\documentclass[table, 10pt]{article} 
\usepackage[table,xcdraw]{xcolor}
\usepackage[preprint]{rlc}

\usepackage{amssymb}            
\usepackage{mathtools}          
\usepackage{mathrsfs}           
\mathtoolsset{showonlyrefs}     
\usepackage{graphicx}           
\usepackage{subcaption}         
\usepackage[space]{grffile}     
\usepackage{url}                

\usepackage{tabularx, makecell}
\usepackage[utf8]{inputenc} 
\usepackage[T1]{fontenc}    
\usepackage{hyperref}       
\usepackage{url}            
\usepackage{booktabs}       
\usepackage{amsfonts}       
\usepackage{nicefrac}       
\usepackage{microtype}      
\usepackage{placeins}
\usepackage{enumitem}
\usepackage{comment}
\usepackage{todonotes}
\usepackage{blindtext}
\usepackage{caption}
\usepackage{subcaption}
\usepackage{natbib}
\usepackage{amssymb}
\usepackage{amsbsy}
\usepackage{upgreek}
\usepackage{multirow}
\usepackage{wrapfig}

\usepackage{soul}

\usepackage{ifthen}
\newboolean{anonymized} 
\setboolean{anonymized}{false} 

\definecolor{blue}{RGB}{0,0,255}

\newcommand{\rebut}[1]{#1}
\newcommand{\eg}{\emph{e.g.}~} 

\newcommand{\ie}{\emph{i.e.}~} 

\newcommand{\cf}{\emph{cf.}~}

\newcommand{\anonymizedlink}[1]{\ifanonymized \small \url{https://anonymous.4open.science/r/OCAtari-52B9} \else #1~\fi}

\title{OCAtari: Object-Centric Atari 2600 \\ Reinforcement Learning Environments}

\author{\\\name Quentin Delfosse\thanks{Equal contribution}$^{\ ,1}$, \ \textbf{Jannis Blüml}\(^{*, 1,2}\), \ \textbf{Bjarne Gregori}$^1$, \\ \textbf{Sebastian Sztwiertnia}$^1$ \ \textbf{\&} \ \textbf{Kristian Kersting}$^{1,2,3,4}$ \\
\addr \href{mailto:quentin.delfosse@cs.tu-darmstadt.de}{quentin.delfosse@cs.tu-darmstadt.de}, \href{mailto:blueml@cs.tu-darmstadt.de}{blueml@cs.tu-darmstadt.de} \\ \\
$^1$AI and ML Group, Technical University of Darmstadt, Germany \\
$^2$Hessian Center for Artificial Intelligence (hessian.AI) \\
$^3$Centre for Cognitive Science of Darmstadt \\
$^4$German Research Center for Artificial Intelligence (DFKI)  
}

\begin{document}

\maketitle

\begin{abstract}
    Cognitive science and psychology suggest that object-centric representations of complex scenes are a promising step towards enabling efficient abstract reasoning from low-level perceptual features. Yet, most deep reinforcement learning approaches only rely on pixel-based representations that do not capture the compositional properties of natural scenes. 
    For this, we need environments and datasets that allow us to work and evaluate object-centric approaches. 
    In our work, we extend the Atari Learning Environments, the most-used evaluation framework for deep RL approaches, by introducing OCAtari, that performs resource-efficient extractions of the object-centric states for these games. Our framework allows for object discovery, object representation learning, as well as object-centric RL. 
    We evaluate OCAtari's detection capabilities and resource efficiency. 
    Our source code is available at \anonymizedlink{\href{https://github.com/k4ntz/OC_Atari}{\url{github.com/k4ntz/OC_Atari}}}.
\end{abstract}


\section{Introduction}

\begin{figure}[t]
    \centering
         \includegraphics[width=1.\textwidth]{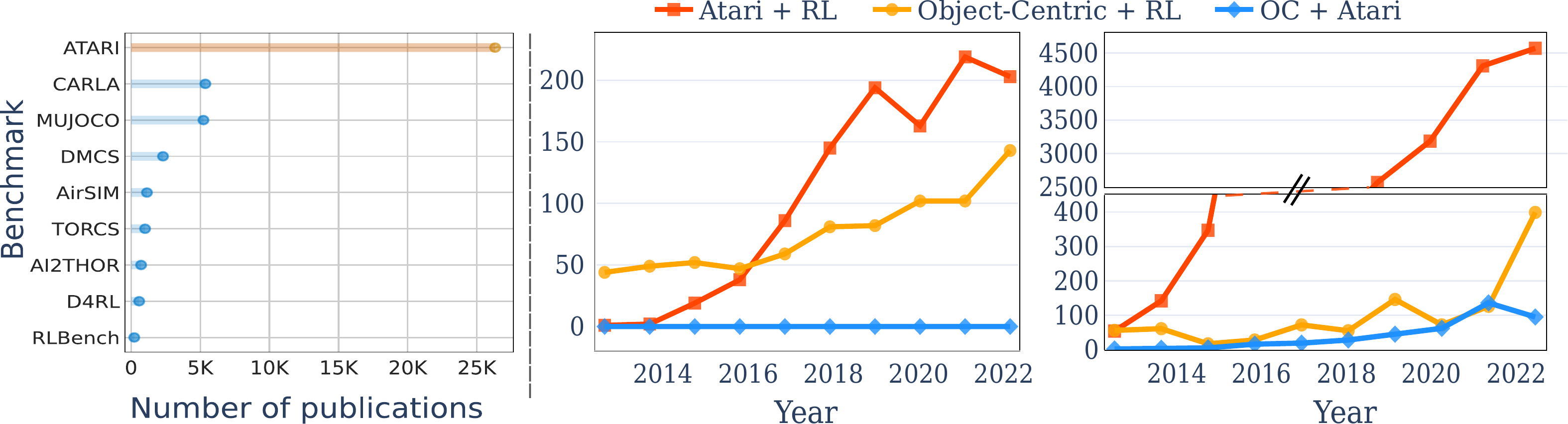}
         \label{fig:oc_publication}
     \caption{\textbf{RL research needs Object-Centric Atari environments.} The Atari Learning Environments (ALE) is, by far, the most used RL benchmark among the ones listed on \url{paperswithcode.com} (left). Publications using ALE are increasing, together with the number of papers concerned on object-centric RL. As no Object-centric ALE is available yet, the amount of papers concerned with object-centric approaches in Atari is however negligible. Data queried
 using \url{dimensions.ai}, based on keyword occurrence in title and abstract (center) or in full text (right). These graphs show that RL researchers would make use of object-centric atari environments, if they would be available.}
\label{fig:publications}
\vspace{-0.3cm}
\end{figure}

Since the introduction of the Arcade Learning Environment (ALE) by~\citet{BellemareNVB13}, Atari 2600 games have become the most common environments to test and evaluate RL algorithms (\cf~Figure~\ref{fig:publications}, left). 
As RL methods are challenging to evaluate, compare and reproduce, benchmarks need to encompass a variety tasks and challenges to allow for balancing advantages and drawbacks of the different approaches~\citep{Henderson18deepbenchmarking, PineauReproducible21}. ALE games incorporate many RL challenges, such as difficult credit assignment (Skiing), sparse reward (Montezuma’s Revenge, Pitfall), and allow for testing approaches with different focuses, such as partial observability~\citep{HausknechtS15}, generalization~\citep{Farebrother2018Generalization}, sample efficiency~\citep{Espeholt2018impala}, environment modeling~\citep{Hafner2021Mastering, Schrittwieser2020muzero}, ...etc.

In order to solve complex tasks, human use \textit{abstraction}, \ie they first extract object-centred representations and abstract relational concepts, on which they base their reasoning~\citep{grill2005visual,tenenbaum2011abstraction, lake2017building}. 
Deep reinforcement learning (RL) agents do not incorporate explicit object-centric intermediate representations, necessary to check if suboptimal behaviors are \eg caused by misdetections, wrong object identifications, or a reasoning failure. 
Numerous studies on RL research highlight the importance of object-centricity (\cf Figure~\ref{fig:publications}, right), notably in understanding the agents' reasoning, detect potential misalignment and potentially correct it~\citep{Langosco2022goal} . 
Notably,
\citet{Delfosse2024InterpretableCB} show that deep agents, that do not make use of interpretable object centric representations, can learn misaligned policies on games as simple as Pong, that post-hoc explanation techniques cannot detect.
Object-centricity also permits to use logic to encode the policy, leading to interpretable agents with better generalization capability~\citep{Delfosse2023InterpretableAE}, and ease knowledge transfer between humans and learning agents, or among different tasks~\citep{dubeyICLRW18human}.
Further studies also underline that the extraction of object-centric states is a necessary step to obtain agent that can make use of large language model together with contextual data (\eg the games instruction manuals) to improve the reward signals, notably allowing agents to learn in difficult credit assignment environments~\citep{ZhongHWNZ21, Wu23Read}.
This underscores the need to produce transparent object-centric RL agents, that can ensure their proper alignment with the intended objectives.

More specifically on the set of Atari RL environments, 
\cite{lake2017building} illustrated that deep agents trained on ALE games lack the ability to create multi-step sub-goals (such as acquiring certain objects while avoiding others) and introduced the ``Frostbite challenge'' to assess that RL agents integrate such human-like capabilities. \citet{Badia2020agent57} also suggested to enhance the internal representations of suboptimal ALE trained agents.

As no benchmark to test object-centric methods exists yet, we introduce OCAtari, a set of object-centric versions of the ALE environments. 
OCAtari runs the ALE games while maintaining object-centric states (\ie a list of the depicted objects and their properties). 
Our framework can be used to train object-centric RL algorithms, making it a resource-efficient replacement for otherwise necessary object discovery methods. To train and evaluate these object discovery methods, we also propose the Object-centric Dataset for Atari (ODA), that uses OCAtari to generate a set of Atari frames, together with the properties of the objects present in each game.

Our contributions can be summarized as follows:
\begin{itemize}[leftmargin=15pt,itemsep=4pt,parsep=2pt,topsep=3pt,partopsep=2pt]
\item We introduce OCAtari, an RL framework to train and evaluate object-detection and object-centered RL methods on the widely-used Arcade Learning Environments. 
\item We evaluate OCAtari capability to detect the depicted game objects in a resource efficient way and demonstrate that it allows for object-centric RL.
\item To ease the comparison of object-discovery methods, we introduce ODA, a collection of frames from Atari games together with their object-centric states. 
\end{itemize}

We start off by introducing the Object-Centric Atari framework. We experimentally evaluate its detection and speed performances. Before concluding, we touch upon related work. 

\section{The Object-Centric Atari Environments}
\label{sec:envs}
We here discuss the definition of objects and how they can be used in RL, then introduce the OCAtari benchmark, and detail its two extraction methods.

\subsection{Using Object-Centric Descriptions to Learn}
According to~\citet{thiel2011early}, objects are physical entities that possess properties, attributes, and behaviors that can be observed, measured, and described. 
\cite{Rettler2017RETO} define objects as the fundamental building blocks that human reasoning relies on. 
Breaking down the world into objects enables abstraction, generalization, cognitive efficiency, understanding of cause and effect, clear communication, logical inference, and more (\citet{spelke1992origins, grill2005visual, tenenbaum2011abstraction, lake2017building}, \cf Appendix~\ref{app:objects_advantages} for further details).

 \begin{figure}[t]
     \centering
     \includegraphics[width=0.9\linewidth]{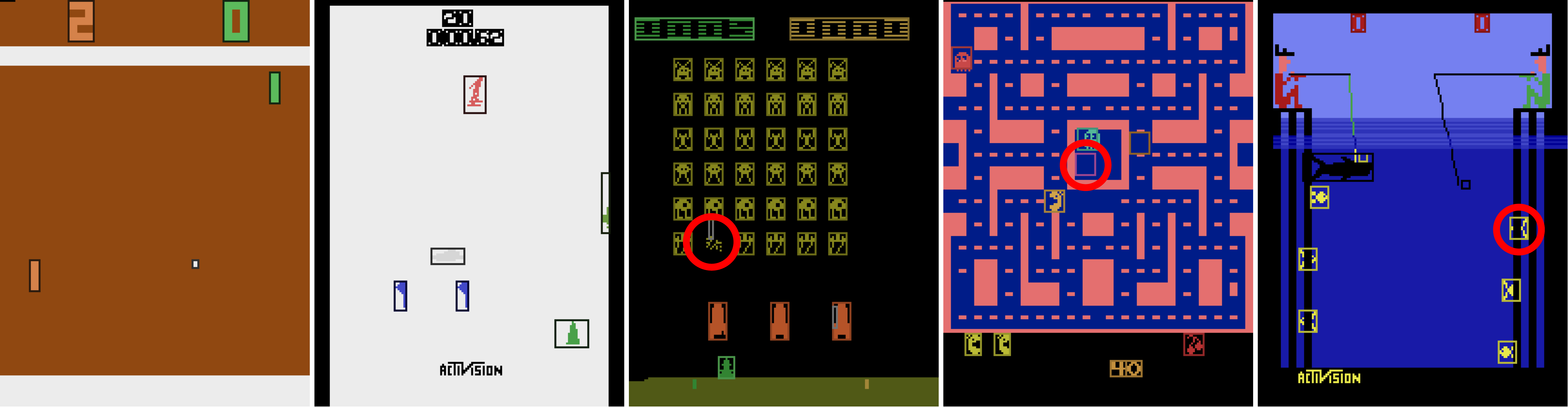}
     \caption{\textbf{Qualitative evaluation of OCAtari's REM.} Frames from our OCAtari framework on $5$ environments (Pong, Skiing, SpaceInvaders, MsPacman, FishingDerby). Bounding boxes surround the detected objects. \rebut{REM automatically detects blinking (MsPacman), occluded (FishingDerby) objects, and ignore \eg exploded objects (SpaceInvaders) that vision methods falsely can pick up.}}
     \label{fig:idea}
     \vspace{-4mm}
 \end{figure}


\begin{wrapfigure}{r}{0.36\linewidth}
\vspace{-4mm}
    \centering
    \includegraphics[width=.9\linewidth]{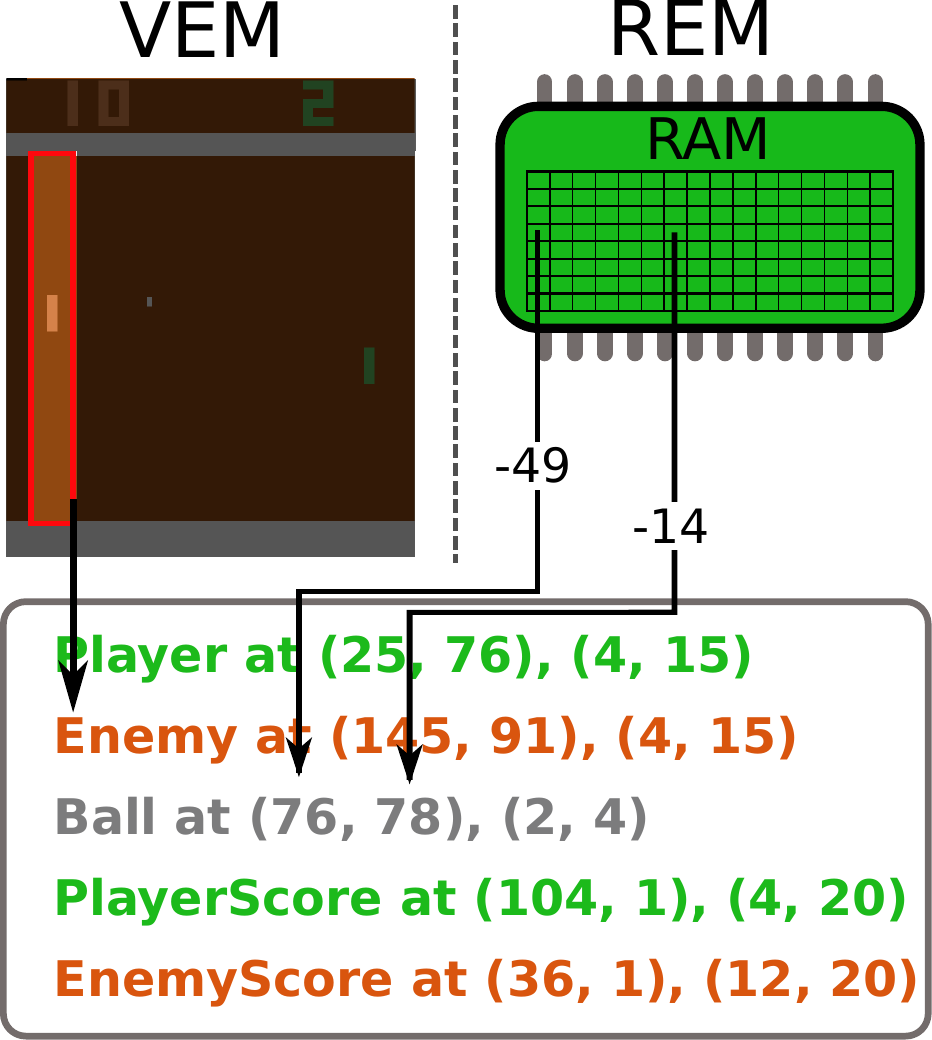}
    \vspace{-1mm}
    \caption{\textbf{OCAtari extract object-centric descriptions}: using its RAM Extraction method (REM) or Vision Extraction method (VEM).}  
    \vspace{-10mm}
    \label{fig:VEMvREM}
\end{wrapfigure}

In artificial approaches, object-centric visual learning often involves the extraction of objects withing bounding boxes that contain them and distinguish them from the background~\citep{lin2020space, Delfosse2021MOC}. In these approaches, static objects, such as the maze in MsPacman or the walls in Pong (\cf~\autoref{fig:idea}), are considered as part of the background. 
In our work, we define \textit{objects} as small elements (relative to the agent) with which it can interact. Excluding "background objects" when learning to play Pong with object-centric inputs is not problematic. However, it can lead to problems when learning on \eg MsPacman. 
The learning agents can learn to incorporate \eg Pong's boundaries when learning to play, but may have difficulties to accurately encode the maze structures of Pacman games. 
As it may be necessary to provide a background representation to the agent, OCAtari provides both renderings and object-centric descriptions of the states.

\subsection{The OCAtari framework}

In OCAtari, every object is defined by its category  (\eg ``Pacman''), position (\texttt{x} and \texttt{y}), size (\texttt{w} and \texttt{h}), and its  \texttt{RGB} values. Objects may have additional characteristics such as orientation (\eg the Player in Skiing, \cf \autoref{fig:idea}) or value (\eg oxygen bars or scores) if required. Objects that are vital for gameplay are distinguished from those that are components of the Head-up-Display (HUD) 
elements (\eg score, number of lives). 
The role of HUD objects is to provide additional information about the performance of the playing agent. 
Although learning agents should, in principle, be capable of ignoring such elements, in our environments a boolean parameter is available to filter out HUD objects. 
A list of the considered objects for each game can be found in \autoref{app:detailed_results}.

To extract objects, OCAtari uses either its Vision Extraction Method (VEM) or its resource efficient RAM Extraction Method (REM), that are depicted in Figure~\ref{fig:VEMvREM}. 

\textbf{VEM: the Vision Extraction Method.} The most straightforward method for extracting objects from Atari frames involves using simple computer vision techniques. 
Considering the limited memory available to Atari developers, most objects are defined by a restricted set of pre-established colors (i.e., RGB values). 
At each stage, the Vision Extraction Method extracts objects using color-based filtering and priors about the objects' positions. 
For example, Pong consists of $3$ moving objects and $2$ HUD objects, each assigned fixed RGB values (\cf Figure~\ref{fig:VEMvREM}). 
The enemy's paddle and scores share common RGB values (orange in~\autoref{fig:VEMvREM}), but contrary to the scores, the paddles always appears between the white threshold. The enemy's paddle is always positioned within the red rectangle. 
Using this technique, it is possible to accurately extract all present objects. 
This detection method can only detect what is depicted in the frame, and not objects that are \eg blinking, overlapping, etc.

\textbf{REM: the RAM Extraction Method.} 
ALE provides the state of the emulator's RAM, which contains information about the games' objects. This has led~\citet{Sygnowski2016LearningFT} to use the raw RAM states for RL states to train agents. 
However, much of the non-relevant information is present in the RAM (\eg time counter, HUD element information). Moreover, several games, use \eg bitmaps or encode various information quantities such as object orientation, offset from the anchor, and object category together within one byte. 
These noisy inputs and entangled representations prevent obscure these agents decision process and remove any interpretation possibilities.  
To address these problems,~\citet{Anand19AtariARI} have proposed AtariARI, a wrapper around some Atari environments, that provides some the RAM positions, describing where some specific information is encoded.
Nonetheless, raw RAM information is not enough.
Take, for instance, in \textit{Kangaroo}, the player's position corresponds to various RAM values, that also encode its heights using categorical values. Simply providing some uninfluenced RAM positions does not reflect the object-centric state. 
Similar to AtariARI, our Ram Extraction Method extracts the information from the RAM, but processes it to provide an interpretable object-centric state, that matches VEM's one (\cf Figure~\ref{fig:VEMvREM}).
To determine how the game's program processes the RAM information, we task human, random, or DQN agents with playing the games while using VEM to track the depicted objects. We then establish correlations between objects properties (\eg positions) and each of the $128$ bytes of the Atari RAM representation. We can also modify each RAM byte and track the resulting changes in the rendered frames. All these scripts are documented and released along with this manuscript.

REM, being based on semantic information, allows for tracking moving objects. Conversely, VEM only furnishes consecutive object-centric descriptions, where the lists of objects are independently extracted at each state. 
REM thus enables tracking of blinking objects and moving instances, as proven useful for RL approaches using tracklets~\citep{Agnew2020RelevanceGuidedMO, Liu2021SemanticTA}.

\textbf{The OCAtari package.}
We provide an easy-to-use \texttt{ocatari} package, with its documentation\footnote{\anonymizedlink{\url{https://oc-atari.readthedocs.io}}}. OCAtari includes wrappers for the Arcade Learning Environments (ALE) of~\citet{BellemareNVB13}. 
To allow an easy swap between ALE and OCAtari environments, we follow the logic and naming system of ALE. 
We have reimplemented its methods for OCAtari (\eg~\texttt{step}, \texttt{render}, \texttt{seed}, \dots), and added new methods like \texttt{get\_ram} and \texttt{set\_ram}, to easily allow RAM lookup and manipulation. \texttt{OCAtari} environments also maintain a list of the depicted objects and can provide a buffer of the last $4$ transformed (\ie black and white, $84\!\times\!84$) frames of the game, as it has become a standard of RL state representations~\citep{Mnih2015dqn, vanHasseltGS16ddqn, hessel2018rainbow}.

As shown in \autoref{tab:supported games}, our image processing method VEM covers \rebut{46} games, while REM covers \rebut{44} games at the time of writing. 
While these already constitute a diverse set of environments, we will continue to add newly supported games in both REM and VEM and complete what we have started. 
Along with this work, we release ODA, a dataset that contains frames with the object-centric states obtained from REM and VEM, collected using Random and trained DQN agents (\cf ~\autoref{app:oda} for further details). ODA and OCAtari are openly accessible under the MIT license.

\section{Evaluating OCAtari}
\label{sec:eval}

In this section, we evaluate the detection and speed performances of OCAtari methods, then explain how it can be used for object-centric RL agents training. Finally, we compare OCAtari to AtariARI.

\textbf{Setup}.
To evaluate the detection capabilities of REM, we use a random agent (that represents any untrained RL agent), as well as a DQN and, if available, a C51 agent~\citep{BellemareDM17}, both obtained from~\citet{gogianu2022agents}\footnote{\url{https://github.com/floringogianu/atari-agents}}. 
For reproducibility, every used agent is provided with our along with our codebase. The RL experiments utilized the PPO implementation from stable-baselines3~\citep{stable-baselines3} on a 40 GB DGX A100 server. In each seeded run, $1$ critic and $8$ actors are utilized per seed over $20$M frames. 
Since these experiments do not involve visual representation learning, we utilize the default $2\times64$ MLP architecture (with the hyperbolic tangent as activation functions). As developing RL agents is not our focus, we did not conduct any fine-tuning or hyperparameter search. 
Further details on these experiments can be found in Appendix \ref{app:exp_details}.

\subsection{Evaluating OCAtari for Object Extraction }

\textbf{Correctness and Completeness of the Object Extraction.}
As explained previously, REM needs to decode the game objects' properties from RAM values. For example, objects' position in \eg Riverraid either require adding an offset (for the agent) or being reconstructed from anchor and offsets position in a grid.
To assert that REM accurately reconstruct these values, we compare the object-centric states of both extraction methods (VEM and REM). We let the Random, and trained DQN and C51 agents play for $500$ frames, and compute IOU~\citep{Rezatofighi2019IOU} for each agent on each game. As this metric's relevance is debatable for small objects (\eg~the ball in Pong, Tennis, or missiles in Atlantis, Space Invaders), we also calculate precision, recall, and F1-scores for each object category in every game. 
For these metrics, an object is considered correctly detected if it is within $5$ pixels of the center for both detection methods. 

\begin{table}
\centering 
\small
 \setlength{\tabcolsep}{1.2pt}
\caption{\textbf{REM reliably detects the objects within the frames of each developed games}. Measuring precision, Recall, F1-Score and IOU of REM (using VEM as baseline) in a diverse set of Atari games using trained DQN agents. 
High values being displayed in blue going over green to red for low values. A more detailed table, with Radom and C51 agents is provided in~\autoref{app:detailed_results}.}

\hspace{-0.6cm}
\begin{subtable}{0.99\linewidth}
\begin{tabular}{lrrrrrrrrrrrrrrrrrrrrrrrrrrrrrrrrrrrrrrrrrrr}

 & \rotatebox{90}{Alien} & \rotatebox{90}{Amidar} & \rotatebox{90}{Assau.} & \rotatebox{90}{Aster.} & \rotatebox{90}{Atlan.} & \rotatebox{90}{BankH.} & \rotatebox{90}{Battl.} & \rotatebox{90}{Berze.} & \rotatebox{90}{Bowli.} & \rotatebox{90}{Boxing} & \rotatebox{90}{Break.} & \rotatebox{90}{Carni.} & \rotatebox{90}{Centi.} & \rotatebox{90}{Chopp.} & \rotatebox{90}{Crazy.} & \rotatebox{90}{Demon.} & \rotatebox{90}{Donke.} & \rotatebox{90}{Fishi.} & \rotatebox{90}{Freew.} & \rotatebox{90}{Frost.} & \rotatebox{90}{Gopher} & \rotatebox{90}{Hero} & \rotatebox{90}{IceHo.} & \rotatebox{90}{James.} & \rotatebox{90}{Kanga.} & \rotatebox{90}{Krull} & \rotatebox{90}{Monte.} & \rotatebox{90}{MsPac.} & \rotatebox{90}{Pacma.} & \rotatebox{90}{Pitfa.} & \rotatebox{90}{Pong} & \rotatebox{90}{Priva.} & \rotatebox{90}{Qbert} & \rotatebox{90}{River.} & \rotatebox{90}{RoadR.} & \rotatebox{90}{Seaqu.} & \rotatebox{90}{Skiing} & \rotatebox{90}{Space.} & \rotatebox{90}{Tennis} & \rotatebox{90}{TimeP.} & \rotatebox{90}{UpNDo.} & \rotatebox{90}{Ventu.} & \rotatebox{90}{Video.} \\

Prec. & {\cellcolor[HTML]{FCFEBA}} \color[HTML]{000000}  & {\cellcolor[HTML]{459EB4}} \color[HTML]{F1F1F1}  & {\cellcolor[HTML]{525FA9}} \color[HTML]{F1F1F1}  & {\cellcolor[HTML]{496AAF}} \color[HTML]{F1F1F1}  & {\cellcolor[HTML]{5061AA}} \color[HTML]{F1F1F1}  & {\cellcolor[HTML]{4E63AC}} \color[HTML]{F1F1F1}  & {\cellcolor[HTML]{5CB7AA}} \color[HTML]{F1F1F1}  & {\cellcolor[HTML]{466EB1}} \color[HTML]{F1F1F1}  & {\cellcolor[HTML]{5B53A4}} \color[HTML]{F1F1F1}  & {\cellcolor[HTML]{4E63AC}} \color[HTML]{F1F1F1}  & {\cellcolor[HTML]{5C51A3}} \color[HTML]{F1F1F1}  & {\cellcolor[HTML]{486CB0}} \color[HTML]{F1F1F1}  & {\cellcolor[HTML]{4D65AD}} \color[HTML]{F1F1F1}  & {\cellcolor[HTML]{71C6A5}} \color[HTML]{000000}  & {\cellcolor[HTML]{555AA7}} \color[HTML]{F1F1F1}  & {\cellcolor[HTML]{E7F59A}} \color[HTML]{000000}  & {\cellcolor[HTML]{5956A5}} \color[HTML]{F1F1F1}  & {\cellcolor[HTML]{378EBB}} \color[HTML]{F1F1F1}  & {\cellcolor[HTML]{5956A5}} \color[HTML]{F1F1F1}  & {\cellcolor[HTML]{3D95B8}} \color[HTML]{F1F1F1}  & {\cellcolor[HTML]{5758A6}} \color[HTML]{F1F1F1}  & {\cellcolor[HTML]{6BC4A5}} \color[HTML]{000000}  & {\cellcolor[HTML]{3D79B6}} \color[HTML]{F1F1F1}  & {\cellcolor[HTML]{4175B4}} \color[HTML]{F1F1F1}  & {\cellcolor[HTML]{5758A6}} \color[HTML]{F1F1F1}  & {\cellcolor[HTML]{4B68AE}} \color[HTML]{F1F1F1}  & {\cellcolor[HTML]{5E4FA2}} \color[HTML]{F1F1F1}  & {\cellcolor[HTML]{9CD7A4}} \color[HTML]{000000}  & {\cellcolor[HTML]{FCFEBA}} \color[HTML]{000000}  & {\cellcolor[HTML]{5E4FA2}} \color[HTML]{F1F1F1}  & {\cellcolor[HTML]{466EB1}} \color[HTML]{F1F1F1}  & {\cellcolor[HTML]{5061AA}} \color[HTML]{F1F1F1}  & {\cellcolor[HTML]{89D0A4}} \color[HTML]{000000}  & {\cellcolor[HTML]{358BBC}} \color[HTML]{F1F1F1}  & {\cellcolor[HTML]{49A2B2}} \color[HTML]{F1F1F1}  & {\cellcolor[HTML]{3A7EB8}} \color[HTML]{F1F1F1}  & {\cellcolor[HTML]{4471B2}} \color[HTML]{F1F1F1}  & {\cellcolor[HTML]{3585BB}} \color[HTML]{F1F1F1}  & {\cellcolor[HTML]{4471B2}} \color[HTML]{F1F1F1}  & {\cellcolor[HTML]{3880B9}} \color[HTML]{F1F1F1}  & {\cellcolor[HTML]{4175B4}} \color[HTML]{F1F1F1}  & {\cellcolor[HTML]{ECF7A1}} \color[HTML]{000000}  & {\cellcolor[HTML]{5C51A3}} \color[HTML]{F1F1F1}  \\
Rec. & {\cellcolor[HTML]{525FA9}} \color[HTML]{F1F1F1}  & {\cellcolor[HTML]{5E4FA2}} \color[HTML]{F1F1F1}  & {\cellcolor[HTML]{4273B3}} \color[HTML]{F1F1F1}  & {\cellcolor[HTML]{5C51A3}} \color[HTML]{F1F1F1}  & {\cellcolor[HTML]{486CB0}} \color[HTML]{F1F1F1}  & {\cellcolor[HTML]{4E63AC}} \color[HTML]{F1F1F1}  & {\cellcolor[HTML]{FBFDB8}} \color[HTML]{000000}  & {\cellcolor[HTML]{5061AA}} \color[HTML]{F1F1F1}  & {\cellcolor[HTML]{5956A5}} \color[HTML]{F1F1F1}  & {\cellcolor[HTML]{4EA7B0}} \color[HTML]{F1F1F1}  & {\cellcolor[HTML]{5E4FA2}} \color[HTML]{F1F1F1}  & {\cellcolor[HTML]{4E63AC}} \color[HTML]{F1F1F1}  & {\cellcolor[HTML]{525FA9}} \color[HTML]{F1F1F1}  & {\cellcolor[HTML]{69C3A5}} \color[HTML]{000000}  & {\cellcolor[HTML]{486CB0}} \color[HTML]{F1F1F1}  & {\cellcolor[HTML]{76C8A5}} \color[HTML]{000000}  & {\cellcolor[HTML]{5956A5}} \color[HTML]{F1F1F1}  & {\cellcolor[HTML]{4EA7B0}} \color[HTML]{F1F1F1}  & {\cellcolor[HTML]{3F97B7}} \color[HTML]{F1F1F1}  & {\cellcolor[HTML]{545CA8}} \color[HTML]{F1F1F1}  & {\cellcolor[HTML]{FFF7B2}} \color[HTML]{000000}  & {\cellcolor[HTML]{3990BA}} \color[HTML]{F1F1F1}  & {\cellcolor[HTML]{5E4FA2}} \color[HTML]{F1F1F1}  & {\cellcolor[HTML]{555AA7}} \color[HTML]{F1F1F1}  & {\cellcolor[HTML]{4175B4}} \color[HTML]{F1F1F1}  & {\cellcolor[HTML]{5061AA}} \color[HTML]{F1F1F1}  & {\cellcolor[HTML]{5E4FA2}} \color[HTML]{F1F1F1}  & {\cellcolor[HTML]{5C51A3}} \color[HTML]{F1F1F1}  & {\cellcolor[HTML]{3990BA}} \color[HTML]{F1F1F1}  & {\cellcolor[HTML]{5E4FA2}} \color[HTML]{F1F1F1}  & {\cellcolor[HTML]{5956A5}} \color[HTML]{F1F1F1}  & {\cellcolor[HTML]{5956A5}} \color[HTML]{F1F1F1}  & {\cellcolor[HTML]{5758A6}} \color[HTML]{F1F1F1}  & {\cellcolor[HTML]{555AA7}} \color[HTML]{F1F1F1}  & {\cellcolor[HTML]{6EC5A5}} \color[HTML]{000000}  & {\cellcolor[HTML]{5EB9A9}} \color[HTML]{F1F1F1}  & {\cellcolor[HTML]{466EB1}} \color[HTML]{F1F1F1}  & {\cellcolor[HTML]{4D65AD}} \color[HTML]{F1F1F1}  & {\cellcolor[HTML]{5956A5}} \color[HTML]{F1F1F1}  & {\cellcolor[HTML]{466EB1}} \color[HTML]{F1F1F1}  & {\cellcolor[HTML]{555AA7}} \color[HTML]{F1F1F1}  & {\cellcolor[HTML]{5E4FA2}} \color[HTML]{F1F1F1}  & {\cellcolor[HTML]{496AAF}} \color[HTML]{F1F1F1}  \\
F1 & {\cellcolor[HTML]{BCE4A0}} \color[HTML]{000000}  & {\cellcolor[HTML]{3F77B5}} \color[HTML]{F1F1F1}  & {\cellcolor[HTML]{496AAF}} \color[HTML]{F1F1F1}  & {\cellcolor[HTML]{525FA9}} \color[HTML]{F1F1F1}  & {\cellcolor[HTML]{4B68AE}} \color[HTML]{F1F1F1}  & {\cellcolor[HTML]{4E63AC}} \color[HTML]{F1F1F1}  & {\cellcolor[HTML]{D1ED9C}} \color[HTML]{000000}  & {\cellcolor[HTML]{4B68AE}} \color[HTML]{F1F1F1}  & {\cellcolor[HTML]{5B53A4}} \color[HTML]{F1F1F1}  & {\cellcolor[HTML]{3387BC}} \color[HTML]{F1F1F1}  & {\cellcolor[HTML]{5E4FA2}} \color[HTML]{F1F1F1}  & {\cellcolor[HTML]{4B68AE}} \color[HTML]{F1F1F1}  & {\cellcolor[HTML]{5061AA}} \color[HTML]{F1F1F1}  & {\cellcolor[HTML]{6EC5A5}} \color[HTML]{000000}  & {\cellcolor[HTML]{4E63AC}} \color[HTML]{F1F1F1}  & {\cellcolor[HTML]{BAE3A1}} \color[HTML]{000000}  & {\cellcolor[HTML]{5956A5}} \color[HTML]{F1F1F1}  & {\cellcolor[HTML]{439BB5}} \color[HTML]{F1F1F1}  & {\cellcolor[HTML]{3F77B5}} \color[HTML]{F1F1F1}  & {\cellcolor[HTML]{3D79B6}} \color[HTML]{F1F1F1}  & {\cellcolor[HTML]{CDEB9D}} \color[HTML]{000000}  & {\cellcolor[HTML]{54AEAD}} \color[HTML]{F1F1F1}  & {\cellcolor[HTML]{4D65AD}} \color[HTML]{F1F1F1}  & {\cellcolor[HTML]{4B68AE}} \color[HTML]{F1F1F1}  & {\cellcolor[HTML]{4B68AE}} \color[HTML]{F1F1F1}  & {\cellcolor[HTML]{4E63AC}} \color[HTML]{F1F1F1}  & {\cellcolor[HTML]{5E4FA2}} \color[HTML]{F1F1F1}  & {\cellcolor[HTML]{52ABAE}} \color[HTML]{F1F1F1}  & {\cellcolor[HTML]{C8E99E}} \color[HTML]{000000}  & {\cellcolor[HTML]{5E4FA2}} \color[HTML]{F1F1F1}  & {\cellcolor[HTML]{5061AA}} \color[HTML]{F1F1F1}  & {\cellcolor[HTML]{545CA8}} \color[HTML]{F1F1F1}  & {\cellcolor[HTML]{4BA4B1}} \color[HTML]{F1F1F1}  & {\cellcolor[HTML]{4273B3}} \color[HTML]{F1F1F1}  & {\cellcolor[HTML]{5CB7AA}} \color[HTML]{F1F1F1}  & {\cellcolor[HTML]{459EB4}} \color[HTML]{F1F1F1}  & {\cellcolor[HTML]{466EB1}} \color[HTML]{F1F1F1}  & {\cellcolor[HTML]{4175B4}} \color[HTML]{F1F1F1}  & {\cellcolor[HTML]{4E63AC}} \color[HTML]{F1F1F1}  & {\cellcolor[HTML]{3F77B5}} \color[HTML]{F1F1F1}  & {\cellcolor[HTML]{496AAF}} \color[HTML]{F1F1F1}  & {\cellcolor[HTML]{94D4A4}} \color[HTML]{000000}  & {\cellcolor[HTML]{545CA8}} \color[HTML]{F1F1F1}  \\
IOU & {\cellcolor[HTML]{545CA8}} \color[HTML]{F1F1F1}  & {\cellcolor[HTML]{3D79B6}} \color[HTML]{F1F1F1}  & {\cellcolor[HTML]{4471B2}} \color[HTML]{F1F1F1}  & {\cellcolor[HTML]{4E63AC}} \color[HTML]{F1F1F1}  & {\cellcolor[HTML]{496AAF}} \color[HTML]{F1F1F1}  & {\cellcolor[HTML]{486CB0}} \color[HTML]{F1F1F1}  & {\cellcolor[HTML]{4273B3}} \color[HTML]{F1F1F1}  & {\cellcolor[HTML]{76C8A5}} \color[HTML]{000000}  & {\cellcolor[HTML]{5C51A3}} \color[HTML]{F1F1F1}  & {\cellcolor[HTML]{4273B3}} \color[HTML]{F1F1F1}  & {\cellcolor[HTML]{5E4FA2}} \color[HTML]{F1F1F1}  & {\cellcolor[HTML]{3A7EB8}} \color[HTML]{F1F1F1}  & {\cellcolor[HTML]{4D65AD}} \color[HTML]{F1F1F1}  & {\cellcolor[HTML]{439BB5}} \color[HTML]{F1F1F1}  & {\cellcolor[HTML]{5061AA}} \color[HTML]{F1F1F1}  & {\cellcolor[HTML]{50A9AF}} \color[HTML]{F1F1F1}  & {\cellcolor[HTML]{5B53A4}} \color[HTML]{F1F1F1}  & {\cellcolor[HTML]{91D3A4}} \color[HTML]{000000}  & {\cellcolor[HTML]{3387BC}} \color[HTML]{F1F1F1}  & {\cellcolor[HTML]{4199B6}} \color[HTML]{F1F1F1}  & {\cellcolor[HTML]{50A9AF}} \color[HTML]{F1F1F1}  & {\cellcolor[HTML]{459EB4}} \color[HTML]{F1F1F1}  & {\cellcolor[HTML]{C1E6A0}} \color[HTML]{000000}  & {\cellcolor[HTML]{486CB0}} \color[HTML]{F1F1F1}  & {\cellcolor[HTML]{486CB0}} \color[HTML]{F1F1F1}  & {\cellcolor[HTML]{358BBC}} \color[HTML]{F1F1F1}  & {\cellcolor[HTML]{555AA7}} \color[HTML]{F1F1F1}  & {\cellcolor[HTML]{56B0AD}} \color[HTML]{F1F1F1}  & {\cellcolor[HTML]{76C8A5}} \color[HTML]{000000}  & {\cellcolor[HTML]{5061AA}} \color[HTML]{F1F1F1}  & {\cellcolor[HTML]{56B0AD}} \color[HTML]{F1F1F1}  & {\cellcolor[HTML]{4B68AE}} \color[HTML]{F1F1F1}  & {\cellcolor[HTML]{5758A6}} \color[HTML]{F1F1F1}  & {\cellcolor[HTML]{3682BA}} \color[HTML]{F1F1F1}  & {\cellcolor[HTML]{3F97B7}} \color[HTML]{F1F1F1}  & {\cellcolor[HTML]{3880B9}} \color[HTML]{F1F1F1}  & {\cellcolor[HTML]{358BBC}} \color[HTML]{F1F1F1}  & {\cellcolor[HTML]{545CA8}} \color[HTML]{F1F1F1}  & {\cellcolor[HTML]{52ABAE}} \color[HTML]{F1F1F1}  & {\cellcolor[HTML]{486CB0}} \color[HTML]{F1F1F1}  & {\cellcolor[HTML]{4175B4}} \color[HTML]{F1F1F1}  & {\cellcolor[HTML]{3880B9}} \color[HTML]{F1F1F1}  & {\cellcolor[HTML]{496AAF}} \color[HTML]{F1F1F1}  \\

\end{tabular}
\end{subtable}
\hspace*{-0.8cm}
\begin{subtable}{.006\linewidth}
\renewcommand{\arraystretch}{0.36}
\vspace{-10.8mm}
\begin{tabular}{r}
\rotatebox{90}{ Scale} \\
\hline
{\cellcolor[HTML]{5E4FA2}} \color[HTML]{F1F1F1}  \\
{\cellcolor[HTML]{3387BC}} \color[HTML]{F1F1F1}  \\
{\cellcolor[HTML]{66C2A5}} \color[HTML]{000000}  \\
{\cellcolor[HTML]{AADCA4}} \color[HTML]{000000}  \\
{\cellcolor[HTML]{E6F598}} \color[HTML]{000000}  \\
{\cellcolor[HTML]{FFFFBE}} \color[HTML]{000000}  \\
{\cellcolor[HTML]{FEE08B}} \color[HTML]{000000}  \\
{\cellcolor[HTML]{FDAD60}} \color[HTML]{000000}  \\
{\cellcolor[HTML]{F46D43}} \color[HTML]{F1F1F1}  \\
{\cellcolor[HTML]{D43D4F}} \color[HTML]{F1F1F1}  \\
{\cellcolor[HTML]{9E0142}} \color[HTML]{F1F1F1}  \\
\end{tabular}
\end{subtable}

\label{tab:f1}
\end{table}

In \autoref{tab:f1}, we report these metrics for DQN agents averaged over every object category. Similar results, obtained using Random and C51 agents are provided in~\autoref{app:detailed_results}. 
Lower precisions indicate that some objects detected using REM are not detected by VEM, and lower recalls imply the opposite situation.
In MsPacman, the ghost can blink and objects can overlap, which explains why the precision is slightly lower. 
This can be observed in the per-category tables (\cf~\autoref{app:detailed_results}). 
We opted for allowing the RAM extraction method to monitor hidden or blinking objects, regardless of its effects on the precision of our framework, as it can be used to train object tracking methods that employ tracklets (e.g.,~\citealt{Agnew2020RelevanceGuidedMO}) or Kalman filters (e.g.,~\citealt{welch1995introduction}).
The F1-score aggregates both previously mentioned metrics, using a harmonic mean, hastily punishing both for false positives and false negatives. Perfect F1-scores means that every object-centered state extracted using REM is identical to the VEM ones. 

In general, the table results indicate that the games covered by REM have high detection performances. 
Misdetection can be caused by overlap of other objects or the background (\cf~\autoref{fig:idea}, \textit{FishingDerby}). 
Potential rendering instabilities cause slight differences in ball position and size, which decreases the IOU in \eg Pong and Tennis. 
In many games, the rendering freezes after specific events (\eg when the player dies) while the RAM is unaltered. 
Some objects are then not rendered for a few frames, but our RAM extraction approach can keeps them in the list. 
Although this decreases the detection scores, it does not affect gameplay since, for these frames, the environment is not interactive.

In~\autoref{tab:SPACEandSPOC}, we compare the detection performances (F1-scores) of REM~(94\%) on the games used by the $2$ object-discovery methods used on ALE: SPACE~(\citet{SPACE2020}, $31\%$) and SPOC~(\citet{Delfosse2021MOC}, $77\%$). 
REM largely outperforms both. As highlighted by SPOC's authors, the detection of Atari games' objects, composed of few pixels, remains a challenge for neural networks. OCAtari does not extract encodings for objects, but directly provides their classes (from the deterministic RAM information process), that can be used to train these objects discovery methods.

\textbf{Comparing the RAM and Visual Extraction Method.}
As explained in the previous section, REM relies on accurate information decoding, but allows for tracking blinking or overlapping objects.
Its most significant advantage over VEM is the computational efficiency of the RAM extraction procedure. While VEM must perform colour filtering for each object category, REM needs few simple operations to extract objects' properties. 
Getting object-centric states using REM is on average $50$ times faster than with VEM (\cf \autoref{fig:speed} in Appendix~\ref{app:speed_perfs}). 
RL agents can use REM to efficiently train the reasoning part of the policy, as shown bellow, and later be fine-tuned to work with neural-based object extraction. 
To evaluate such extraction methods, on \eg independently drawn frames (without tracking), VEM can reliably extract only the visible objects. The (slower) extraction is then performed only once, as such training is usually run using a dataset, such as ODA. 

\begin{table}[t]
\begin{minipage}{.38\textwidth}
    \centering
    \resizebox{\textwidth}{!}{
    \renewcommand{\arraystretch}{0.76}
    \begin{tabular}{lccc}  \toprule
        Game & SPACE & SPOC & REM  \\
        \midrule
        Boxing & $24.5$ & $70.5$ & $\textbf{90.1}$ \\
        Carnival & $48.6$ & $90.6$ & $\textbf{93.7}$ \\
        MsPacm. & $0.4$ & $\textbf{90.5}$ & $87.4$ \\
        Pong & $10.7$ & $87.4$ & $\textbf{94.3}$ \\
        Riverraid & $45.0$ & $76.6$ & $\textbf{95.7}$ \\
        SpaceInv. & $87.5$ & $85.2$ & $\textbf{96.9}$ \\
        Tennis & $3.6$ & $40.2$ & $\textbf{99.3}$ \\
        \midrule
        Average & $31.5$ & $77.3$ & $\textbf{93.9}$ \\
        \bottomrule
    \end{tabular}}
    \vspace{-1mm}
    \caption{\textbf{Object detection is still challenging in Atari.} SPACE and SPOC, SOTA in object discovery, are inferior in terms of F1 scores.}
    \label{tab:SPACEandSPOC}
\end{minipage}\hfill
\begin{minipage}{.58\textwidth}
\centering
\includegraphics[width=1.\textwidth]{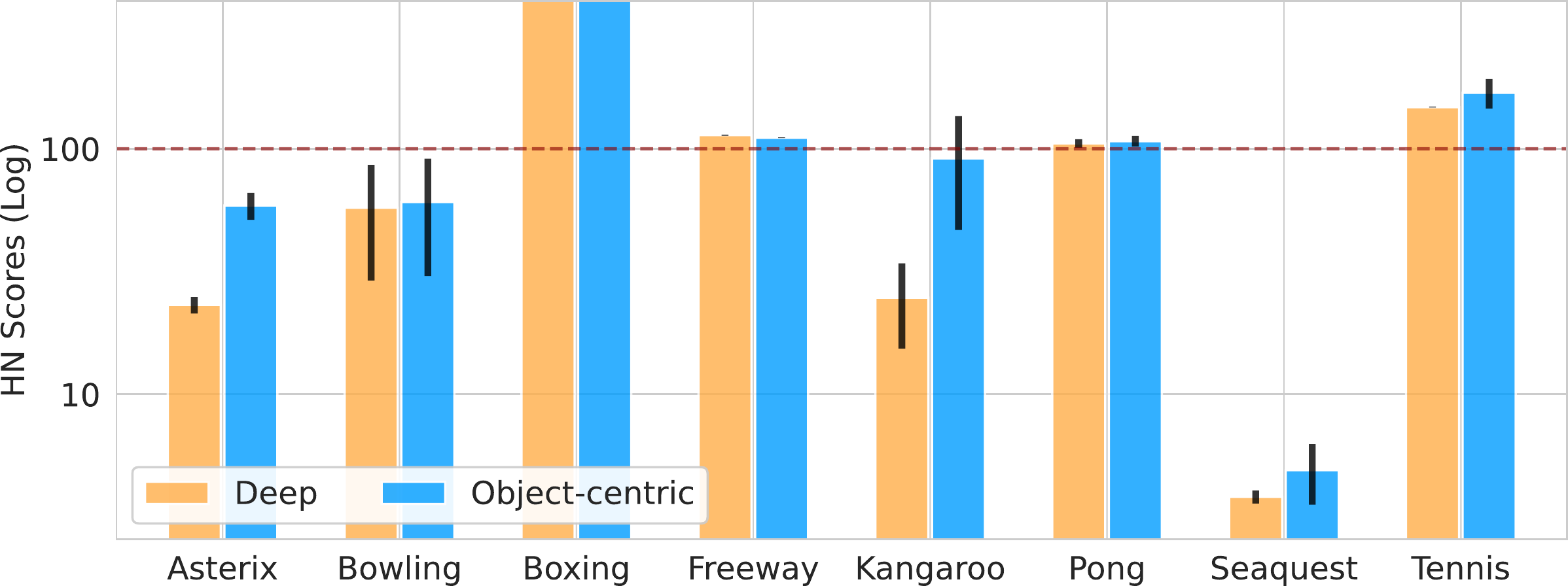}
\captionof{figure}{\textbf{OCAtari (REM) permits learning of object-centric RL agents}. The object-centric PPO agents perform at least on par with the pixel-based PPO (Deep) agents' and humans on $8$ Atari games. }
\label{fig:oc_vs_deep}
\end{minipage}

\vspace{-0.2cm}
\end{table}

\subsection{Using OCAtari to train Object-centric RL agents}

To show that OCAtari allows training object-centric RL agents, we trained RL agents using our REM with $3$ seeded Proximal Policy Optimization (PPO) agents in $8$ different environments. These agents are provided with the positional information of each moving object. Specifically, these correspond to the \texttt{x} and \texttt{y} positions of each object detected by REM in the last two frames at each timestep. Our trained models are available in our public repository, as well as our the scripts used to generate our data sets (\cf \autoref{app:oda}). 
As depicted in Figure~\ref{fig:oc_vs_deep},
OCAtari allows object-centric PPO agents to learn to master Atari games, as they perform on par or better than their deep counterparts. 

Overall, we have shown that OCAtari can be used to train or evaluate any part of an object-centric RL agent, from object extractors (preferably with VEM) to object-centric policies. Since REM allows object tracking, it can also be used on methods that track object through time, and can directly be integrated for resource efficient object-centric policy training.

\newcommand{\rtexttt}[1]{\rotatebox[origin=c]{90}{\texttt{#1}}}

\newcommand{\rtexbttt}[1]{\rotatebox[origin=c]{90}{\texttt{#1}}}

\begin{table}[t]
\setlength{\tabcolsep}{1pt}
\centering
\caption{\textbf{Games supported by AtariARI and OCAtari.} $\checkmark$ describes that all necessary information about the objects are given. $\sim$ denotes that some necessary information to play the game is lacking. We provide detailed explanation for each of these games in \autoref{app:insufficent}. All games missing in this table are neither supported by AtariARI nor OCAtari yet.}

\resizebox{\textwidth}{!}{
\begin{tabular}{c|cccccccccccccccccccccccccccccccccccccccccccccc|c}
\multicolumn{1}{c|}{\rotatebox[origin=c]{90}{\makecell{Extraction \\Method}}} &
  \multicolumn{1}{l}{\rtexbttt{Alien}} &
  \multicolumn{1}{l}{\rtexbttt{Amidar}} &
  \multicolumn{1}{l}{\rtexttt{Assault}} &
  \multicolumn{1}{l}{\rtexttt{Asterix}} &
  \multicolumn{1}{l}{\rtexbttt{Asteroids}} &
  \multicolumn{1}{l}{\rtexttt{Atlantis}} &
  \multicolumn{1}{l}{\rtexbttt{BattleZone}} &
  \multicolumn{1}{l}{\rtexbttt{BankHeist}} &
  \multicolumn{1}{l}{\rtexttt{BeamR.}} &
  \multicolumn{1}{l}{\rtexttt{Berzerk}} &
  \multicolumn{1}{l}{\rtexttt{Bowling}} &
  \multicolumn{1}{l}{\rtexttt{Boxing}} &
  \multicolumn{1}{l}{\rtexttt{Breakout}} &
  \multicolumn{1}{l}{\rtexttt{Carnival}} &
  \multicolumn{1}{l}{\rtexttt{Centipede}} &
  \multicolumn{1}{l}{\rtexttt{ChopperC.}} &
  \multicolumn{1}{l}{\rtexbttt{CrazyC.}} &
  \multicolumn{1}{l}{\rtexttt{DemonA.}} &
  \multicolumn{1}{l}{\rtexttt{DonkeyK.}} &
  \multicolumn{1}{l}{\rtexttt{\ FishingD.}} &
  \multicolumn{1}{l}{\rtexttt{Freeway}} &
  \multicolumn{1}{l}{\rtexttt{Frostbite}} &
  \multicolumn{1}{l}{\rtexbttt{Gopher}} &
  \multicolumn{1}{l}{\rtexttt{Hero}} &
  \multicolumn{1}{l}{\rtexttt{IceHockey}} &
  \multicolumn{1}{l}{\rtexttt{Jamesbond}} &
  \multicolumn{1}{l}{\rtexttt{Kangaroo}} &
  \multicolumn{1}{l}{\rtexttt{Krull}} &
  \multicolumn{1}{l}{\rtexttt{Montezum.}} &
  \multicolumn{1}{l}{\rtexttt{Ms.Pacman}} &
  \multicolumn{1}{l}{\rtexttt{Pacman}} &
  \multicolumn{1}{l}{\rtexttt{Pitfall}} &
  \multicolumn{1}{l}{\rtexttt{Pong}} &
  \multicolumn{1}{l}{\rtexttt{PrivateE.}} &
  \multicolumn{1}{l}{\rtexttt{Q$\ast$Bert}} &
  \multicolumn{1}{l}{\rtexttt{RiverRaid}} &
  \multicolumn{1}{l}{\rtexttt{RoadR.}} &
  \multicolumn{1}{l}{\rtexttt{Seaquest}} &
  \multicolumn{1}{l}{\rtexttt{Skiing}} &
  \multicolumn{1}{l}{\rtexttt{SpaceInv.}} &
  \multicolumn{1}{l}{\rtexttt{Tennis}} &
  \multicolumn{1}{l}{\rtexttt{TimePilot}} &
  \multicolumn{1}{l}{\rtexttt{UpNDown}} &
  \multicolumn{1}{l}{\rtexttt{Venture}} &
  \multicolumn{1}{l}{\rtexttt{VideoP.}} &
  \multicolumn{1}{l}{\rtexbttt{YarsR.}} &
 \multicolumn{1}{|c}{\rotatebox[origin=c]{90}{Sum of G.}}\\ \midrule
ARI &
   &
   &
   &
   &
  $\checkmark$ & 
   &
   &
   &
   &
  $\checkmark$ &
  $\checkmark$ &
  $\checkmark$ &
  $\checkmark$ &
   &
   &
   &
   &
  $\sim$ &
  &
  &
  $\checkmark$ &
  $\checkmark$ &
  & 
  $\sim$ & 
   & 
   & 
   &
   &
  $\checkmark$ &
  $\checkmark$ & 
   & 
  $\checkmark$ &
  $\checkmark$ &
  $\checkmark$ &
  $\sim$ &
  $\sim$ &
   &
  $\sim$ &
   &
  $\sim$ &
  $\checkmark$ &
   &
   &
  $\checkmark$ &
  $\checkmark$ &
  $\checkmark$ & 
  16(22) \\
  \midrule
REM &
  ${\checkmark}$ &
  ${\checkmark}$ &
  ${\checkmark}$ &
  ${\checkmark}$ &
  ${\checkmark}$ &
  ${\checkmark}$ &
  ${\checkmark}$ &
  ${\checkmark}$ &
    &
  $\checkmark$ &
  $\checkmark$ &
  $\checkmark$ &
  $\checkmark$ &
  ${\checkmark}$ &
  ${\checkmark}$ &
  ${\checkmark}$ &
  ${\checkmark}$ &
   $\checkmark$ & 
   $\checkmark$ & 
   $\checkmark$ &
  $\checkmark$ &
   $\checkmark$ &
  ${\checkmark}$ & 
  $\checkmark$  &   
  ${\checkmark}$ &  
  ${\checkmark}$ & 
  ${\checkmark}$ & 
  ${\checkmark}$ & 
  ${\checkmark}$ &
  $\checkmark$ &
  ${\checkmark}$ &  
  $\checkmark$ & 
  $\checkmark$ &
  $\checkmark$ &  
  $\checkmark$  & 
  ${\checkmark}$ &
  $\checkmark$  & 
  $\checkmark$ &
  ${\checkmark}$ &
  ${\checkmark}$ &
  $\checkmark$ &
  $\checkmark$ &
  $\checkmark$ &
  $\checkmark$ &
  $\checkmark$ &
  $\sim$ &
  42(43) \\
VEM &
  ${\checkmark}$ &
  ${\checkmark}$ &
  ${\checkmark}$ &
  ${\checkmark}$ &
  ${\checkmark}$ &
  ${\checkmark}$ &
  ${\checkmark}$ &
  ${\checkmark}$ &
  ${\checkmark}$ &
  ${\checkmark}$ &
  ${\checkmark}$ &
  ${\checkmark}$ &
  ${\checkmark}$ &
  ${\checkmark}$ &
  ${\checkmark}$ &
  ${\checkmark}$ &
  ${\checkmark}$ &
  ${\checkmark}$ &
  ${\checkmark}$ &
  ${\checkmark}$&
  ${\checkmark}$ &
  ${\checkmark}$ &
  $\checkmark$ & 
  ${\checkmark}$ & 
  ${\checkmark}$ &
  ${\checkmark}$ &
  ${\checkmark}$ &
  ${\checkmark}$ & 
  ${\checkmark}$ &
  ${\checkmark}$ &
  ${\checkmark}$  &  
  ${\checkmark}$  &  
  ${\checkmark}$ &
  ${\checkmark}$  &
  ${\checkmark}$ &
  ${\checkmark}$ &
  ${\checkmark}$ &
  ${\checkmark}$ &
  ${\checkmark}$ &
  ${\checkmark}$ &
  ${\checkmark}$ &
  $\checkmark$ & 
  $\checkmark$  &
  $\checkmark$  &
  $\checkmark$  &
  $\checkmark$  &
  44 \\
  \midrule

\end{tabular}}
\label{tab:supported games}
\end{table}

\subsection{OCAtari vs AtariARI}
\label{sec:comparingWithARI}

For their AtariARI framework,~\citet{Anand19AtariARI} disassembled the source code of various games to find the RAM location of the objects' properties. 
AtariARI thus provides information of where a specific information is encoded in the RAM. 
Providing only the RAM positions is however not enough to get a directly human-interpretable, object-centric description of the state. 
As shown in \autoref{fig:VEMvREM}, even when positions are encoded directly, offsets are applied to objects during the rendering phase, which the raw RAM information does not provide. 
Some games the information provided by AtariARI is thus incomplete or insufficient to play the game (\cf \autoref{tab:supported games} and \autoref{app:insufficent}). 
Our OCAtari framework makes use of intricate computations, such as deriving the \texttt{x} and \texttt{y} positions from grid anchors and offsets, looking up potential presence indicators (e.g. for objects that have been destroyed).
This ensures that RL agents genuinely acquire human understandable object-centric state descriptions, on which they can base their policies.
Finally, OCAtari is already covering \rebut{($28$)} more games than AtariARI, and we are continually adapting the rest of the game collection of ALE. 

\section{Related Work}
Atari games to benchmark deep RL agents has a well-established history. 
\citet{Mnih2015dqn} introduced the direct use of frames with DQN, tested on $7$ different games of ALE. 
In the following years, Atari games was repeatedly used as a test bed for various approaches, well-known ones being Rainbow~\citep{hessel2018rainbow}, Dreamer~\citep{Hafner2020dreamer}, MuZero~\citep{Schrittwieser2020muzero}, Agent57~\citep{Badia2020agent57} or GDI~\citep{Fan2021GDI}.
Although deep RL agent already achieve superhuman performance on Atari games, lots of challenges are left, like efficient exploration~\citep{Bellemare2016intrinsic, Ecoffet2019goexplore, ecoffet2021first}, efficiency~\citep{Kapturowski2022faster}, planning with sparse~\citep{Hafner2020dreamer, Schrittwieser2020muzero}, sample inefficiency, missgeneralization~\citep{ZambaldiRSBLBTR19, Mabelli2022MARL, stanic2022an}, etc. As underlined by~\citet{Toromanoff2019Saber}, these challenges can greatly benefit or might even require human like reasoning, and thus, object-centricity.

Other work have highlighted the need for augmented Atari benchmarks. \citet{Toromanoff2019Saber} and \citet{Fan2021AtariBenchmark} have both proposed to integrate many additional metrics to accurately measure performance, and \citet{MachadoBTVHB18} insisting on integrating the learning efficiency. This was tackled 
by~\citet{KaiserBMOCCEFKL20}, with their Atari $100k$ benchmark.
\citet{AitchisonSH23} have selected representative subsets of $5$ ALE environments, by looking at the performance variances of commonly used agents.
\citet{Shao2022POMDP} introduced a partial observable Atari benchmark, called Mask Atari, designed to test specifically POMDPs. 
These extensions can easily integrate OCAtari environments, as they can be swapped with ALE ones.
Many other object-centic representations learning methods, that tackle these challenges, have also been explored outside of RL~\citep{eslami2016air, kosiorek2018sqair, JiangL19NLRL, greff2019iodine, engelcke2020genesis, locatello2020slotattention, kipf2021SaVi, Elsayed2022endtoendoc, Singh2022dalle, singh2022ocvideos}. \citet{DittadiPVSWL22, Yoon2023Eval} look at objects properties' extractions, and generalization, required for downstream tasks, while \citet{lin2020space} and \citep{Delfosse2021MOC} already rely on ALE to evaluate representation learning.

Finally, several object-centric RL environments have been developed, such as VirtualHome~\citep{puig2018virtualhome}, AI2-THOR~\citep{ai2thor} or iGibson 2.0~\citep{Li21Gibson}. 
While these benchmarks excel in providing realistic 3D environments conducive to AI research, they introduce high-dimensional observations and emphasizes physical interactions, particularly suitable for robotics-oriented studies. 

\section{Discussion}
\label{sec:discussion}

Our OCAtari environments are suitable for training object detection and object-tracking methods, as well as developing new object-centric RL approaches. 
OCAtari offers an information bottleneck in the form of list of objects and their properties rather than exhaustive details per game.
While evaluating performance using the ALE is one of the most recognized benchmarks in RL, these evaluations are not without flaws, as explained by~\cite{agarwal2021deep}. 
The noisy scores do not linearly reflect the agents' learning ability. 
These games are also created to be played by humans and offer many shortcut learning possibilities \citep{Delfosse2024InterpretableCB}.
Directly evaluating the representations performance helps to understand and measure the quality of the learned internal representation and minimize other effects within the training, as proposed by~\citet{pmlr-v139-stooke21a}. 
The object-centricity offered by OCAtari also allows to provide extra information to the algorithms, such as additional reward signal based on objects properties or relations, as done by \citet{Wu23Read}. 
Finally, our provided repository includes many scripts for locating and analyzing RAM representation information, that can be adapted to other simulation benchmarks, and thus also equip them with object-centric states. 

\textbf{Societal and environmental impact.}
This work introduces a set of RL games. Such environments can be used for training object-tracking algorithms, which present potential ethical risks if misapplied. 
However, its main impact lies in advancing transparent object-centric RL methods, which can enhance the understanding of upcoming agents' decision-making processes and reduce misalignment issues~\citep{Friedrich2022ATF}. 
Improving transparency can also potentially help uncovering existing biases in learning algorithms with possible negative societal consequences~\citep{Schramowski2020MakingDN, Steinmann2023LearningTI}. 
OCAtari can also save resources while training RL policies.
We do not incorporate and have not found any personal or offensive content in our framework.

\textbf{Limitations.}
\label{sec:limits}
OCAtari extracts object-centric representations of ALE games. 
In most games, there are hardcoded static elements, which we did not consider as objects. 
For instance, no information about the mazes in Pacaman and MsPacman are encoded in the RAM. 
As such, we cannot extract this information, or only partially. We could in the future decide to hardcode suitable representations for it, but we have not found one yet.
However, this information being static, it could be learned by agents, but the integration of such information as input can help agents understand that \eg they cannot move through it. 
An interesting consideration here would be whether a combination of our two modes, object-centric states and frames, can be used to extract not only objects but also important information from the backgrounds. Using this additional information like the position of objects in MsPacman to run A* or similar path finding algorithms could also be an interesting way forward. 

\section{Conclusion}
Representing scenes in terms of objects and their relations is a crucial human ability that allows one to focus on the essentials when reasoning. While object-centric reinforcement learning and unsupervised detection algorithms are increasingly successful, we lack benchmarks and datasets to evaluate and compare such methods. 
OCAtari fills this gap and provides an easy-to-use diverse set of environments to develop and test object-centric learning methods on many games of ALE, by far the most commonly used RL benchmark. 
Overall, we hope that our work inspires other researchers to create object-centric approaches, allowing for more interpretable algorithms that humans can interact with and maybe learn from in the future. 
OCAtari will also permit AI practitioners to create novel challenges among the existing Atari games, usable on object-centric, deep or hybrid approaches.

\clearpage


\vskip 0.2in
\bibliography{bibliography}
\bibliographystyle{rlc}

\clearpage

\appendix
\clearpage

\section{ODA, an Object-centric Dataset for Atari.}
\label{app:oda}
OCAtari enables training policies using an object-centric approach to describe RL states for various Atari games. It can serve as a fast and dependable alternative to methods that discover objects. To compare object-centric agents to classic deep ones, it is necessary to train an object detection method and integrate it into the object-centric playing agent, \eg, as shown by~\citet{Delfosse2021MOC}. To train and compare the object detection methods, we introduce the \textbf{O}bject-centric \textbf{D}ataset for \textbf{A}tari (ODA), a preset selection of frames from the Atari games covered by OCAtari.
For each game, ODAs incorporates sequential states, where for each state, the $210\!\times\!160$ RGB frame is stored with the list of objects found by both VEM and REM procedure (otherwise the game sequence is discarded). The HUD elements are separated from the game objects. Every additional object information contained from the RAM is also saved. As trained agents with varying capabilities can expose different parts of the environment, especially in progressive games where agents must achieve a certain level of mastery to reveal new parts of the game, it is necessary to fix the agents that are used to capture these frames~\citep{Delfosse2021AdaptiveRA}. The frames are extracted using both a random and a trained DQN agent to cover numerous possible states within each game, that should incorporate states encountered by learning agents. In many games, \eg, Montezuma's Revenge or Kangaroo, such agents are not good enough to access every level of the game. However, as the level part is also stored in RAM, we let the agent start in different part of the game by manipulating the RAM. We choose to build our dataset out of 30\% of games from the random agent and 70\% of the games based on the DQN agent. All needed information, as well as the models used to generate ODA, are provided within the OCAtari repository.

\section{Details on object perception and its advantages}
\label{app:objects_advantages}
As described in our manuscript, decomposing the world in terms of objects incorporates many advantages, some of them are:

\textbf{Abstraction and Generalization} \\ Objects allow us to abstract and generalize information. By categorizing similar objects together, we can create concepts and classifications that help us make sense of a wide variety of individual instances.

\textbf{Cognitive Efficiency} \\ Our brains are more efficient at processing and remembering information when it's organized into meaningful chunks. Objects provide a natural way to group related information, making it easier for us to reason about complex situations.

\textbf{Predictive Reasoning} \\ Objects have properties and behaviors that can be predicted based on their past interactions and characteristics. This predictive reasoning is crucial for making informed decisions and anticipating outcomes.

\textbf{Cause and Effect} \\ Objects play a key role in understanding cause-and-effect relationships. By observing how objects interact and how changes in one object lead to changes in others, we can infer causal connections and predict future outcomes.

\textbf{Communication} \\ Objects provide a shared vocabulary that facilitates communication and understanding. When we refer to objects, we can convey complex ideas more efficiently than describing individual instances or specific situations..

\textbf{Logical Inference} \\ Objects provide a basis for logical reasoning. By identifying relationships between objects, we can deduce logical conclusions and make valid inferences.

\clearpage
\section{Details on OCAtari}
\begin{figure}[h!]
    \centering
    \includegraphics[width=0.55\linewidth]{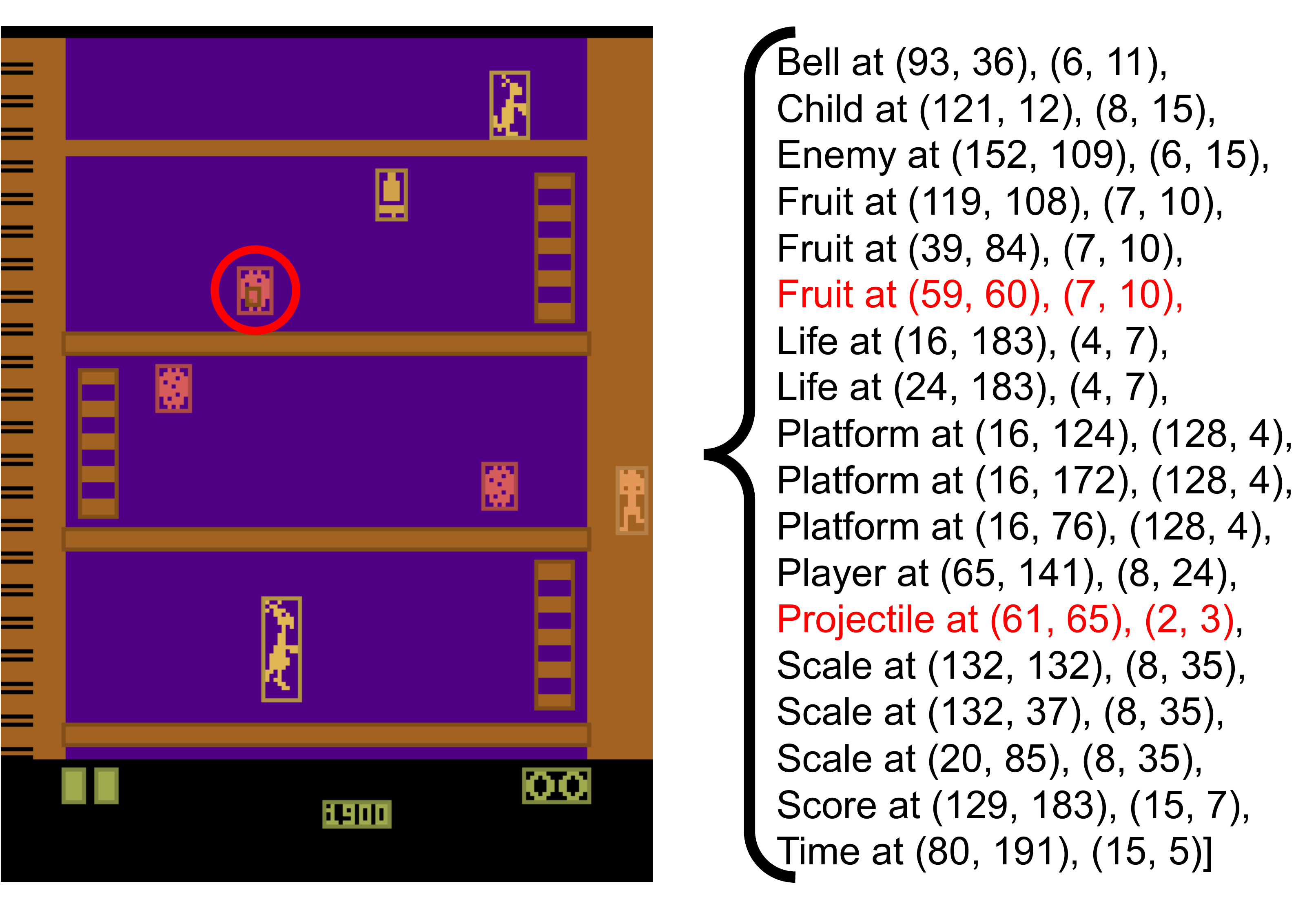}
    \caption{\textbf{OCAtari: The object-centric Atari benchmark.} OCAtari maintains a list of existing objects via processing the information from the RAM. Our framework enables training and evaluating object discovery methods and object-centric RL algorithms on the widely used Atari Learning Environments benchmark.}
    \label{fig:benchmark}
\end{figure}
\section{Reproducing our Results}
\begin{figure}[h]
    \centering
\includegraphics[width=0.8\textwidth]{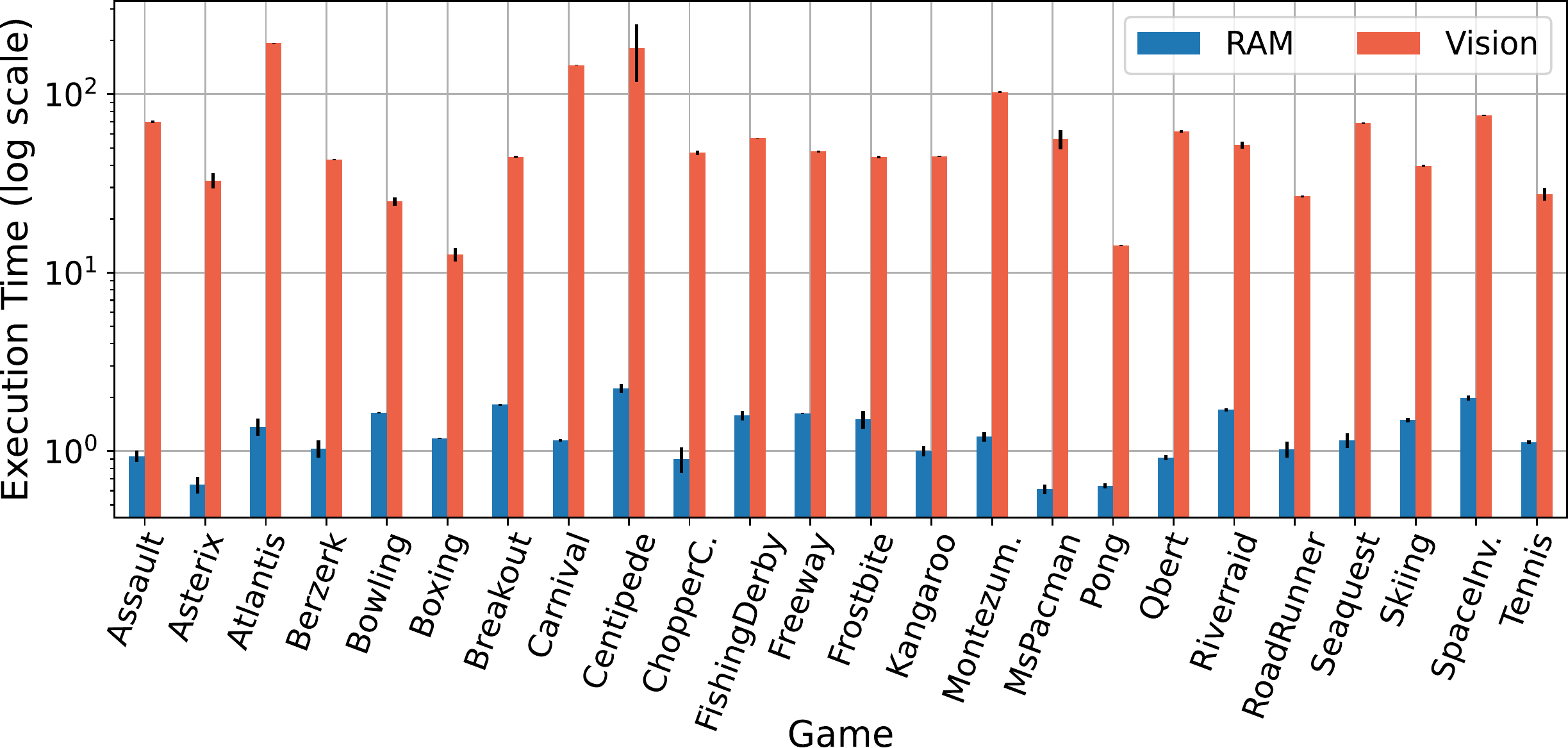}
\caption{\textbf{Using the RAM extraction procedures leads to 50$\!\boldsymbol{\times}$ faster environments.} The average time needed to perform $10^4$ steps in each OCAtari game, using RAM extraction (REM), and our vision extraction (VEM).}
\label{fig:speed}
\end{figure}

To reproduce our results, we included the option to run the experiments deterministically. For this purpose, a seed can be specified in the respective scripts. In our experiments, we used the seeds 0 and 42. 
All supported games can be found in \autoref{tab:supported games}. Since we are extending the environment permanently, you can also find all supported games in the ReadMe of our repository. To test if a game is supported, you can also use the scripts ``test\_game'' or ``test\_game\_both'' depending on if you want to test only one or both modes of OCAtari. \autoref{tab:f1} and all tables in section \ref{app:detailed_results} are generated by the script ``get\_metrics''. To reproduce and measure the time needed for evaluation, see \autoref{fig:speed}, the script ``test\_speed'' was used. For further information, we recommend checking the documentation of OCAtari under \anonymizedlink{\url{https://oc-atari.readthedocs.io/}}.

As mentioned before, we are using the models obtained from~\citet{gogianu2022agents}\footnote{\url{https://github.com/floringogianu/atari-agents}}. However for the games recently added to gymnasium, i.e. Pacman and Donkeykong, we needed to train our own agents. For this purpose we were using the \textit{cleanRL} framework by \citet{huang2022cleanrl}\footnote{\url{https://github.com/vwxyzjn/cleanrl}}.

\clearpage
\section{Experimental details}
\label{app:exp_details}

\begin{wraptable}{r}{6cm}
\centering
\vspace{-0.5cm}
\begin{tabular}{lr}
Actors $N$ & 8 \\
Minibatch size & $32 * 8$ \\
Horizon $T$ & 128 \\
Num. epochs $K$ & $3$ \\
Adam stepsize & $2.5 * 10^{-4} * \alpha$ \\
Discount $\gamma$ & $0.99$ \\
GAE parameter $\lambda$ & $0.95$ \\
Clipping parameter $\epsilon$ & $0.1 * \alpha$ \\
VF coefficient $c_1$ & $1$ \\
Entropy coefficient $c_2$ & $0.01$ \\
MLP architecture & $2\times64$ \\
MLP activation fn. & Tanh
\end{tabular}
\caption{PPO Hyperparameter Values. $\alpha$ linearly increases from $0$ to $1$ over the course of training.}
\label{app:tab:ppo_hp}
\end{wraptable} 

In our case, all experiments on object extraction and dataset generation were run on a machine with an AMD Ryzen 7 processor, 64GB of RAM, and no dedicated GPU. The dataset generation script takes approximately $3$ minutes for one game. We use the same hyperparameters as the~\citet{SchulmanWDRK17PPO} PPO agents that learned to master the games. Hyperparameter values for Atari environments are derived from the original PPO paper. The same applies to the definitions and values of VF coefficient $c_1$ and entropy coefficient $c_2$. The PPO implementation used and respective MLP hyperparameters are based on stable-baselines3~\citep{stable-baselines3}. Deep agents have the same hyperparameter values as OCAtari agents but use 'CnnPolicy' in stable-baselines3 for the policy architecture and frame stacking of 4. The Atari environment version used in gymnasium is v4 \& v5. This version defines a deterministic skipping of 5 frames per action taken and sets the probability to repeat the last action taken to 0.25. This is aligned with recommended best practices by~\citet{MachadoBTVHB18}. \rebut{We also used the \textit{Deterministic} and \textit{NoFrameskip} features of gymnasium when necessary to make our experiments easier to reproduce.}
A list of all hyperparameter values used is provided in \autoref{app:tab:ppo_hp}. 

\section{Generating Datasets}
With OCAtari it is possible to create object-centric datasets for all supported games. The dataset consists primarily of a csv file. In addition to a sequential \textbf{index}, based on the game number and state number, this file contains the respective image as a list of pixels, called \textbf{OBS}. An image in the form of a png file is also stored separately. Furthermore, the csv file contains a list of all HUD elements that could be extracted from the RAM, called \textbf{HUD}, as well as a list of all objects that were read from the RAM, called \textbf{RAM}. Finally, we provide a list of all elements that could be generated using the vision mode, called \textbf{VIS}. An example is given in \autoref{app:tab:dataset}.

The generation of the dataset can also be made reproducible by setting a seed. For our tests, we used the seeds 0 and 42. More information at \anonymizedlink{\url{https://github.com/k4ntz/OC_Atari/tree/master/dataset_generation}}.

\begin{table}[h]
    \centering
    \caption{An example how an object-centric dataset for Atari looks like after generation.}
    \resizebox{\textwidth}{!}{
    \begin{tabular}{ccccc} \toprule
        Index & OBS & HUD & RAM & VIS  \\ \midrule
        00001\_00001 & [[0,0,0]...[255,255,255]] & score at (x,y)(width, height),... & ball at (x,y)(width, height),... & ball at (x,y)(width, height),.... \\
        00001\_00002 & [[0,0,0]...[255,255,255]] & score at (x,y)(width, height),... & ball at (x,y)(width, height),... & ball at (x,y)(width, height),.... \\
        00001\_00003 & [[0,0,0]...[255,255,255]] & score at (x,y)(width, height),... & ball at (x,y)(width, height),... & ball at (x,y)(width, height),.... \\
        ... & & & & \\ 
        00008\_00678 & [[0,0,0]...[255,255,255]] & score at (x,y)(width, height),... & ball at (x,y)(width, height),... & ball at (x,y)(width, height),.... \\ \bottomrule
    \end{tabular}
    }
    \label{app:tab:dataset}
\end{table}

\clearpage
\section{Detailed Per Object Category results on each game.}
\label{app:detailed_results}
In this section, we provide descriptions of each covered game (obtained from \url{https://gymnasium.farama.org/environments/atari/}) with example frames. For a more detailed documentation, see the game's respective AtariAge manual page\footnote{\url{https://atariage.com/system_items.php?SystemID=2600&itemTypeID=MANUAL}}. We also share detailed statistics on the object detection capacities of OCAtari for every class of objects detected in each game. 

\renewcommand{\arraystretch}{1.1}

\newcommand{\bl}[1]{{\texttt{\color{blue}{#1}}}}

\begin{table}[h!]
\centering
\caption{A more detailed version of \autoref{tab:f1}. Precision, Recall, F1-scores of REM, and intersection over union (IOU) metrics. Frames are obtained using random, DQN and C51 (if available) agents.}
\resizebox{\textwidth}{!}{

\begin{tabular}{l|rrrr|rrrr|rrrrr}
\toprule
 & \multicolumn{4}{c|}{Random} & \multicolumn{4}{c|}{DQN} & \multicolumn{4}{c}{C51} \\
 & precision & recall & f-score & iou & precision & recall & f-score & iou & precision & recall & f-score & iou \\
\midrule
Alien & {\cellcolor[HTML]{FCFEBA}} \color[HTML]{000000} 51.4 & {\cellcolor[HTML]{545CA8}} \color[HTML]{F1F1F1} 97.3 & {\cellcolor[HTML]{BAE3A1}} \color[HTML]{000000} 67.3 & {\cellcolor[HTML]{555AA7}} \color[HTML]{F1F1F1} 97.7 & {\cellcolor[HTML]{FCFEBA}} \color[HTML]{000000} 51.2 & {\cellcolor[HTML]{525FA9}} \color[HTML]{F1F1F1} 97.2 & {\cellcolor[HTML]{BCE4A0}} \color[HTML]{000000} 67.1 & {\cellcolor[HTML]{545CA8}} \color[HTML]{F1F1F1} 97.4 & {\cellcolor[HTML]{000000}} \color[HTML]{F1F1F1} {\cellcolor{gray}} N/A & {\cellcolor[HTML]{000000}} \color[HTML]{F1F1F1} {\cellcolor{gray}} N/A & {\cellcolor[HTML]{000000}} \color[HTML]{F1F1F1} {\cellcolor{gray}} N/A & {\cellcolor[HTML]{000000}} \color[HTML]{F1F1F1} {\cellcolor{gray}} N/A \\
Amidar & {\cellcolor[HTML]{81CDA5}} \color[HTML]{000000} 75.8 & {\cellcolor[HTML]{5E4FA2}} \color[HTML]{F1F1F1} 99.9 & {\cellcolor[HTML]{459EB4}} \color[HTML]{F1F1F1} 86.2 & {\cellcolor[HTML]{525FA9}} \color[HTML]{F1F1F1} 97.0 & {\cellcolor[HTML]{459EB4}} \color[HTML]{F1F1F1} 86.3 & {\cellcolor[HTML]{5E4FA2}} \color[HTML]{F1F1F1} 99.9 & {\cellcolor[HTML]{3F77B5}} \color[HTML]{F1F1F1} 92.6 & {\cellcolor[HTML]{3D79B6}} \color[HTML]{F1F1F1} 92.4 & {\cellcolor[HTML]{000000}} \color[HTML]{F1F1F1} {\cellcolor{gray}} N/A & {\cellcolor[HTML]{000000}} \color[HTML]{F1F1F1} {\cellcolor{gray}} N/A & {\cellcolor[HTML]{000000}} \color[HTML]{F1F1F1} {\cellcolor{gray}} N/A & {\cellcolor[HTML]{000000}} \color[HTML]{F1F1F1} {\cellcolor{gray}} N/A \\
Assault & {\cellcolor[HTML]{4B68AE}} \color[HTML]{F1F1F1} 95.4 & {\cellcolor[HTML]{466EB1}} \color[HTML]{F1F1F1} 94.2 & {\cellcolor[HTML]{486CB0}} \color[HTML]{F1F1F1} 94.8 & {\cellcolor[HTML]{496AAF}} \color[HTML]{F1F1F1} 95.3 & {\cellcolor[HTML]{525FA9}} \color[HTML]{F1F1F1} 97.1 & {\cellcolor[HTML]{4273B3}} \color[HTML]{F1F1F1} 93.6 & {\cellcolor[HTML]{496AAF}} \color[HTML]{F1F1F1} 95.3 & {\cellcolor[HTML]{4471B2}} \color[HTML]{F1F1F1} 93.8 & {\cellcolor[HTML]{000000}} \color[HTML]{F1F1F1} {\cellcolor{gray}} N/A & {\cellcolor[HTML]{000000}} \color[HTML]{F1F1F1} {\cellcolor{gray}} N/A & {\cellcolor[HTML]{000000}} \color[HTML]{F1F1F1} {\cellcolor{gray}} N/A & {\cellcolor[HTML]{000000}} \color[HTML]{F1F1F1} {\cellcolor{gray}} N/A \\
Asterix & {\cellcolor[HTML]{4175B4}} \color[HTML]{F1F1F1} 93.1 & {\cellcolor[HTML]{5E4FA2}} \color[HTML]{F1F1F1} 99.8 & {\cellcolor[HTML]{4E63AC}} \color[HTML]{F1F1F1} 96.3 & {\cellcolor[HTML]{4D65AD}} \color[HTML]{F1F1F1} 96.0 & {\cellcolor[HTML]{496AAF}} \color[HTML]{F1F1F1} 95.0 & {\cellcolor[HTML]{5C51A3}} \color[HTML]{F1F1F1} 99.6 & {\cellcolor[HTML]{525FA9}} \color[HTML]{F1F1F1} 97.2 & {\cellcolor[HTML]{4E63AC}} \color[HTML]{F1F1F1} 96.1 & {\cellcolor[HTML]{486CB0}} \color[HTML]{F1F1F1} 94.8 & {\cellcolor[HTML]{5E4FA2}} \color[HTML]{F1F1F1} 99.8 & {\cellcolor[HTML]{525FA9}} \color[HTML]{F1F1F1} 97.2 & {\cellcolor[HTML]{4E63AC}} \color[HTML]{F1F1F1} 96.2 \\
Atlantis & {\cellcolor[HTML]{4E63AC}} \color[HTML]{F1F1F1} 96.3 & {\cellcolor[HTML]{486CB0}} \color[HTML]{F1F1F1} 94.6 & {\cellcolor[HTML]{4B68AE}} \color[HTML]{F1F1F1} 95.5 & {\cellcolor[HTML]{4D65AD}} \color[HTML]{F1F1F1} 95.8 & {\cellcolor[HTML]{5061AA}} \color[HTML]{F1F1F1} 96.6 & {\cellcolor[HTML]{486CB0}} \color[HTML]{F1F1F1} 94.7 & {\cellcolor[HTML]{4B68AE}} \color[HTML]{F1F1F1} 95.7 & {\cellcolor[HTML]{496AAF}} \color[HTML]{F1F1F1} 95.0 & {\cellcolor[HTML]{000000}} \color[HTML]{F1F1F1} {\cellcolor{gray}} N/A & {\cellcolor[HTML]{000000}} \color[HTML]{F1F1F1} {\cellcolor{gray}} N/A & {\cellcolor[HTML]{000000}} \color[HTML]{F1F1F1} {\cellcolor{gray}} N/A & {\cellcolor[HTML]{000000}} \color[HTML]{F1F1F1} {\cellcolor{gray}} N/A \\
BankHeist & {\cellcolor[HTML]{3B92B9}} \color[HTML]{F1F1F1} 87.9 & {\cellcolor[HTML]{4D65AD}} \color[HTML]{F1F1F1} 95.8 & {\cellcolor[HTML]{3A7EB8}} \color[HTML]{F1F1F1} 91.7 & {\cellcolor[HTML]{3F97B7}} \color[HTML]{F1F1F1} 87.3 & {\cellcolor[HTML]{4E63AC}} \color[HTML]{F1F1F1} 96.2 & {\cellcolor[HTML]{4E63AC}} \color[HTML]{F1F1F1} 96.2 & {\cellcolor[HTML]{4E63AC}} \color[HTML]{F1F1F1} 96.2 & {\cellcolor[HTML]{486CB0}} \color[HTML]{F1F1F1} 94.7 & {\cellcolor[HTML]{000000}} \color[HTML]{F1F1F1} {\cellcolor{gray}} N/A & {\cellcolor[HTML]{000000}} \color[HTML]{F1F1F1} {\cellcolor{gray}} N/A & {\cellcolor[HTML]{000000}} \color[HTML]{F1F1F1} {\cellcolor{gray}} N/A & {\cellcolor[HTML]{000000}} \color[HTML]{F1F1F1} {\cellcolor{gray}} N/A \\
BattleZone & {\cellcolor[HTML]{60BBA8}} \color[HTML]{000000} 81.1 & {\cellcolor[HTML]{F1F9A9}} \color[HTML]{000000} 55.7 & {\cellcolor[HTML]{C3E79F}} \color[HTML]{000000} 66.0 & {\cellcolor[HTML]{496AAF}} \color[HTML]{F1F1F1} 95.1 & {\cellcolor[HTML]{5CB7AA}} \color[HTML]{F1F1F1} 81.8 & {\cellcolor[HTML]{FBFDB8}} \color[HTML]{000000} 51.8 & {\cellcolor[HTML]{D1ED9C}} \color[HTML]{000000} 63.4 & {\cellcolor[HTML]{4273B3}} \color[HTML]{F1F1F1} 93.5 & {\cellcolor[HTML]{000000}} \color[HTML]{F1F1F1} {\cellcolor{gray}} N/A & {\cellcolor[HTML]{000000}} \color[HTML]{F1F1F1} {\cellcolor{gray}} N/A & {\cellcolor[HTML]{000000}} \color[HTML]{F1F1F1} {\cellcolor{gray}} N/A & {\cellcolor[HTML]{000000}} \color[HTML]{F1F1F1} {\cellcolor{gray}} N/A \\
Berzerk & {\cellcolor[HTML]{4471B2}} \color[HTML]{F1F1F1} 94.1 & {\cellcolor[HTML]{496AAF}} \color[HTML]{F1F1F1} 95.2 & {\cellcolor[HTML]{486CB0}} \color[HTML]{F1F1F1} 94.6 & {\cellcolor[HTML]{71C6A5}} \color[HTML]{000000} 78.4 & {\cellcolor[HTML]{466EB1}} \color[HTML]{F1F1F1} 94.3 & {\cellcolor[HTML]{5061AA}} \color[HTML]{F1F1F1} 96.5 & {\cellcolor[HTML]{4B68AE}} \color[HTML]{F1F1F1} 95.4 & {\cellcolor[HTML]{76C8A5}} \color[HTML]{000000} 77.4 & {\cellcolor[HTML]{000000}} \color[HTML]{F1F1F1} {\cellcolor{gray}} N/A & {\cellcolor[HTML]{000000}} \color[HTML]{F1F1F1} {\cellcolor{gray}} N/A & {\cellcolor[HTML]{000000}} \color[HTML]{F1F1F1} {\cellcolor{gray}} N/A & {\cellcolor[HTML]{000000}} \color[HTML]{F1F1F1} {\cellcolor{gray}} N/A \\
Bowling & {\cellcolor[HTML]{5C51A3}} \color[HTML]{F1F1F1} 99.5 & {\cellcolor[HTML]{5B53A4}} \color[HTML]{F1F1F1} 99.2 & {\cellcolor[HTML]{5C51A3}} \color[HTML]{F1F1F1} 99.3 & {\cellcolor[HTML]{5C51A3}} \color[HTML]{F1F1F1} 99.6 & {\cellcolor[HTML]{5B53A4}} \color[HTML]{F1F1F1} 99.2 & {\cellcolor[HTML]{5956A5}} \color[HTML]{F1F1F1} 98.8 & {\cellcolor[HTML]{5B53A4}} \color[HTML]{F1F1F1} 99.0 & {\cellcolor[HTML]{5C51A3}} \color[HTML]{F1F1F1} 99.4 & {\cellcolor[HTML]{5C51A3}} \color[HTML]{F1F1F1} 99.4 & {\cellcolor[HTML]{5B53A4}} \color[HTML]{F1F1F1} 99.1 & {\cellcolor[HTML]{5C51A3}} \color[HTML]{F1F1F1} 99.3 & {\cellcolor[HTML]{5C51A3}} \color[HTML]{F1F1F1} 99.5 \\
Boxing & {\cellcolor[HTML]{5061AA}} \color[HTML]{F1F1F1} 96.5 & {\cellcolor[HTML]{4EA7B0}} \color[HTML]{F1F1F1} 84.5 & {\cellcolor[HTML]{3387BC}} \color[HTML]{F1F1F1} 90.1 & {\cellcolor[HTML]{4273B3}} \color[HTML]{F1F1F1} 93.5 & {\cellcolor[HTML]{4E63AC}} \color[HTML]{F1F1F1} 96.1 & {\cellcolor[HTML]{4EA7B0}} \color[HTML]{F1F1F1} 84.5 & {\cellcolor[HTML]{3387BC}} \color[HTML]{F1F1F1} 89.9 & {\cellcolor[HTML]{4273B3}} \color[HTML]{F1F1F1} 93.4 & {\cellcolor[HTML]{5061AA}} \color[HTML]{F1F1F1} 96.8 & {\cellcolor[HTML]{47A0B3}} \color[HTML]{F1F1F1} 85.6 & {\cellcolor[HTML]{3682BA}} \color[HTML]{F1F1F1} 90.9 & {\cellcolor[HTML]{4471B2}} \color[HTML]{F1F1F1} 94.1 \\
Breakout & {\cellcolor[HTML]{5C51A3}} \color[HTML]{F1F1F1} 99.5 & {\cellcolor[HTML]{5E4FA2}} \color[HTML]{F1F1F1} 100 & {\cellcolor[HTML]{5E4FA2}} \color[HTML]{F1F1F1} 99.7 & {\cellcolor[HTML]{5E4FA2}} \color[HTML]{F1F1F1} 100 & {\cellcolor[HTML]{5C51A3}} \color[HTML]{F1F1F1} 99.5 & {\cellcolor[HTML]{5E4FA2}} \color[HTML]{F1F1F1} 100 & {\cellcolor[HTML]{5E4FA2}} \color[HTML]{F1F1F1} 99.7 & {\cellcolor[HTML]{5E4FA2}} \color[HTML]{F1F1F1} 100 & {\cellcolor[HTML]{5E4FA2}} \color[HTML]{F1F1F1} 100 & {\cellcolor[HTML]{5E4FA2}} \color[HTML]{F1F1F1} 100 & {\cellcolor[HTML]{5E4FA2}} \color[HTML]{F1F1F1} 100 & {\cellcolor[HTML]{5E4FA2}} \color[HTML]{F1F1F1} 100 \\
Carnival & {\cellcolor[HTML]{4175B4}} \color[HTML]{F1F1F1} 93.2 & {\cellcolor[HTML]{466EB1}} \color[HTML]{F1F1F1} 94.2 & {\cellcolor[HTML]{4273B3}} \color[HTML]{F1F1F1} 93.7 & {\cellcolor[HTML]{3682BA}} \color[HTML]{F1F1F1} 90.7 & {\cellcolor[HTML]{486CB0}} \color[HTML]{F1F1F1} 94.6 & {\cellcolor[HTML]{4E63AC}} \color[HTML]{F1F1F1} 96.4 & {\cellcolor[HTML]{4B68AE}} \color[HTML]{F1F1F1} 95.5 & {\cellcolor[HTML]{3A7EB8}} \color[HTML]{F1F1F1} 91.5 & {\cellcolor[HTML]{000000}} \color[HTML]{F1F1F1} {\cellcolor{gray}} N/A & {\cellcolor[HTML]{000000}} \color[HTML]{F1F1F1} {\cellcolor{gray}} N/A & {\cellcolor[HTML]{000000}} \color[HTML]{F1F1F1} {\cellcolor{gray}} N/A & {\cellcolor[HTML]{000000}} \color[HTML]{F1F1F1} {\cellcolor{gray}} N/A \\
Centipede & {\cellcolor[HTML]{4B68AE}} \color[HTML]{F1F1F1} 95.7 & {\cellcolor[HTML]{525FA9}} \color[HTML]{F1F1F1} 97.0 & {\cellcolor[HTML]{4E63AC}} \color[HTML]{F1F1F1} 96.3 & {\cellcolor[HTML]{496AAF}} \color[HTML]{F1F1F1} 95.1 & {\cellcolor[HTML]{4D65AD}} \color[HTML]{F1F1F1} 95.9 & {\cellcolor[HTML]{525FA9}} \color[HTML]{F1F1F1} 97.2 & {\cellcolor[HTML]{5061AA}} \color[HTML]{F1F1F1} 96.6 & {\cellcolor[HTML]{4D65AD}} \color[HTML]{F1F1F1} 96.0 & {\cellcolor[HTML]{000000}} \color[HTML]{F1F1F1} {\cellcolor{gray}} N/A & {\cellcolor[HTML]{000000}} \color[HTML]{F1F1F1} {\cellcolor{gray}} N/A & {\cellcolor[HTML]{000000}} \color[HTML]{F1F1F1} {\cellcolor{gray}} N/A & {\cellcolor[HTML]{000000}} \color[HTML]{F1F1F1} {\cellcolor{gray}} N/A \\
ChopperComma. & {\cellcolor[HTML]{358BBC}} \color[HTML]{F1F1F1} 89.2 & {\cellcolor[HTML]{358BBC}} \color[HTML]{F1F1F1} 89.4 & {\cellcolor[HTML]{358BBC}} \color[HTML]{F1F1F1} 89.3 & {\cellcolor[HTML]{74C7A5}} \color[HTML]{000000} 78.1 & {\cellcolor[HTML]{71C6A5}} \color[HTML]{000000} 78.3 & {\cellcolor[HTML]{69C3A5}} \color[HTML]{000000} 79.5 & {\cellcolor[HTML]{6EC5A5}} \color[HTML]{000000} 78.9 & {\cellcolor[HTML]{439BB5}} \color[HTML]{F1F1F1} 86.7 & {\cellcolor[HTML]{9CD7A4}} \color[HTML]{000000} 72.1 & {\cellcolor[HTML]{84CEA5}} \color[HTML]{000000} 75.6 & {\cellcolor[HTML]{91D3A4}} \color[HTML]{000000} 73.8 & {\cellcolor[HTML]{4273B3}} \color[HTML]{F1F1F1} 93.5 \\
CrazyClimber & {\cellcolor[HTML]{545CA8}} \color[HTML]{F1F1F1} 97.6 & {\cellcolor[HTML]{4D65AD}} \color[HTML]{F1F1F1} 96.0 & {\cellcolor[HTML]{5061AA}} \color[HTML]{F1F1F1} 96.8 & {\cellcolor[HTML]{545CA8}} \color[HTML]{F1F1F1} 97.6 & {\cellcolor[HTML]{555AA7}} \color[HTML]{F1F1F1} 97.9 & {\cellcolor[HTML]{486CB0}} \color[HTML]{F1F1F1} 94.8 & {\cellcolor[HTML]{4E63AC}} \color[HTML]{F1F1F1} 96.3 & {\cellcolor[HTML]{5061AA}} \color[HTML]{F1F1F1} 96.7 & {\cellcolor[HTML]{000000}} \color[HTML]{F1F1F1} {\cellcolor{gray}} N/A & {\cellcolor[HTML]{000000}} \color[HTML]{F1F1F1} {\cellcolor{gray}} N/A & {\cellcolor[HTML]{000000}} \color[HTML]{F1F1F1} {\cellcolor{gray}} N/A & {\cellcolor[HTML]{000000}} \color[HTML]{F1F1F1} {\cellcolor{gray}} N/A \\
DemonAttack & {\cellcolor[HTML]{D6EE9B}} \color[HTML]{000000} 62.6 & {\cellcolor[HTML]{6EC5A5}} \color[HTML]{000000} 78.6 & {\cellcolor[HTML]{ACDDA4}} \color[HTML]{000000} 69.7 & {\cellcolor[HTML]{66C2A5}} \color[HTML]{000000} 79.9 & {\cellcolor[HTML]{E7F59A}} \color[HTML]{000000} 59.5 & {\cellcolor[HTML]{76C8A5}} \color[HTML]{000000} 77.6 & {\cellcolor[HTML]{BAE3A1}} \color[HTML]{000000} 67.3 & {\cellcolor[HTML]{50A9AF}} \color[HTML]{F1F1F1} 84.1 & {\cellcolor[HTML]{000000}} \color[HTML]{F1F1F1} {\cellcolor{gray}} N/A & {\cellcolor[HTML]{000000}} \color[HTML]{F1F1F1} {\cellcolor{gray}} N/A & {\cellcolor[HTML]{000000}} \color[HTML]{F1F1F1} {\cellcolor{gray}} N/A & {\cellcolor[HTML]{000000}} \color[HTML]{F1F1F1} {\cellcolor{gray}} N/A \\
DonkeyKong. & {\cellcolor[HTML]{4D65AD}} \color[HTML]{F1F1F1} 96.0 & {\cellcolor[HTML]{5956A5}} \color[HTML]{F1F1F1} 98.6 & {\cellcolor[HTML]{545CA8}} \color[HTML]{F1F1F1} 97.3 & {\cellcolor[HTML]{5B53A4}} \color[HTML]{F1F1F1} 99.1 & {\cellcolor[HTML]{5956A5}} \color[HTML]{F1F1F1} 98.5 & {\cellcolor[HTML]{5956A5}} \color[HTML]{F1F1F1} 98.7 & {\cellcolor[HTML]{5956A5}} \color[HTML]{F1F1F1} 98.6 & {\cellcolor[HTML]{5B53A4}} \color[HTML]{F1F1F1} 99.1 & {\cellcolor[HTML]{5956A5}} \color[HTML]{F1F1F1} 98.7 & {\cellcolor[HTML]{5956A5}} \color[HTML]{F1F1F1} 98.5 & {\cellcolor[HTML]{5956A5}} \color[HTML]{F1F1F1} 98.6 & {\cellcolor[HTML]{5B53A4}} \color[HTML]{F1F1F1} 99.1 \\
FishingDerby & {\cellcolor[HTML]{358BBC}} \color[HTML]{F1F1F1} 89.2 & {\cellcolor[HTML]{47A0B3}} \color[HTML]{F1F1F1} 85.6 & {\cellcolor[HTML]{3F97B7}} \color[HTML]{F1F1F1} 87.3 & {\cellcolor[HTML]{86CFA5}} \color[HTML]{000000} 75.2 & {\cellcolor[HTML]{378EBB}} \color[HTML]{F1F1F1} 88.8 & {\cellcolor[HTML]{4EA7B0}} \color[HTML]{F1F1F1} 84.6 & {\cellcolor[HTML]{439BB5}} \color[HTML]{F1F1F1} 86.6 & {\cellcolor[HTML]{91D3A4}} \color[HTML]{000000} 73.6 & {\cellcolor[HTML]{56B0AD}} \color[HTML]{F1F1F1} 83.2 & {\cellcolor[HTML]{74C7A5}} \color[HTML]{000000} 77.9 & {\cellcolor[HTML]{62BDA7}} \color[HTML]{000000} 80.5 & {\cellcolor[HTML]{84CEA5}} \color[HTML]{000000} 75.7 \\
Freeway & {\cellcolor[HTML]{5956A5}} \color[HTML]{F1F1F1} 98.7 & {\cellcolor[HTML]{3F97B7}} \color[HTML]{F1F1F1} 87.3 & {\cellcolor[HTML]{3F77B5}} \color[HTML]{F1F1F1} 92.6 & {\cellcolor[HTML]{3387BC}} \color[HTML]{F1F1F1} 90.2 & {\cellcolor[HTML]{5956A5}} \color[HTML]{F1F1F1} 98.6 & {\cellcolor[HTML]{3F97B7}} \color[HTML]{F1F1F1} 87.3 & {\cellcolor[HTML]{3F77B5}} \color[HTML]{F1F1F1} 92.6 & {\cellcolor[HTML]{3387BC}} \color[HTML]{F1F1F1} 90.2 & {\cellcolor[HTML]{5061AA}} \color[HTML]{F1F1F1} 96.5 & {\cellcolor[HTML]{3F97B7}} \color[HTML]{F1F1F1} 87.2 & {\cellcolor[HTML]{3A7EB8}} \color[HTML]{F1F1F1} 91.6 & {\cellcolor[HTML]{3B92B9}} \color[HTML]{F1F1F1} 87.9 \\
Frostbite & {\cellcolor[HTML]{545CA8}} \color[HTML]{F1F1F1} 97.6 & {\cellcolor[HTML]{5C51A3}} \color[HTML]{F1F1F1} 99.5 & {\cellcolor[HTML]{5956A5}} \color[HTML]{F1F1F1} 98.6 & {\cellcolor[HTML]{3F77B5}} \color[HTML]{F1F1F1} 92.7 & {\cellcolor[HTML]{3D95B8}} \color[HTML]{F1F1F1} 87.5 & {\cellcolor[HTML]{545CA8}} \color[HTML]{F1F1F1} 97.5 & {\cellcolor[HTML]{3D79B6}} \color[HTML]{F1F1F1} 92.2 & {\cellcolor[HTML]{4199B6}} \color[HTML]{F1F1F1} 87.1 & {\cellcolor[HTML]{49A2B2}} \color[HTML]{F1F1F1} 85.5 & {\cellcolor[HTML]{525FA9}} \color[HTML]{F1F1F1} 97.1 & {\cellcolor[HTML]{3682BA}} \color[HTML]{F1F1F1} 90.9 & {\cellcolor[HTML]{49A2B2}} \color[HTML]{F1F1F1} 85.4 \\
Gopher & {\cellcolor[HTML]{5758A6}} \color[HTML]{F1F1F1} 98.3 & {\cellcolor[HTML]{FFFAB6}} \color[HTML]{000000} 48.2 & {\cellcolor[HTML]{CAEA9E}} \color[HTML]{000000} 64.7 & {\cellcolor[HTML]{71C6A5}} \color[HTML]{000000} 78.4 & {\cellcolor[HTML]{5758A6}} \color[HTML]{F1F1F1} 98.3 & {\cellcolor[HTML]{FFF7B2}} \color[HTML]{000000} 47.6 & {\cellcolor[HTML]{CDEB9D}} \color[HTML]{000000} 64.1 & {\cellcolor[HTML]{50A9AF}} \color[HTML]{F1F1F1} 84.0 & {\cellcolor[HTML]{000000}} \color[HTML]{F1F1F1} {\cellcolor{gray}} N/A & {\cellcolor[HTML]{000000}} \color[HTML]{F1F1F1} {\cellcolor{gray}} N/A & {\cellcolor[HTML]{000000}} \color[HTML]{F1F1F1} {\cellcolor{gray}} N/A & {\cellcolor[HTML]{000000}} \color[HTML]{F1F1F1} {\cellcolor{gray}} N/A \\
Hero & {\cellcolor[HTML]{3D79B6}} \color[HTML]{F1F1F1} 92.4 & {\cellcolor[HTML]{486CB0}} \color[HTML]{F1F1F1} 94.6 & {\cellcolor[HTML]{4273B3}} \color[HTML]{F1F1F1} 93.5 & {\cellcolor[HTML]{3B92B9}} \color[HTML]{F1F1F1} 88.2 & {\cellcolor[HTML]{6BC4A5}} \color[HTML]{000000} 79.0 & {\cellcolor[HTML]{3990BA}} \color[HTML]{F1F1F1} 88.4 & {\cellcolor[HTML]{54AEAD}} \color[HTML]{F1F1F1} 83.4 & {\cellcolor[HTML]{459EB4}} \color[HTML]{F1F1F1} 86.3 & {\cellcolor[HTML]{62BDA7}} \color[HTML]{000000} 80.8 & {\cellcolor[HTML]{3A7EB8}} \color[HTML]{F1F1F1} 91.7 & {\cellcolor[HTML]{47A0B3}} \color[HTML]{F1F1F1} 85.9 & {\cellcolor[HTML]{439BB5}} \color[HTML]{F1F1F1} 86.7 \\
IceHockey & {\cellcolor[HTML]{358BBC}} \color[HTML]{F1F1F1} 89.2 & {\cellcolor[HTML]{5C51A3}} \color[HTML]{F1F1F1} 99.6 & {\cellcolor[HTML]{4471B2}} \color[HTML]{F1F1F1} 94.1 & {\cellcolor[HTML]{C1E6A0}} \color[HTML]{000000} 66.2 & {\cellcolor[HTML]{3D79B6}} \color[HTML]{F1F1F1} 92.4 & {\cellcolor[HTML]{5E4FA2}} \color[HTML]{F1F1F1} 99.7 & {\cellcolor[HTML]{4D65AD}} \color[HTML]{F1F1F1} 95.9 & {\cellcolor[HTML]{C1E6A0}} \color[HTML]{000000} 66.3 & {\cellcolor[HTML]{000000}} \color[HTML]{F1F1F1} {\cellcolor{gray}} N/A & {\cellcolor[HTML]{000000}} \color[HTML]{F1F1F1} {\cellcolor{gray}} N/A & {\cellcolor[HTML]{000000}} \color[HTML]{F1F1F1} {\cellcolor{gray}} N/A & {\cellcolor[HTML]{000000}} \color[HTML]{F1F1F1} {\cellcolor{gray}} N/A \\
Jamesbond & {\cellcolor[HTML]{3D79B6}} \color[HTML]{F1F1F1} 92.5 & {\cellcolor[HTML]{5C51A3}} \color[HTML]{F1F1F1} 99.5 & {\cellcolor[HTML]{4D65AD}} \color[HTML]{F1F1F1} 95.9 & {\cellcolor[HTML]{4B68AE}} \color[HTML]{F1F1F1} 95.6 & {\cellcolor[HTML]{4175B4}} \color[HTML]{F1F1F1} 93.3 & {\cellcolor[HTML]{555AA7}} \color[HTML]{F1F1F1} 98.0 & {\cellcolor[HTML]{4B68AE}} \color[HTML]{F1F1F1} 95.6 & {\cellcolor[HTML]{486CB0}} \color[HTML]{F1F1F1} 94.8 & {\cellcolor[HTML]{000000}} \color[HTML]{F1F1F1} {\cellcolor{gray}} N/A & {\cellcolor[HTML]{000000}} \color[HTML]{F1F1F1} {\cellcolor{gray}} N/A & {\cellcolor[HTML]{000000}} \color[HTML]{F1F1F1} {\cellcolor{gray}} N/A & {\cellcolor[HTML]{000000}} \color[HTML]{F1F1F1} {\cellcolor{gray}} N/A \\
Kangaroo & {\cellcolor[HTML]{5061AA}} \color[HTML]{F1F1F1} 96.7 & {\cellcolor[HTML]{4175B4}} \color[HTML]{F1F1F1} 93.1 & {\cellcolor[HTML]{486CB0}} \color[HTML]{F1F1F1} 94.9 & {\cellcolor[HTML]{4B68AE}} \color[HTML]{F1F1F1} 95.6 & {\cellcolor[HTML]{5758A6}} \color[HTML]{F1F1F1} 98.3 & {\cellcolor[HTML]{4175B4}} \color[HTML]{F1F1F1} 93.2 & {\cellcolor[HTML]{4B68AE}} \color[HTML]{F1F1F1} 95.7 & {\cellcolor[HTML]{486CB0}} \color[HTML]{F1F1F1} 94.8 & {\cellcolor[HTML]{4E63AC}} \color[HTML]{F1F1F1} 96.1 & {\cellcolor[HTML]{4175B4}} \color[HTML]{F1F1F1} 93.1 & {\cellcolor[HTML]{486CB0}} \color[HTML]{F1F1F1} 94.6 & {\cellcolor[HTML]{496AAF}} \color[HTML]{F1F1F1} 95.2 \\
Krull & {\cellcolor[HTML]{486CB0}} \color[HTML]{F1F1F1} 94.8 & {\cellcolor[HTML]{5061AA}} \color[HTML]{F1F1F1} 96.8 & {\cellcolor[HTML]{4D65AD}} \color[HTML]{F1F1F1} 95.8 & {\cellcolor[HTML]{358BBC}} \color[HTML]{F1F1F1} 89.1 & {\cellcolor[HTML]{4B68AE}} \color[HTML]{F1F1F1} 95.6 & {\cellcolor[HTML]{5061AA}} \color[HTML]{F1F1F1} 96.7 & {\cellcolor[HTML]{4E63AC}} \color[HTML]{F1F1F1} 96.2 & {\cellcolor[HTML]{358BBC}} \color[HTML]{F1F1F1} 89.4 & {\cellcolor[HTML]{000000}} \color[HTML]{F1F1F1} {\cellcolor{gray}} N/A & {\cellcolor[HTML]{000000}} \color[HTML]{F1F1F1} {\cellcolor{gray}} N/A & {\cellcolor[HTML]{000000}} \color[HTML]{F1F1F1} {\cellcolor{gray}} N/A & {\cellcolor[HTML]{000000}} \color[HTML]{F1F1F1} {\cellcolor{gray}} N/A \\
MontezumaRev. & {\cellcolor[HTML]{5C51A3}} \color[HTML]{F1F1F1} 99.5 & {\cellcolor[HTML]{5C51A3}} \color[HTML]{F1F1F1} 99.4 & {\cellcolor[HTML]{5C51A3}} \color[HTML]{F1F1F1} 99.5 & {\cellcolor[HTML]{496AAF}} \color[HTML]{F1F1F1} 95.2 & {\cellcolor[HTML]{5E4FA2}} \color[HTML]{F1F1F1} 100 & {\cellcolor[HTML]{5E4FA2}} \color[HTML]{F1F1F1} 100 & {\cellcolor[HTML]{5E4FA2}} \color[HTML]{F1F1F1} 100 & {\cellcolor[HTML]{555AA7}} \color[HTML]{F1F1F1} 97.9 & {\cellcolor[HTML]{5E4FA2}} \color[HTML]{F1F1F1} 100 & {\cellcolor[HTML]{5E4FA2}} \color[HTML]{F1F1F1} 100 & {\cellcolor[HTML]{5E4FA2}} \color[HTML]{F1F1F1} 100 & {\cellcolor[HTML]{5758A6}} \color[HTML]{F1F1F1} 98.2 \\
MsPacman & {\cellcolor[HTML]{74C7A5}} \color[HTML]{000000} 77.9 & {\cellcolor[HTML]{5C51A3}} \color[HTML]{F1F1F1} 99.4 & {\cellcolor[HTML]{3F97B7}} \color[HTML]{F1F1F1} 87.4 & {\cellcolor[HTML]{50A9AF}} \color[HTML]{F1F1F1} 84.2 & {\cellcolor[HTML]{9CD7A4}} \color[HTML]{000000} 72.1 & {\cellcolor[HTML]{5C51A3}} \color[HTML]{F1F1F1} 99.3 & {\cellcolor[HTML]{52ABAE}} \color[HTML]{F1F1F1} 83.6 & {\cellcolor[HTML]{56B0AD}} \color[HTML]{F1F1F1} 83.1 & {\cellcolor[HTML]{000000}} \color[HTML]{F1F1F1} {\cellcolor{gray}} N/A & {\cellcolor[HTML]{000000}} \color[HTML]{F1F1F1} {\cellcolor{gray}} N/A & {\cellcolor[HTML]{000000}} \color[HTML]{F1F1F1} {\cellcolor{gray}} N/A & {\cellcolor[HTML]{000000}} \color[HTML]{F1F1F1} {\cellcolor{gray}} N/A \\
Pacman & {\cellcolor[HTML]{EAF79E}} \color[HTML]{000000} 58.5 & {\cellcolor[HTML]{3F77B5}} \color[HTML]{F1F1F1} 92.7 & {\cellcolor[HTML]{9FD8A4}} \color[HTML]{000000} 71.7 & {\cellcolor[HTML]{64C0A6}} \color[HTML]{000000} 80.4 & {\cellcolor[HTML]{FCFEBA}} \color[HTML]{000000} 51.3 & {\cellcolor[HTML]{3990BA}} \color[HTML]{F1F1F1} 88.6 & {\cellcolor[HTML]{C8E99E}} \color[HTML]{000000} 65.0 & {\cellcolor[HTML]{76C8A5}} \color[HTML]{000000} 77.4 & {\cellcolor[HTML]{FFF7B2}} \color[HTML]{000000} 47.6 & {\cellcolor[HTML]{56B0AD}} \color[HTML]{F1F1F1} 83.1 & {\cellcolor[HTML]{E4F498}} \color[HTML]{000000} 60.5 & {\cellcolor[HTML]{9CD7A4}} \color[HTML]{000000} 72.1 \\
Pitfall & {\cellcolor[HTML]{5758A6}} \color[HTML]{F1F1F1} 98.2 & {\cellcolor[HTML]{5B53A4}} \color[HTML]{F1F1F1} 99.0 & {\cellcolor[HTML]{5956A5}} \color[HTML]{F1F1F1} 98.6 & {\cellcolor[HTML]{4D65AD}} \color[HTML]{F1F1F1} 95.8 & {\cellcolor[HTML]{5E4FA2}} \color[HTML]{F1F1F1} 100 & {\cellcolor[HTML]{5E4FA2}} \color[HTML]{F1F1F1} 100 & {\cellcolor[HTML]{5E4FA2}} \color[HTML]{F1F1F1} 100 & {\cellcolor[HTML]{5061AA}} \color[HTML]{F1F1F1} 96.6 & {\cellcolor[HTML]{000000}} \color[HTML]{F1F1F1} {\cellcolor{gray}} N/A & {\cellcolor[HTML]{000000}} \color[HTML]{F1F1F1} {\cellcolor{gray}} N/A & {\cellcolor[HTML]{000000}} \color[HTML]{F1F1F1} {\cellcolor{gray}} N/A & {\cellcolor[HTML]{000000}} \color[HTML]{F1F1F1} {\cellcolor{gray}} N/A \\
Pong & {\cellcolor[HTML]{3387BC}} \color[HTML]{F1F1F1} 90.0 & {\cellcolor[HTML]{5B53A4}} \color[HTML]{F1F1F1} 99.1 & {\cellcolor[HTML]{466EB1}} \color[HTML]{F1F1F1} 94.3 & {\cellcolor[HTML]{5CB7AA}} \color[HTML]{F1F1F1} 81.7 & {\cellcolor[HTML]{466EB1}} \color[HTML]{F1F1F1} 94.3 & {\cellcolor[HTML]{5956A5}} \color[HTML]{F1F1F1} 98.8 & {\cellcolor[HTML]{5061AA}} \color[HTML]{F1F1F1} 96.5 & {\cellcolor[HTML]{56B0AD}} \color[HTML]{F1F1F1} 83.2 & {\cellcolor[HTML]{4471B2}} \color[HTML]{F1F1F1} 93.8 & {\cellcolor[HTML]{545CA8}} \color[HTML]{F1F1F1} 97.4 & {\cellcolor[HTML]{4B68AE}} \color[HTML]{F1F1F1} 95.6 & {\cellcolor[HTML]{4EA7B0}} \color[HTML]{F1F1F1} 84.7 \\
PrivateEye & {\cellcolor[HTML]{4B68AE}} \color[HTML]{F1F1F1} 95.7 & {\cellcolor[HTML]{4175B4}} \color[HTML]{F1F1F1} 93.0 & {\cellcolor[HTML]{466EB1}} \color[HTML]{F1F1F1} 94.3 & {\cellcolor[HTML]{525FA9}} \color[HTML]{F1F1F1} 97.0 & {\cellcolor[HTML]{5061AA}} \color[HTML]{F1F1F1} 96.5 & {\cellcolor[HTML]{5956A5}} \color[HTML]{F1F1F1} 98.6 & {\cellcolor[HTML]{545CA8}} \color[HTML]{F1F1F1} 97.5 & {\cellcolor[HTML]{4B68AE}} \color[HTML]{F1F1F1} 95.4 & {\cellcolor[HTML]{000000}} \color[HTML]{F1F1F1} {\cellcolor{gray}} N/A & {\cellcolor[HTML]{000000}} \color[HTML]{F1F1F1} {\cellcolor{gray}} N/A & {\cellcolor[HTML]{000000}} \color[HTML]{F1F1F1} {\cellcolor{gray}} N/A & {\cellcolor[HTML]{000000}} \color[HTML]{F1F1F1} {\cellcolor{gray}} N/A \\
Qbert & {\cellcolor[HTML]{466EB1}} \color[HTML]{F1F1F1} 94.4 & {\cellcolor[HTML]{5B53A4}} \color[HTML]{F1F1F1} 99.0 & {\cellcolor[HTML]{5061AA}} \color[HTML]{F1F1F1} 96.6 & {\cellcolor[HTML]{5C51A3}} \color[HTML]{F1F1F1} 99.6 & {\cellcolor[HTML]{89D0A4}} \color[HTML]{000000} 74.7 & {\cellcolor[HTML]{5758A6}} \color[HTML]{F1F1F1} 98.3 & {\cellcolor[HTML]{4BA4B1}} \color[HTML]{F1F1F1} 84.9 & {\cellcolor[HTML]{5758A6}} \color[HTML]{F1F1F1} 98.4 & {\cellcolor[HTML]{79C9A5}} \color[HTML]{000000} 77.3 & {\cellcolor[HTML]{5758A6}} \color[HTML]{F1F1F1} 98.4 & {\cellcolor[HTML]{439BB5}} \color[HTML]{F1F1F1} 86.6 & {\cellcolor[HTML]{5956A5}} \color[HTML]{F1F1F1} 98.5 \\
Riverraid & {\cellcolor[HTML]{4273B3}} \color[HTML]{F1F1F1} 93.5 & {\cellcolor[HTML]{555AA7}} \color[HTML]{F1F1F1} 98.0 & {\cellcolor[HTML]{4B68AE}} \color[HTML]{F1F1F1} 95.7 & {\cellcolor[HTML]{4273B3}} \color[HTML]{F1F1F1} 93.6 & {\cellcolor[HTML]{358BBC}} \color[HTML]{F1F1F1} 89.3 & {\cellcolor[HTML]{555AA7}} \color[HTML]{F1F1F1} 98.0 & {\cellcolor[HTML]{4273B3}} \color[HTML]{F1F1F1} 93.5 & {\cellcolor[HTML]{3682BA}} \color[HTML]{F1F1F1} 91.0 & {\cellcolor[HTML]{000000}} \color[HTML]{F1F1F1} {\cellcolor{gray}} N/A & {\cellcolor[HTML]{000000}} \color[HTML]{F1F1F1} {\cellcolor{gray}} N/A & {\cellcolor[HTML]{000000}} \color[HTML]{F1F1F1} {\cellcolor{gray}} N/A & {\cellcolor[HTML]{000000}} \color[HTML]{F1F1F1} {\cellcolor{gray}} N/A \\
RoadRunner & {\cellcolor[HTML]{4B68AE}} \color[HTML]{F1F1F1} 95.5 & {\cellcolor[HTML]{545CA8}} \color[HTML]{F1F1F1} 97.5 & {\cellcolor[HTML]{5061AA}} \color[HTML]{F1F1F1} 96.5 & {\cellcolor[HTML]{4175B4}} \color[HTML]{F1F1F1} 93.1 & {\cellcolor[HTML]{49A2B2}} \color[HTML]{F1F1F1} 85.2 & {\cellcolor[HTML]{6EC5A5}} \color[HTML]{000000} 78.7 & {\cellcolor[HTML]{5CB7AA}} \color[HTML]{F1F1F1} 81.8 & {\cellcolor[HTML]{3F97B7}} \color[HTML]{F1F1F1} 87.4 & {\cellcolor[HTML]{000000}} \color[HTML]{F1F1F1} {\cellcolor{gray}} N/A & {\cellcolor[HTML]{000000}} \color[HTML]{F1F1F1} {\cellcolor{gray}} N/A & {\cellcolor[HTML]{000000}} \color[HTML]{F1F1F1} {\cellcolor{gray}} N/A & {\cellcolor[HTML]{000000}} \color[HTML]{F1F1F1} {\cellcolor{gray}} N/A \\
Seaquest & {\cellcolor[HTML]{4471B2}} \color[HTML]{F1F1F1} 94.1 & {\cellcolor[HTML]{3B92B9}} \color[HTML]{F1F1F1} 87.9 & {\cellcolor[HTML]{3682BA}} \color[HTML]{F1F1F1} 90.9 & {\cellcolor[HTML]{3585BB}} \color[HTML]{F1F1F1} 90.3 & {\cellcolor[HTML]{3A7EB8}} \color[HTML]{F1F1F1} 91.5 & {\cellcolor[HTML]{5EB9A9}} \color[HTML]{F1F1F1} 81.3 & {\cellcolor[HTML]{459EB4}} \color[HTML]{F1F1F1} 86.1 & {\cellcolor[HTML]{3880B9}} \color[HTML]{F1F1F1} 91.4 & {\cellcolor[HTML]{3B7CB7}} \color[HTML]{F1F1F1} 92.1 & {\cellcolor[HTML]{58B2AC}} \color[HTML]{F1F1F1} 82.6 & {\cellcolor[HTML]{4199B6}} \color[HTML]{F1F1F1} 87.1 & {\cellcolor[HTML]{3585BB}} \color[HTML]{F1F1F1} 90.6 \\
Skiing & {\cellcolor[HTML]{4D65AD}} \color[HTML]{F1F1F1} 95.8 & {\cellcolor[HTML]{5061AA}} \color[HTML]{F1F1F1} 96.5 & {\cellcolor[HTML]{4E63AC}} \color[HTML]{F1F1F1} 96.2 & {\cellcolor[HTML]{3585BB}} \color[HTML]{F1F1F1} 90.4 & {\cellcolor[HTML]{4471B2}} \color[HTML]{F1F1F1} 94.1 & {\cellcolor[HTML]{466EB1}} \color[HTML]{F1F1F1} 94.2 & {\cellcolor[HTML]{466EB1}} \color[HTML]{F1F1F1} 94.2 & {\cellcolor[HTML]{358BBC}} \color[HTML]{F1F1F1} 89.3 & {\cellcolor[HTML]{000000}} \color[HTML]{F1F1F1} {\cellcolor{gray}} N/A & {\cellcolor[HTML]{000000}} \color[HTML]{F1F1F1} {\cellcolor{gray}} N/A & {\cellcolor[HTML]{000000}} \color[HTML]{F1F1F1} {\cellcolor{gray}} N/A & {\cellcolor[HTML]{000000}} \color[HTML]{F1F1F1} {\cellcolor{gray}} N/A \\
SpaceInv. & {\cellcolor[HTML]{496AAF}} \color[HTML]{F1F1F1} 95.2 & {\cellcolor[HTML]{5956A5}} \color[HTML]{F1F1F1} 98.7 & {\cellcolor[HTML]{525FA9}} \color[HTML]{F1F1F1} 96.9 & {\cellcolor[HTML]{525FA9}} \color[HTML]{F1F1F1} 97.1 & {\cellcolor[HTML]{3585BB}} \color[HTML]{F1F1F1} 90.6 & {\cellcolor[HTML]{4D65AD}} \color[HTML]{F1F1F1} 95.9 & {\cellcolor[HTML]{4175B4}} \color[HTML]{F1F1F1} 93.1 & {\cellcolor[HTML]{545CA8}} \color[HTML]{F1F1F1} 97.3 & {\cellcolor[HTML]{000000}} \color[HTML]{F1F1F1} {\cellcolor{gray}} N/A & {\cellcolor[HTML]{000000}} \color[HTML]{F1F1F1} {\cellcolor{gray}} N/A & {\cellcolor[HTML]{000000}} \color[HTML]{F1F1F1} {\cellcolor{gray}} N/A & {\cellcolor[HTML]{000000}} \color[HTML]{F1F1F1} {\cellcolor{gray}} N/A \\
Tennis & {\cellcolor[HTML]{5956A5}} \color[HTML]{F1F1F1} 98.7 & {\cellcolor[HTML]{5E4FA2}} \color[HTML]{F1F1F1} 99.9 & {\cellcolor[HTML]{5C51A3}} \color[HTML]{F1F1F1} 99.3 & {\cellcolor[HTML]{47A0B3}} \color[HTML]{F1F1F1} 85.7 & {\cellcolor[HTML]{4471B2}} \color[HTML]{F1F1F1} 93.9 & {\cellcolor[HTML]{5956A5}} \color[HTML]{F1F1F1} 98.7 & {\cellcolor[HTML]{4E63AC}} \color[HTML]{F1F1F1} 96.2 & {\cellcolor[HTML]{52ABAE}} \color[HTML]{F1F1F1} 83.9 & {\cellcolor[HTML]{000000}} \color[HTML]{F1F1F1} {\cellcolor{gray}} N/A & {\cellcolor[HTML]{000000}} \color[HTML]{F1F1F1} {\cellcolor{gray}} N/A & {\cellcolor[HTML]{000000}} \color[HTML]{F1F1F1} {\cellcolor{gray}} N/A & {\cellcolor[HTML]{000000}} \color[HTML]{F1F1F1} {\cellcolor{gray}} N/A \\
TimePilot & {\cellcolor[HTML]{4273B3}} \color[HTML]{F1F1F1} 93.5 & {\cellcolor[HTML]{486CB0}} \color[HTML]{F1F1F1} 94.7 & {\cellcolor[HTML]{4471B2}} \color[HTML]{F1F1F1} 94.1 & {\cellcolor[HTML]{5061AA}} \color[HTML]{F1F1F1} 96.6 & {\cellcolor[HTML]{3880B9}} \color[HTML]{F1F1F1} 91.3 & {\cellcolor[HTML]{466EB1}} \color[HTML]{F1F1F1} 94.3 & {\cellcolor[HTML]{3F77B5}} \color[HTML]{F1F1F1} 92.8 & {\cellcolor[HTML]{486CB0}} \color[HTML]{F1F1F1} 94.6 & {\cellcolor[HTML]{000000}} \color[HTML]{F1F1F1} {\cellcolor{gray}} N/A & {\cellcolor[HTML]{000000}} \color[HTML]{F1F1F1} {\cellcolor{gray}} N/A & {\cellcolor[HTML]{000000}} \color[HTML]{F1F1F1} {\cellcolor{gray}} N/A & {\cellcolor[HTML]{000000}} \color[HTML]{F1F1F1} {\cellcolor{gray}} N/A \\
UpNDown & {\cellcolor[HTML]{5061AA}} \color[HTML]{F1F1F1} 96.8 & {\cellcolor[HTML]{5B53A4}} \color[HTML]{F1F1F1} 99.1 & {\cellcolor[HTML]{555AA7}} \color[HTML]{F1F1F1} 97.9 & {\cellcolor[HTML]{545CA8}} \color[HTML]{F1F1F1} 97.4 & {\cellcolor[HTML]{4175B4}} \color[HTML]{F1F1F1} 93.0 & {\cellcolor[HTML]{555AA7}} \color[HTML]{F1F1F1} 97.7 & {\cellcolor[HTML]{496AAF}} \color[HTML]{F1F1F1} 95.3 & {\cellcolor[HTML]{4175B4}} \color[HTML]{F1F1F1} 93.1 & {\cellcolor[HTML]{496AAF}} \color[HTML]{F1F1F1} 95.0 & {\cellcolor[HTML]{5758A6}} \color[HTML]{F1F1F1} 98.3 & {\cellcolor[HTML]{5061AA}} \color[HTML]{F1F1F1} 96.6 & {\cellcolor[HTML]{4D65AD}} \color[HTML]{F1F1F1} 95.9 \\
Venture & {\cellcolor[HTML]{D3ED9C}} \color[HTML]{000000} 63.1 & {\cellcolor[HTML]{5E4FA2}} \color[HTML]{F1F1F1} 99.9 & {\cellcolor[HTML]{76C8A5}} \color[HTML]{000000} 77.4 & {\cellcolor[HTML]{3B7CB7}} \color[HTML]{F1F1F1} 92.1 & {\cellcolor[HTML]{ECF7A1}} \color[HTML]{000000} 57.6 & {\cellcolor[HTML]{5E4FA2}} \color[HTML]{F1F1F1} 100 & {\cellcolor[HTML]{94D4A4}} \color[HTML]{000000} 73.1 & {\cellcolor[HTML]{3880B9}} \color[HTML]{F1F1F1} 91.3 & {\cellcolor[HTML]{000000}} \color[HTML]{F1F1F1} {\cellcolor{gray}} N/A & {\cellcolor[HTML]{000000}} \color[HTML]{F1F1F1} {\cellcolor{gray}} N/A & {\cellcolor[HTML]{000000}} \color[HTML]{F1F1F1} {\cellcolor{gray}} N/A & {\cellcolor[HTML]{000000}} \color[HTML]{F1F1F1} {\cellcolor{gray}} N/A \\
VideoPinball & {\cellcolor[HTML]{5758A6}} \color[HTML]{F1F1F1} 98.3 & {\cellcolor[HTML]{486CB0}} \color[HTML]{F1F1F1} 94.6 & {\cellcolor[HTML]{4E63AC}} \color[HTML]{F1F1F1} 96.4 & {\cellcolor[HTML]{466EB1}} \color[HTML]{F1F1F1} 94.4 & {\cellcolor[HTML]{5C51A3}} \color[HTML]{F1F1F1} 99.5 & {\cellcolor[HTML]{496AAF}} \color[HTML]{F1F1F1} 95.3 & {\cellcolor[HTML]{545CA8}} \color[HTML]{F1F1F1} 97.3 & {\cellcolor[HTML]{496AAF}} \color[HTML]{F1F1F1} 95.3 & {\cellcolor[HTML]{000000}} \color[HTML]{F1F1F1} {\cellcolor{gray}} N/A & {\cellcolor[HTML]{000000}} \color[HTML]{F1F1F1} {\cellcolor{gray}} N/A & {\cellcolor[HTML]{000000}} \color[HTML]{F1F1F1} {\cellcolor{gray}} N/A & {\cellcolor[HTML]{000000}} \color[HTML]{F1F1F1} {\cellcolor{gray}} N/A \\
\bottomrule
mean & {\cellcolor[HTML]{3585BB}} \color[HTML]{F1F1F1} 90.5 & {\cellcolor[HTML]{4273B3}} \color[HTML]{F1F1F1} 93.5 & {\cellcolor[HTML]{3880B9}} \color[HTML]{F1F1F1} 91.3 & {\cellcolor[HTML]{3682BA}} \color[HTML]{F1F1F1} 91.0 & {\cellcolor[HTML]{378EBB}} \color[HTML]{F1F1F1} 88.9 & {\cellcolor[HTML]{3D79B6}} \color[HTML]{F1F1F1} 92.3 & {\cellcolor[HTML]{3389BD}} \color[HTML]{F1F1F1} 89.7 & {\cellcolor[HTML]{3682BA}} \color[HTML]{F1F1F1} 90.7 & {\cellcolor[HTML]{378EBB}} \color[HTML]{F1F1F1} 88.8 & {\cellcolor[HTML]{3B7CB7}} \color[HTML]{F1F1F1} 92.1 & {\cellcolor[HTML]{3387BC}} \color[HTML]{F1F1F1} 90.0 & {\cellcolor[HTML]{3880B9}} \color[HTML]{F1F1F1} 91.4 \\
\bottomrule
\end{tabular}

}
\vspace{-0.5cm}
\label{tab:f1_detailed}
\end{table}
\renewcommand{\arraystretch}{1.}

\subsection{Alien \ details}
\IfFileExists{environment_description/Alien.tex}{
\begin{minipage}{0.60\textwidth}
You are stuck in a maze-like space ship with three aliens. You goal is to destroy their eggs that are scattered all over the ship while simultaneously avoiding the aliens (they are trying to kill you). You have a flamethrower that can help you turn them away in tricky situations. Moreover, you can occasionally collect a power-up (pulsar) that gives you the temporary ability to kill aliens.
\end{minipage}
}{}
\IfFileExists{environment_images/Alien.png}{
\begin{minipage}{0.38\textwidth}
\raggedleft
\includegraphics[scale=0.25]{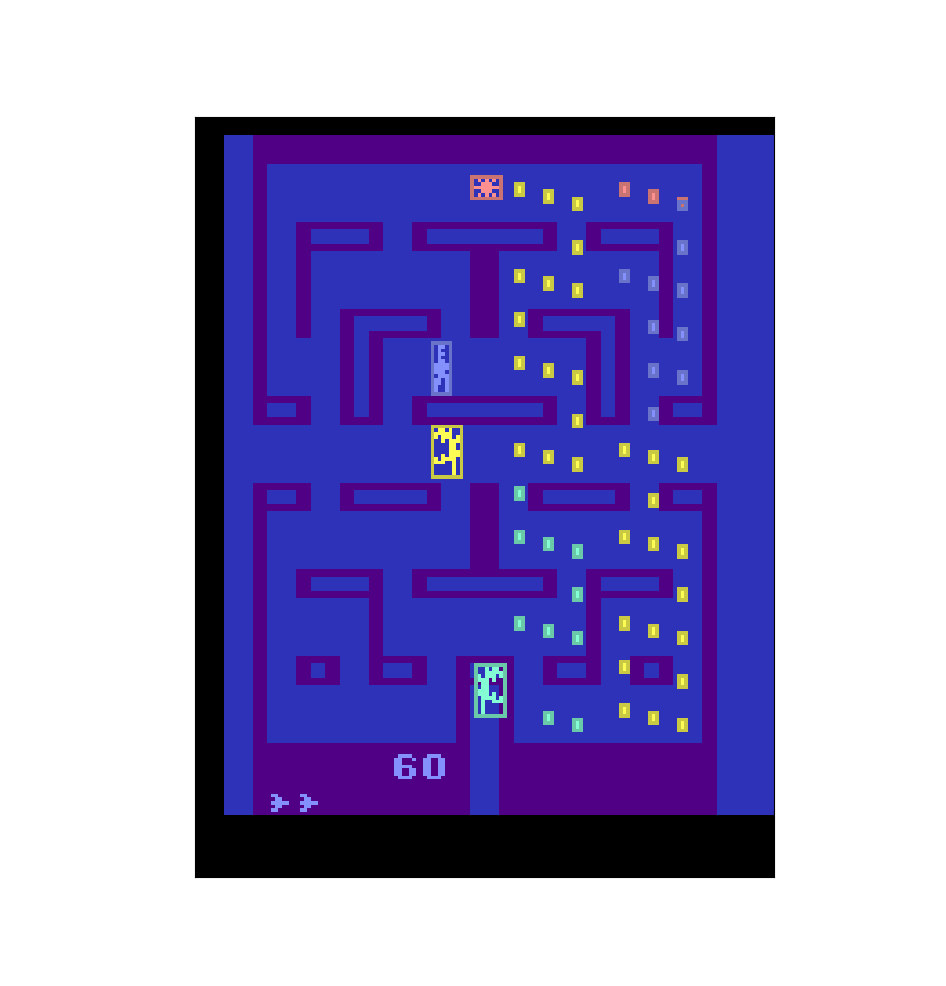}
\end{minipage}{}
}{}

\foreach \n in {Alien}{\input{reports/\n_stat_reports.tex}}

\subsection{Amidar \ details}
\IfFileExists{environment_description/Amidar.tex}{
\begin{minipage}{0.60\textwidth}
This game is similar to Pac-Man: You are trying to visit all places on a 2-dimensional grid while simultaneously avoiding your enemies. You can turn the tables at one point in the game: Your enemies turn into chickens and you can catch them.
\end{minipage}
}{}
\IfFileExists{environment_images/Amidar.png}{
\begin{minipage}{0.38\textwidth}
\raggedleft
\includegraphics[scale=0.25]{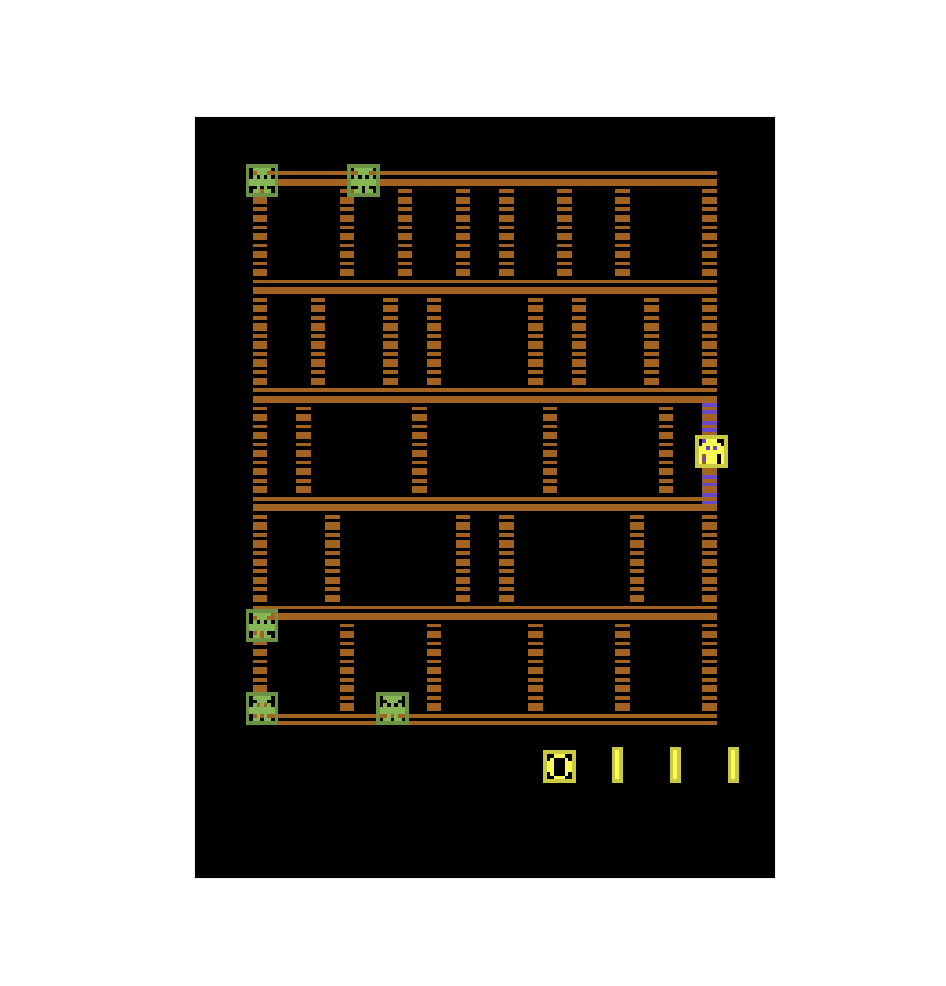}
\end{minipage}{}
}{}

\foreach \n in {Amidar}{\input{reports/\n_stat_reports.tex}}

\subsection{Assault \ details}
\IfFileExists{environment_description/Assault.tex}{
\begin{minipage}{0.60\textwidth}
You control a vehicle that can move sideways. A big mother ship circles overhead and continually deploys smaller drones. You must destroy these enemies and dodge their attacks.
\end{minipage}
}{}
\IfFileExists{environment_images/Assault.png}{
\begin{minipage}{0.38\textwidth}
\raggedleft
\includegraphics[scale=0.25]{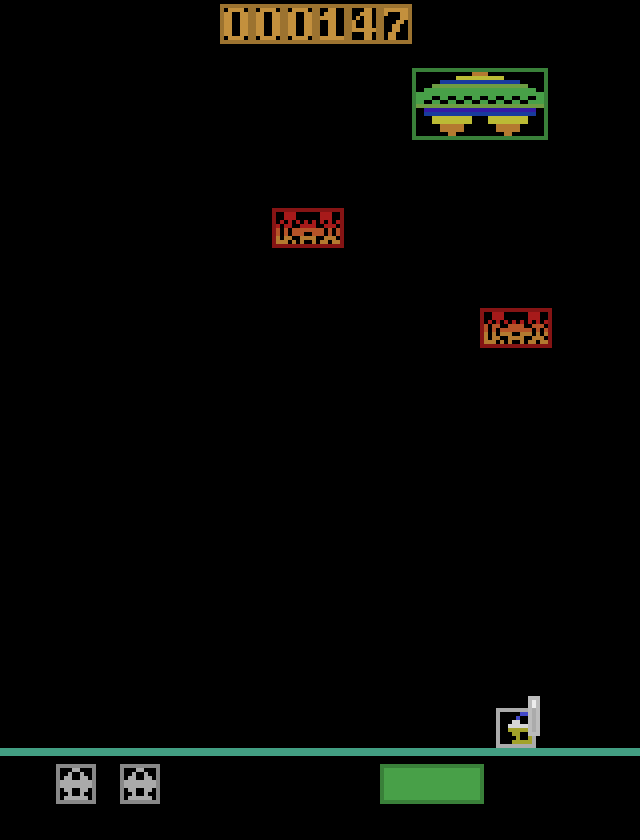}
\end{minipage}{}
}{}

\foreach \n in {Assault}{\input{reports/\n_stat_reports.tex}}

\subsection{Asterix \ details}
\IfFileExists{environment_description/Asterix.tex}{
\begin{minipage}{0.60\textwidth}
You are Asterix and can move horizontally (continuously) and vertically (discretely). Objects move horizontally across the screen: lyres and other (more useful) objects. Your goal is to guide Asterix in such a way as to avoid lyres and collect as many other objects as possible. You score points by collecting objects and lose a life whenever you collect a lyre. You have three lives available at the beginning. If you score sufficiently many points, you will be awarded additional points.
\end{minipage}
}{}
\IfFileExists{environment_images/Asterix.png}{
\begin{minipage}{0.38\textwidth}
\raggedleft
\includegraphics[scale=0.25]{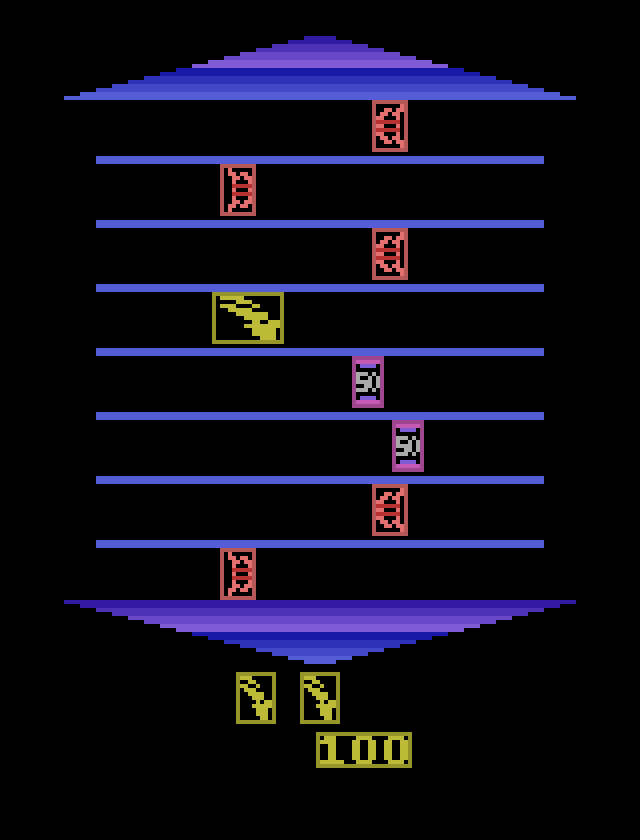}
\end{minipage}{}
}{}

\foreach \n in {Asterix}{\input{reports/\n_stat_reports.tex}}

\subsection{Asteroids \ details}
\IfFileExists{environment_description/Asteroids.tex}{
\begin{minipage}{0.60\textwidth}
This is a well-known arcade game: You control a spaceship in an asteroid field and must break up asteroids by shooting them. Once all asteroids are destroyed, you enter a new level and new asteroids will appear. You will occasionally be attacked by a flying saucer.
\end{minipage}
}{}
\IfFileExists{environment_images/Asteroids.png}{
\begin{minipage}{0.38\textwidth}
\raggedleft
\includegraphics[scale=0.25]{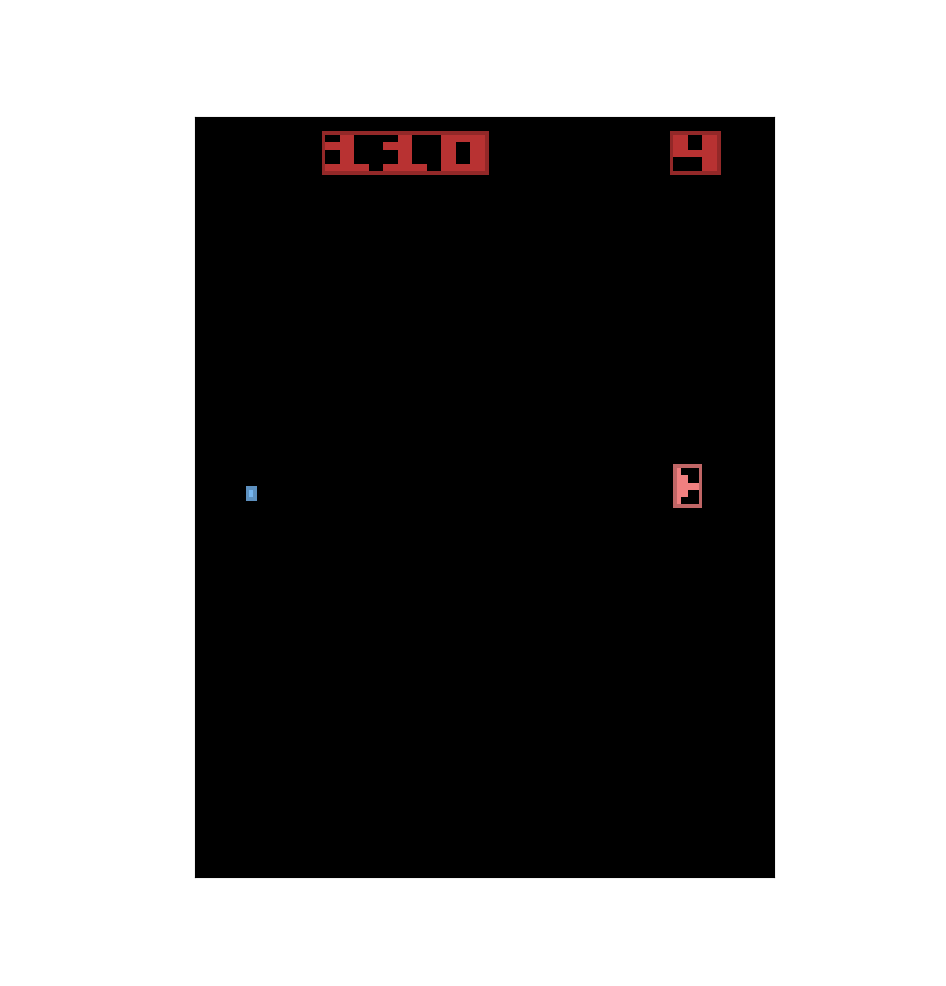}
\end{minipage}{}
}{}

\foreach \n in {Asteroids}{\input{reports/\n_stat_reports.tex}}

\subsection{Atlantis \ details}
\IfFileExists{environment_description/Atlantis.tex}{
\begin{minipage}{0.60\textwidth}
Your job is to defend the submerged city of Atlantis. Your enemies slowly descend towards the city and you must destroy them before they reach striking distance. To this end, you control three defense posts. You lose if your enemies manage to destroy all seven of Atlantis’ installations. You may rebuild installations after you have fought of a wave of enemies and scored a sufficient number of points.
\end{minipage}
}{}
\IfFileExists{environment_images/Atlantis.png}{
\begin{minipage}{0.38\textwidth}
\raggedleft
\includegraphics[scale=0.25]{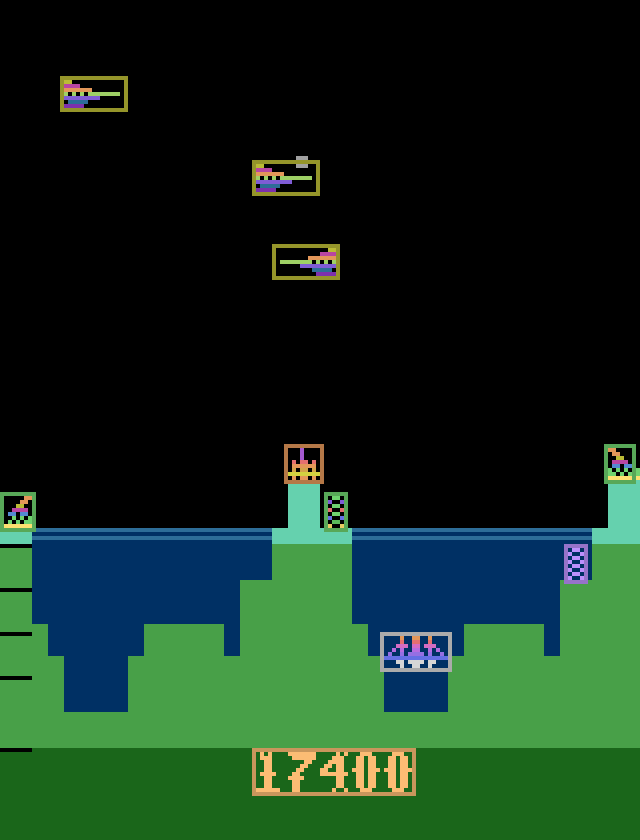}
\end{minipage}{}
}{}

\foreach \n in {Atlantis}{\input{reports/\n_stat_reports.tex}}

\subsection{BankHeist \ details}
\IfFileExists{environment_description/BankHeist.tex}{
\begin{minipage}{0.60\textwidth}
You are a bank robber and (naturally) want to rob as many banks as possible. You control your getaway car and must navigate maze-like cities. The police chases you and will appear whenever you rob a bank. You may destroy police cars by dropping sticks of dynamite. You can fill up your gas tank by entering a new city.At the beginning of the game you have four lives. Lives are lost if you run out of gas, are caught by the police,or run over the dynamite you have previously dropped.
\end{minipage}
}{}
\IfFileExists{environment_images/BankHeist.png}{
\begin{minipage}{0.38\textwidth}
\raggedleft
\includegraphics[scale=0.25]{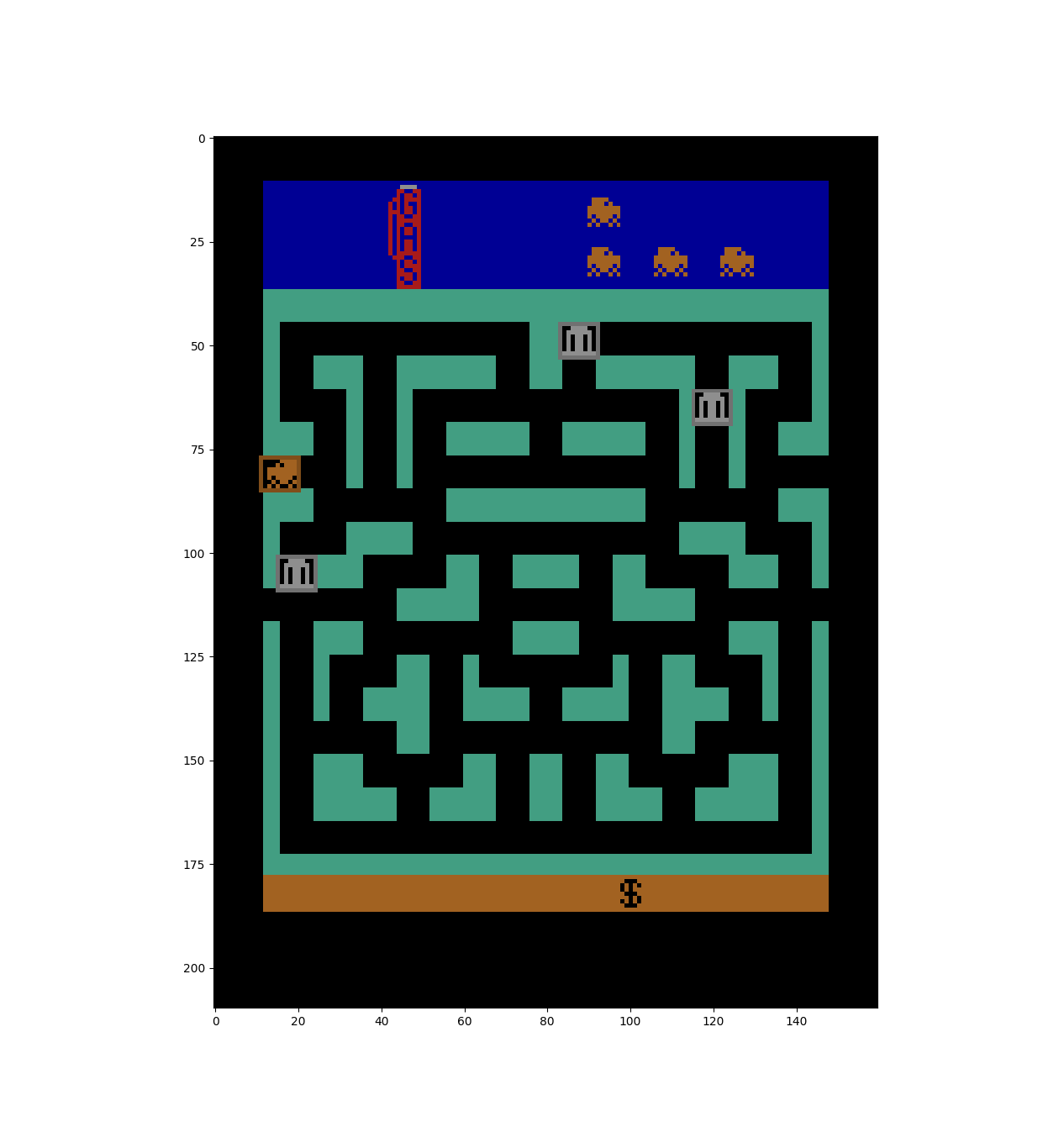}
\end{minipage}{}
}{}

\foreach \n in {BankHeist}{\input{reports/\n_stat_reports.tex}}

\subsection{BattleZone \ details}
\IfFileExists{environment_description/BattleZone.tex}{
\begin{minipage}{0.60\textwidth}
You control a tank and must destroy enemy vehicles. This game is played in a first-person perspective and creates a 3D illusion. A radar screen shows enemies around you. You start with 5 lives and gain up to 2 extra lives if you reach a sufficient score.
\end{minipage}
}{}
\IfFileExists{environment_images/BattleZone.png}{
\begin{minipage}{0.38\textwidth}
\raggedleft
\includegraphics[scale=0.25]{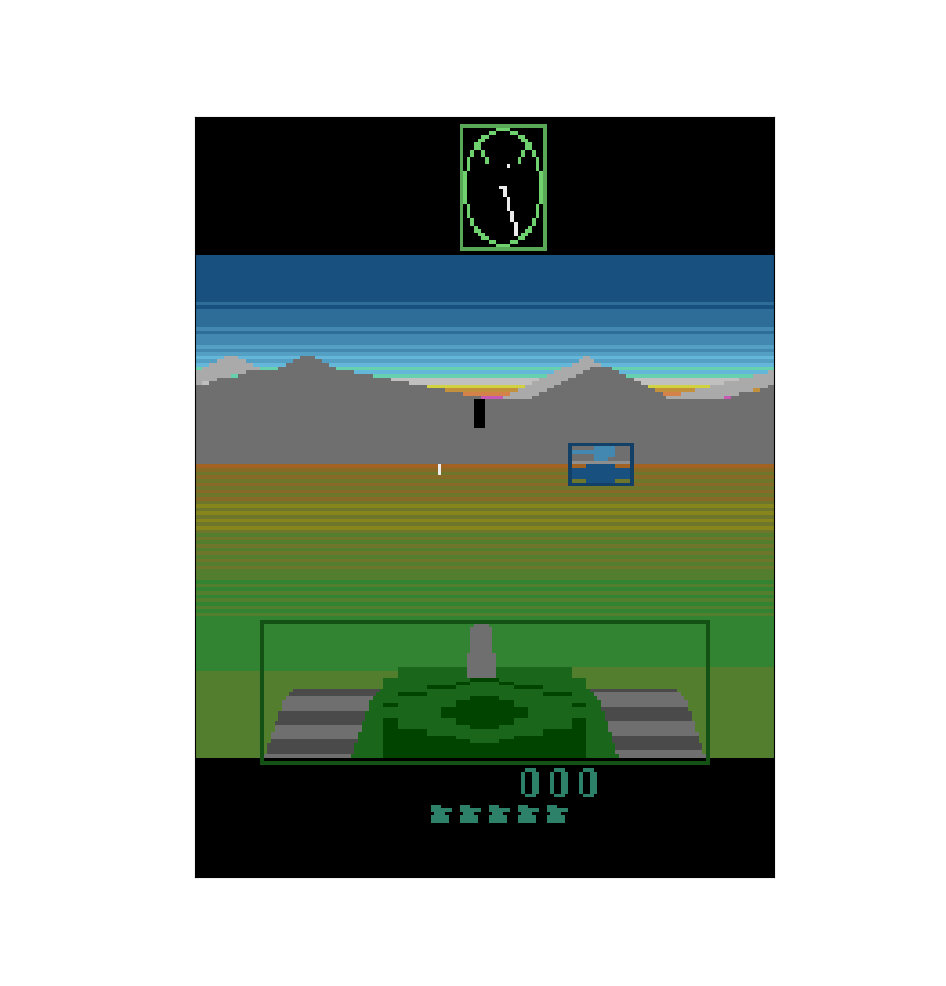}
\end{minipage}{}
}{}

\foreach \n in {BattleZone}{\input{reports/\n_stat_reports.tex}}

\subsection{Berzerk \ details}
\IfFileExists{environment_description/Berzerk.tex}{
\begin{minipage}{0.60\textwidth}
You are stuck in a maze with evil robots. You must destroy them and avoid touching the walls of the maze, as this will kill you. You may be awarded extra lives after scoring a sufficient number of points, depending on the game mode. You may also be chased by an undefeatable enemy, Evil Otto, that you must avoid. Evil Otto does not appear in the default mode.
\end{minipage}
}{}
\IfFileExists{environment_images/Berzerk.png}{
\begin{minipage}{0.38\textwidth}
\raggedleft
\includegraphics[scale=0.25]{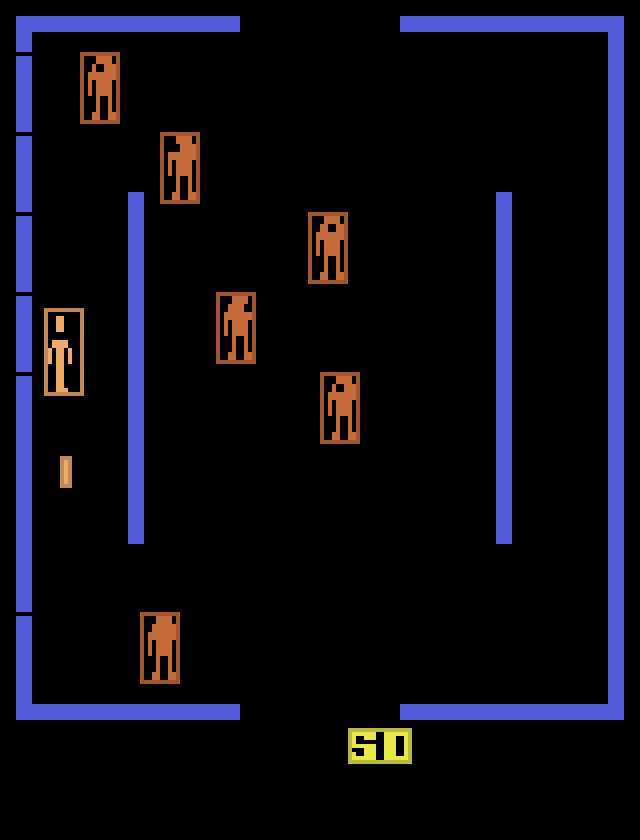}
\end{minipage}{}
}{}

\foreach \n in {Berzerk}{\input{reports/\n_stat_reports.tex}}

\subsection{Bowling \ details}
\IfFileExists{environment_description/Bowling.tex}{
\begin{minipage}{0.60\textwidth}
Your goal is to score as many points as possible in the game of Bowling. A game consists of 10 frames and you have two tries per frame. Knocking down all pins on the first try is called a “strike”. Knocking down all pins on the second roll is called a “spar”. Otherwise, the frame is called “open”.
\end{minipage}
}{}
\IfFileExists{environment_images/Bowling.png}{
\begin{minipage}{0.38\textwidth}
\raggedleft
\includegraphics[scale=0.25]{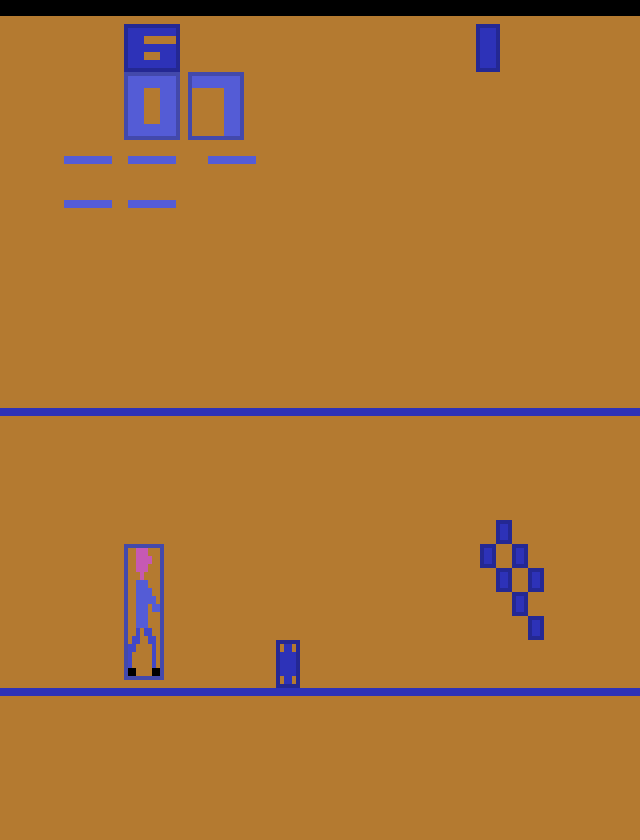}
\end{minipage}{}
}{}

\foreach \n in {Bowling}{\input{reports/\n_stat_reports.tex}}

\subsection{Boxing \ details}
\IfFileExists{environment_description/Boxing.tex}{
\begin{minipage}{0.60\textwidth}
You fight an opponent in a boxing ring. You score points for hitting the opponent. If you score 100 points, your opponent is knocked out.
\end{minipage}
}{}
\IfFileExists{environment_images/Boxing.png}{
\begin{minipage}{0.38\textwidth}
\raggedleft
\includegraphics[scale=0.25]{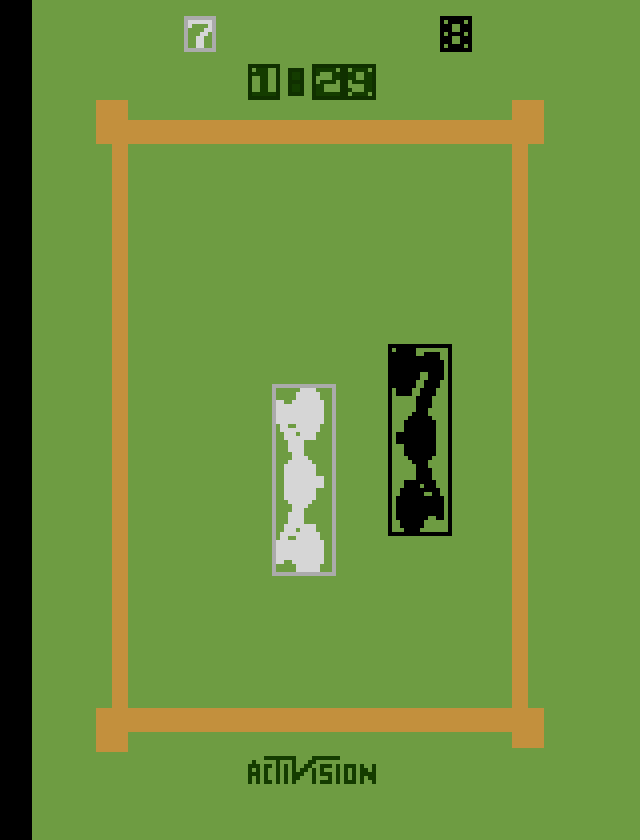}
\end{minipage}{}
}{}

\foreach \n in {Boxing}{\input{reports/\n_stat_reports.tex}}

\subsection{Breakout \ details}
\IfFileExists{environment_description/Breakout.tex}{
\begin{minipage}{0.60\textwidth}
Another famous Atari game. The dynamics are similar to pong: You move a paddle and hit the ball in a brick wall at the top of the screen. Your goal is to destroy the brick wall. You can try to break through the wall and let the ball wreak havoc on the other side, all on its own! You have five lives.
\end{minipage}
}{}
\IfFileExists{environment_images/Breakout.png}{
\begin{minipage}{0.38\textwidth}
\raggedleft
\includegraphics[scale=0.25]{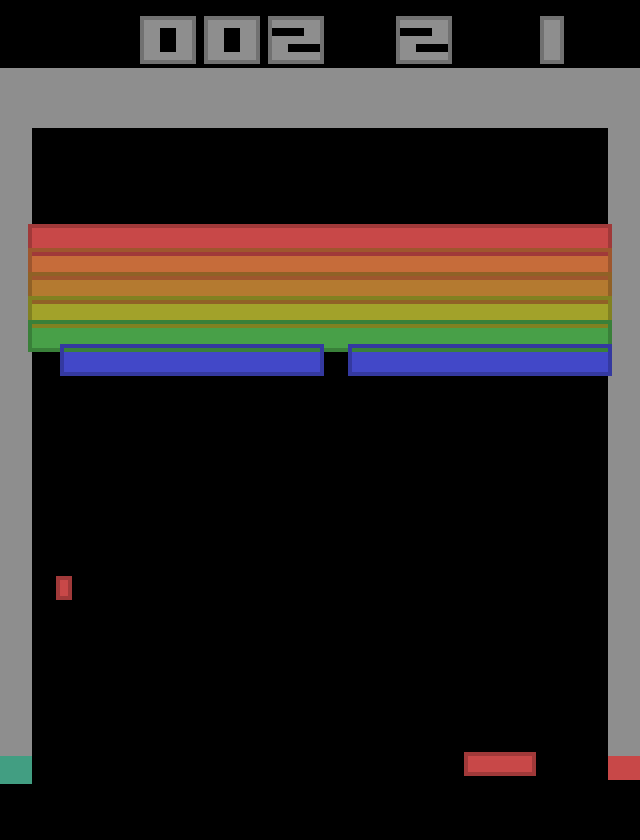}
\end{minipage}{}
}{}

\foreach \n in {Breakout}{\input{reports/\n_stat_reports.tex}}

\subsection{Carnival \ details}
\IfFileExists{environment_description/Carnival.tex}{
\begin{minipage}{0.60\textwidth}
This is a “shoot ‘em up” game. Targets move horizontally across the screen and you must shoot them. You are in control of a gun that can be moved horizontally. The supply of ammunition is limited and chickens may steal some bullets from you if you don’t hit them in time.
\end{minipage}
}{}
\IfFileExists{environment_images/Carnival.png}{
\begin{minipage}{0.38\textwidth}
\raggedleft
\includegraphics[scale=0.25]{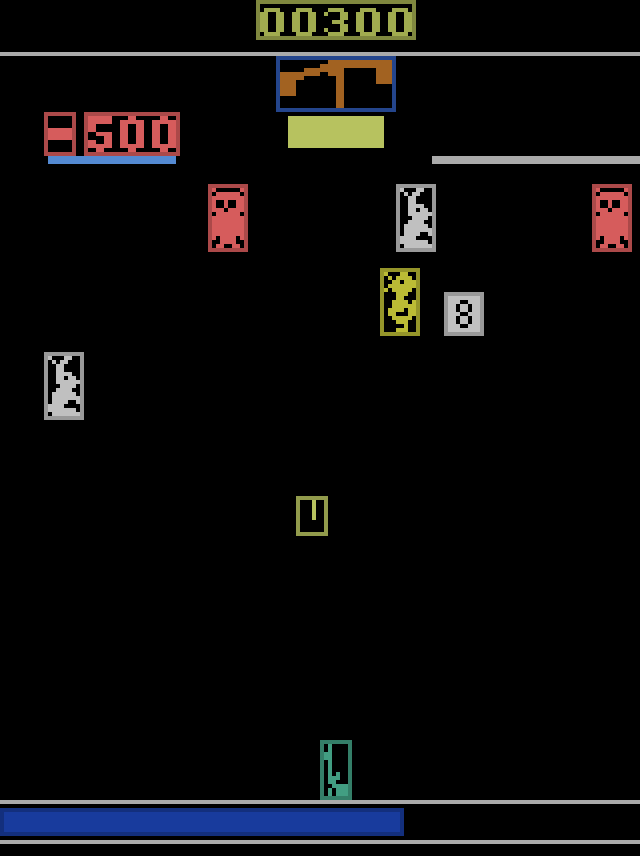}
\end{minipage}{}
}{}

\foreach \n in {Carnival}{\input{reports/\n_stat_reports.tex}}

\subsection{Centipede \ details}
\IfFileExists{environment_description/Centipede.tex}{
\begin{minipage}{0.60\textwidth}
You are an elf and must use your magic wands to fend off spiders, fleas and centipedes. Your goal is to protect mushrooms in an enchanted forest. If you are bitten by a spider, flea or centipede, you will be temporally paralyzed and you will lose a magic wand. The game ends once you have lost all wands. You may receive additional wands after scoring a sufficient number of points.
\end{minipage}
}{}
\IfFileExists{environment_images/Centipede.png}{
\begin{minipage}{0.38\textwidth}
\raggedleft
\includegraphics[scale=0.25]{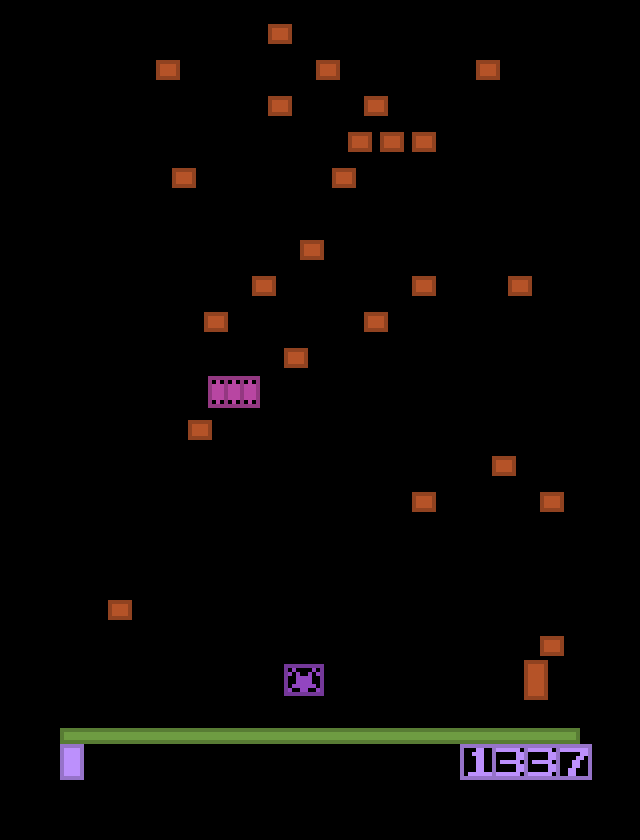}
\end{minipage}{}
}{}

\foreach \n in {Centipede}{\input{reports/\n_stat_reports.tex}}

\subsection{ChopperCommand \ details}
\IfFileExists{environment_description/ChopperCommand.tex}{
\begin{minipage}{0.60\textwidth}
You control a helicopter and must protect truck convoys. To that end, you need to shoot down enemy aircraft.A mini-map is displayed at the bottom of the screen.
\end{minipage}
}{}
\IfFileExists{environment_images/ChopperCommand.png}{
\begin{minipage}{0.38\textwidth}
\raggedleft
\includegraphics[scale=0.25]{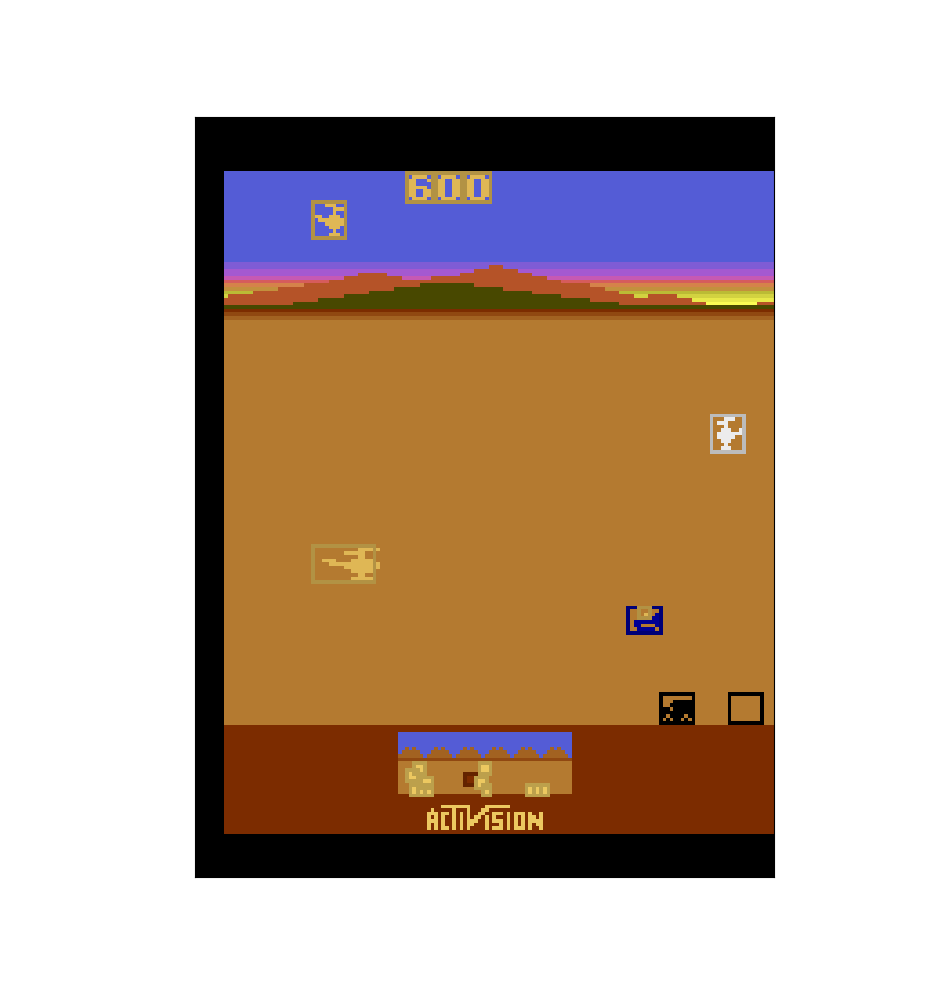}
\end{minipage}{}
}{}

\foreach \n in {ChopperCommand}{\input{reports/\n_stat_reports.tex}}

\subsection{CrazyClimber \ details}
\IfFileExists{environment_description/CrazyClimber.tex}{
\begin{minipage}{0.60\textwidth}
You are a climber trying to reach the top of four buildings, while avoiding obstacles like closing windows and falling objects. When you receive damage (windows closing or objects) you will fall and lose one life; you have a total of 5 lives before the end games. At the top of each building, there’s a helicopter which you need to catch to get to the next building. The goal is to climb as fast as possible while receiving the least amount of damage.
\end{minipage}
}{}
\IfFileExists{environment_images/CrazyClimber.png}{
\begin{minipage}{0.38\textwidth}
\raggedleft
\includegraphics[scale=0.25]{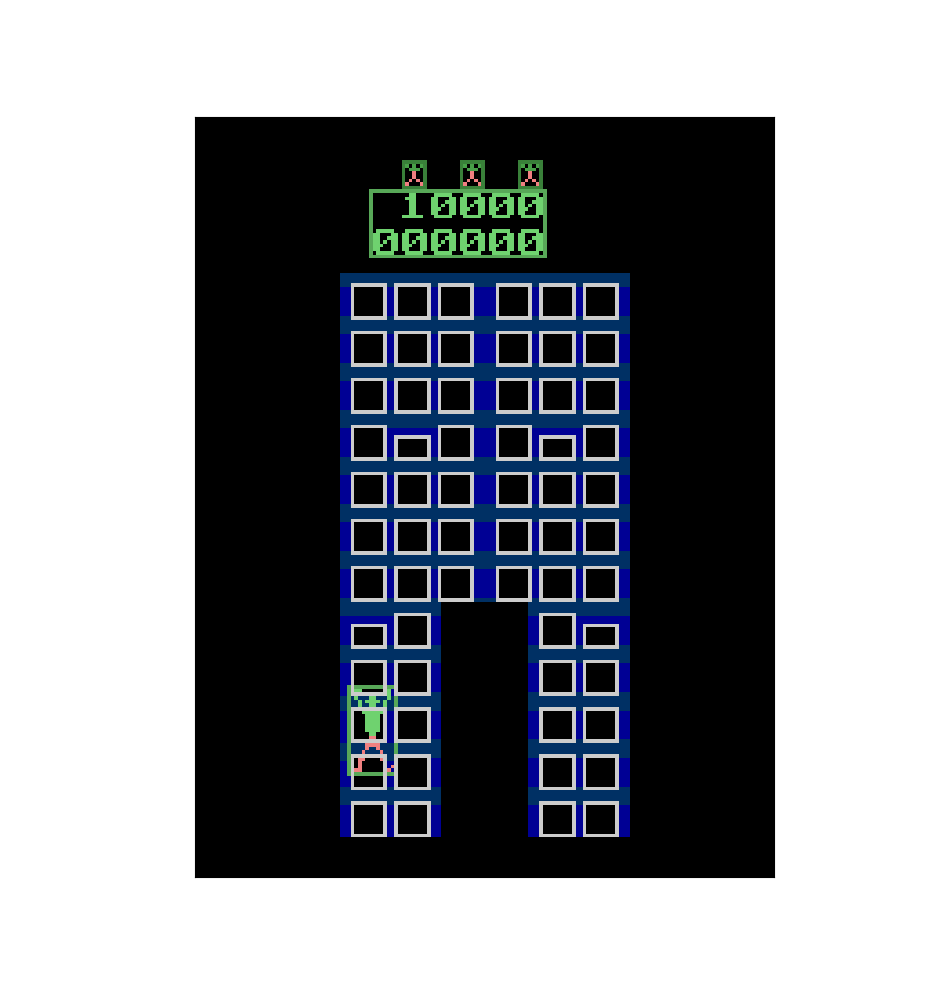}
\end{minipage}{}
}{}

\foreach \n in {CrazyClimber}{\input{reports/\n_stat_reports.tex}}

\subsection{DemonAttack \ details}
\IfFileExists{environment_description/DemonAttack.tex}{
\begin{minipage}{0.60\textwidth}
You are facing waves of demons in the ice planet of Krybor. Points are accumulated by destroying demons. You begin with 3 reserve bunkers, and can increase its number (up to 6) by avoiding enemy attacks. Each attack wave you survive without any hits, grants you a new bunker. Every time an enemy hits you, a bunker is destroyed. When the last bunker falls, the next enemy hit will destroy you and the game ends.
\end{minipage}
}{}
\IfFileExists{environment_images/DemonAttack.png}{
\begin{minipage}{0.38\textwidth}
\raggedleft
\includegraphics[scale=0.25]{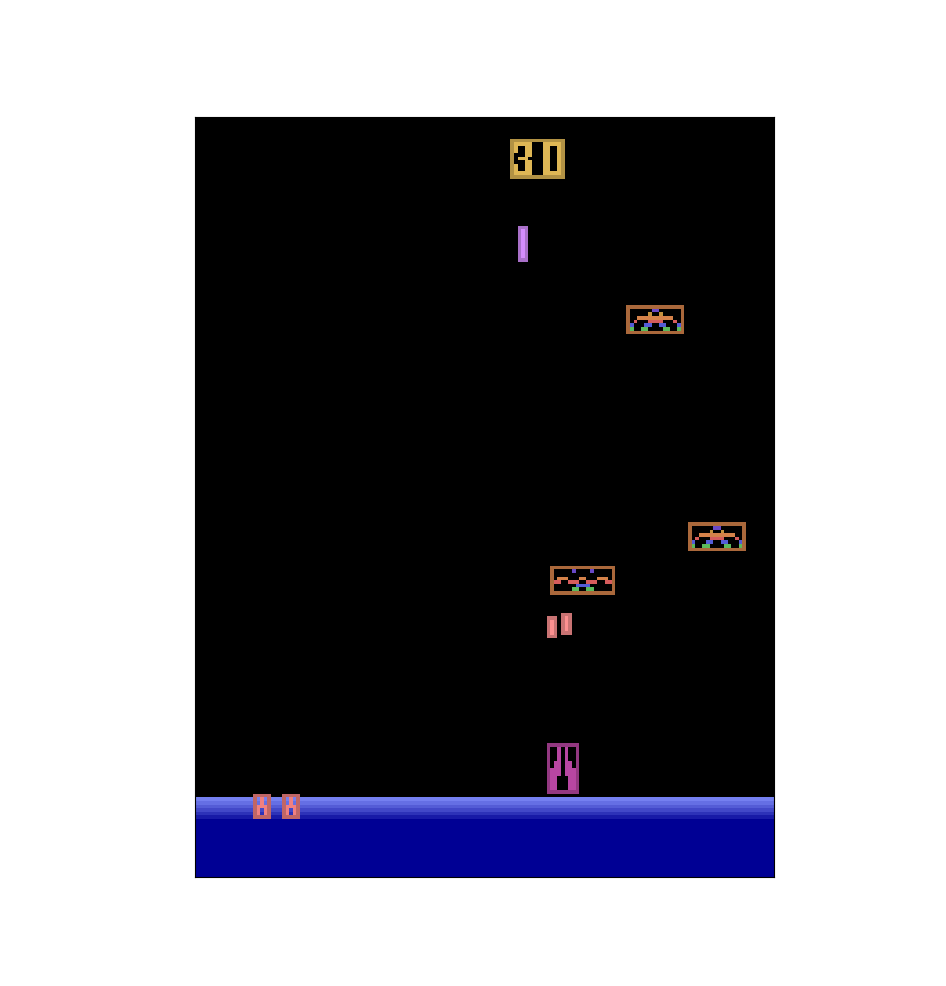}
\end{minipage}{}
}{}

\foreach \n in {DemonAttack}{\input{reports/\n_stat_reports.tex}}

\subsection{DonkeyKong \ details}
\IfFileExists{environment_description/DonkeyKong.tex}{
\begin{minipage}{0.60\textwidth}
You play as Mario trying to save your girlfriend who has been kidnapped by Donkey Kong. Remove rivets and jump over fireballs, with a score that starts high and counts down throughout the game.
\end{minipage}
}{}
\IfFileExists{environment_images/DonkeyKong.png}{
\begin{minipage}{0.38\textwidth}
\raggedleft
\includegraphics[scale=0.25]{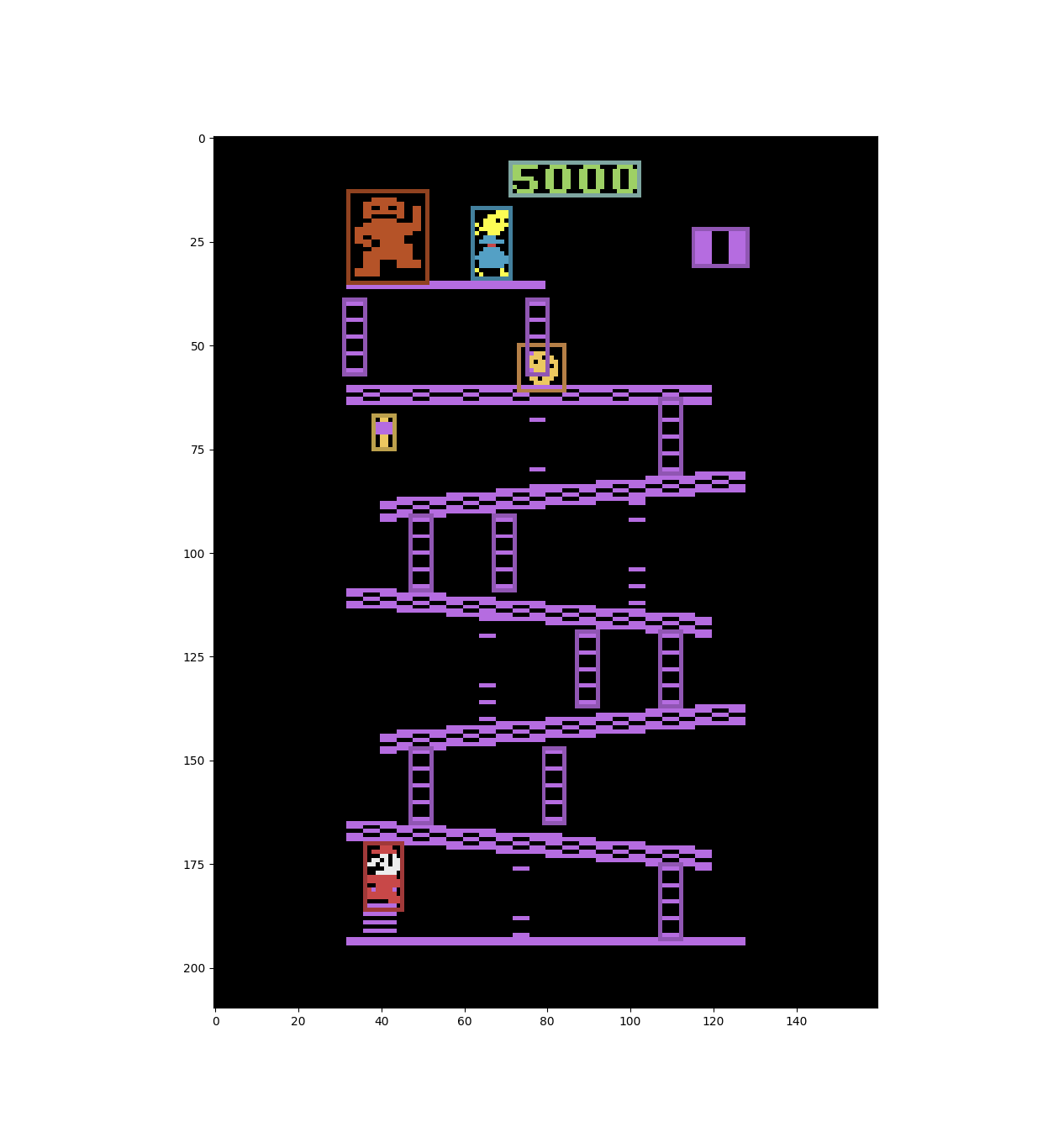}
\end{minipage}{}
}{}

\foreach \n in {DonkeyKong}{\input{reports/\n_stat_reports.tex}}

\subsection{FishingDerby \ details}
\IfFileExists{environment_description/FishingDerby.tex}{
\begin{minipage}{0.60\textwidth}
Your objective is to catch more sunfish than your opponent.
\end{minipage}
}{}
\IfFileExists{environment_images/FishingDerby.png}{
\begin{minipage}{0.38\textwidth}
\raggedleft
\includegraphics[scale=0.25]{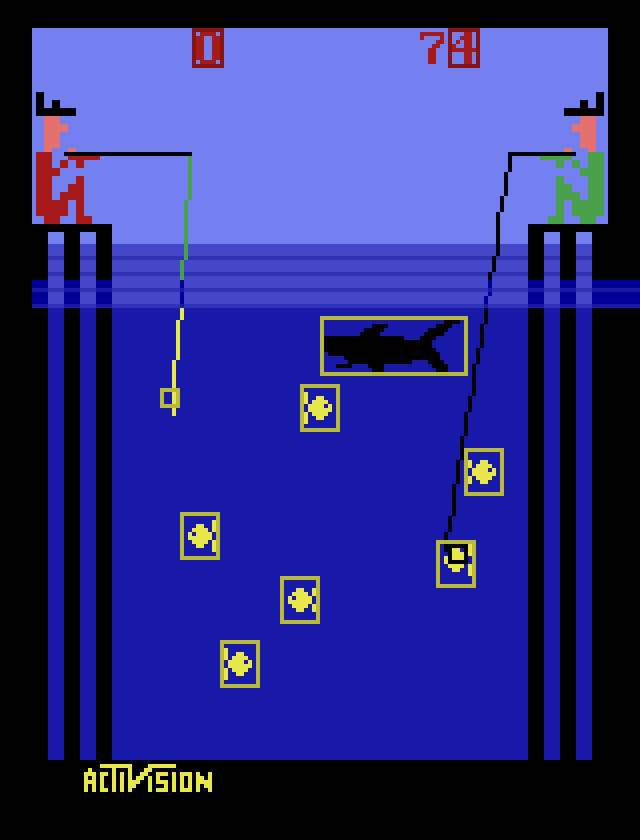}
\end{minipage}{}
}{}

\foreach \n in {FishingDerby}{\input{reports/\n_stat_reports.tex}}

\subsection{Freeway \ details}
\IfFileExists{environment_description/Freeway.tex}{
\begin{minipage}{0.60\textwidth}
Your objective is to guide your chicken across lane after lane of busy rush hour traffic. You receive a point for every chicken that makes it to the top of the screen after crossing all the lanes of traffic.
\end{minipage}
}{}
\IfFileExists{environment_images/Freeway.png}{
\begin{minipage}{0.38\textwidth}
\raggedleft
\includegraphics[scale=0.25]{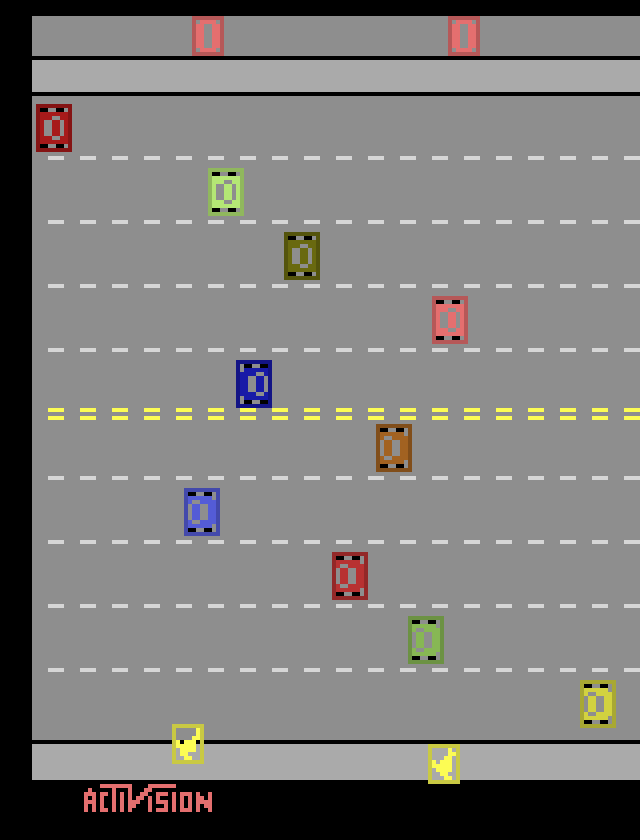}
\end{minipage}{}
}{}

\foreach \n in {Freeway}{\input{reports/\n_stat_reports.tex}}

\subsection{Frostbite \ details}
\IfFileExists{environment_description/Frostbite.tex}{
\begin{minipage}{0.60\textwidth}
In Frostbite, the player controls “Frostbite Bailey” who hops back and forth across across an Arctic river, changing the color of the ice blocks from white to blue. Each time he does so, a block is added to his igloo.
\end{minipage}
}{}
\IfFileExists{environment_images/Frostbite.png}{
\begin{minipage}{0.38\textwidth}
\raggedleft
\includegraphics[scale=0.25]{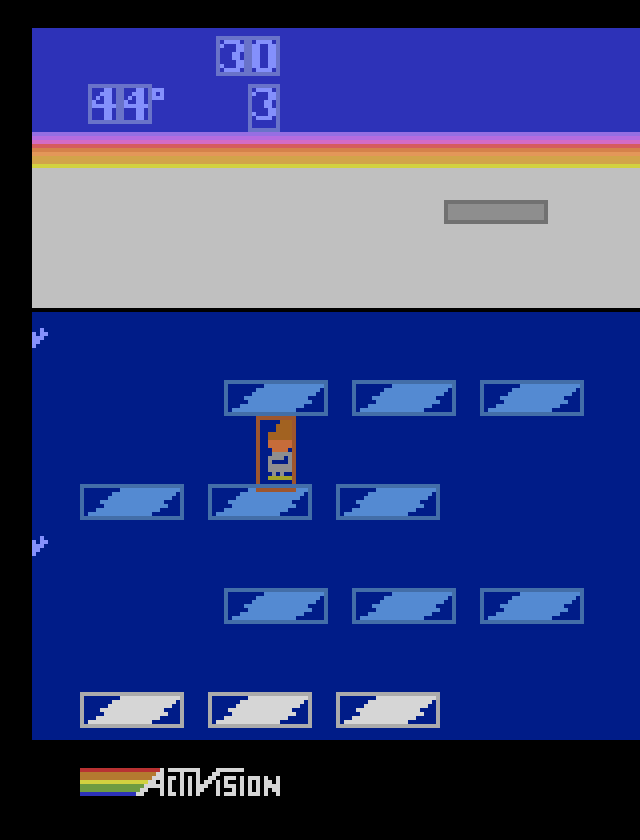}
\end{minipage}{}
}{}

\foreach \n in {Frostbite}{\input{reports/\n_stat_reports.tex}}

\subsection{Gopher \ details}
\IfFileExists{environment_description/Gopher.tex}{
\begin{minipage}{0.60\textwidth}
The player controls a shovel-wielding farmer who protects a crop of three carrots from a gopher.
\end{minipage}
}{}
\IfFileExists{environment_images/Gopher.png}{
\begin{minipage}{0.38\textwidth}
\raggedleft
\includegraphics[scale=0.25]{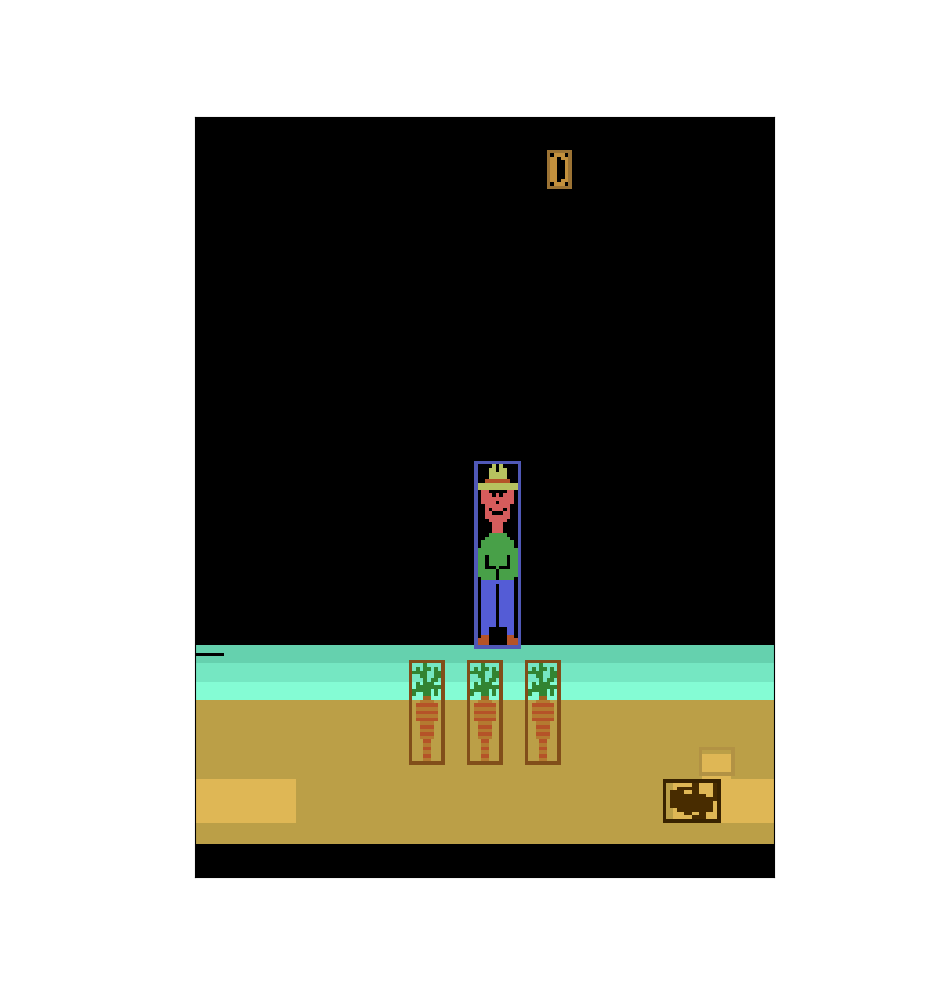}
\end{minipage}{}
}{}

\foreach \n in {Gopher}{\input{reports/\n_stat_reports.tex}}

\subsection{Hero \ details}
\IfFileExists{environment_description/Hero.tex}{
\begin{minipage}{0.60\textwidth}
You need to rescue miners that are stuck in a mine shaft. You have access to various tools: A propeller backpack that allows you to fly wherever you want, sticks of dynamite that can be used to blast through walls, a laser beam to kill vermin, and a raft to float across stretches of lava.You have a limited amount of power. Once you run out, you lose a live.
\end{minipage}
}{}
\IfFileExists{environment_images/Hero.png}{
\begin{minipage}{0.38\textwidth}
\raggedleft
\includegraphics[scale=0.25]{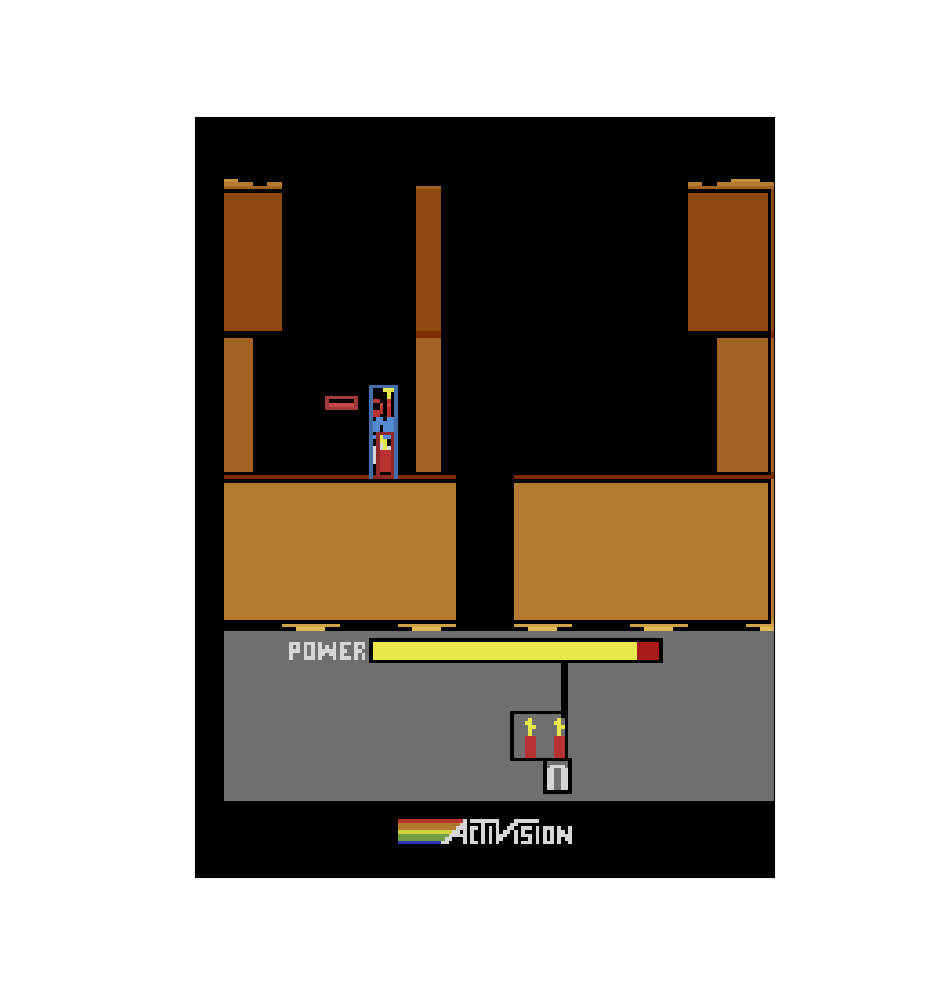}
\end{minipage}{}
}{}

\foreach \n in {Hero}{\input{reports/\n_stat_reports.tex}}

\subsection{IceHockey \ details}
\IfFileExists{environment_description/IceHockey.tex}{
\begin{minipage}{0.60\textwidth}
Your goal is to score as many points as possible in a standard game of Ice Hockey over a 3-minute time period. The ball is usually called “the puck”.There are 32 shot angles ranging from the extreme left to the extreme right. The angles can only aim towards the opponent’s goal.Just as in real hockey, you can pass the puck by shooting it off the sides of the rink. This can be really key when you’re in position to score a goal.
\end{minipage}
}{}
\IfFileExists{environment_images/IceHockey.png}{
\begin{minipage}{0.38\textwidth}
\raggedleft
\includegraphics[scale=0.25]{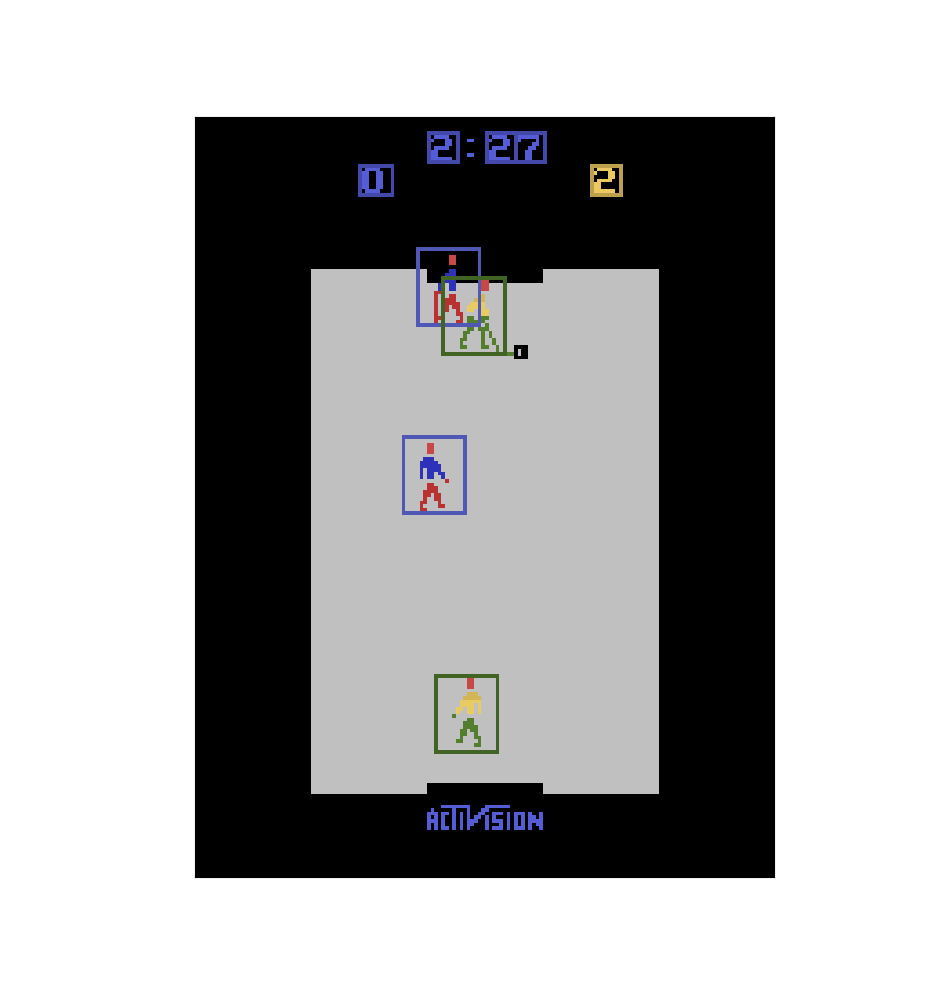}
\end{minipage}{}
}{}

\foreach \n in {IceHockey}{\input{reports/\n_stat_reports.tex}}

\subsection{Jamesbond \ details}
\IfFileExists{environment_description/Jamesbond.tex}{
\begin{minipage}{0.60\textwidth}
Your mission is to control Mr. Bond’s specially designed multipurpose craft to complete a variety of missions.The craft moves forward with a right motion and slightly back with a left motion.An up or down motion causes the craft to jump or dive.You can also fire by either lobbing a bomb to the bottom of the screen or firing a fixed angle shot to the top of the screen.
\end{minipage}
}{}
\IfFileExists{environment_images/Jamesbond.png}{
\begin{minipage}{0.38\textwidth}
\raggedleft
\includegraphics[scale=0.25]{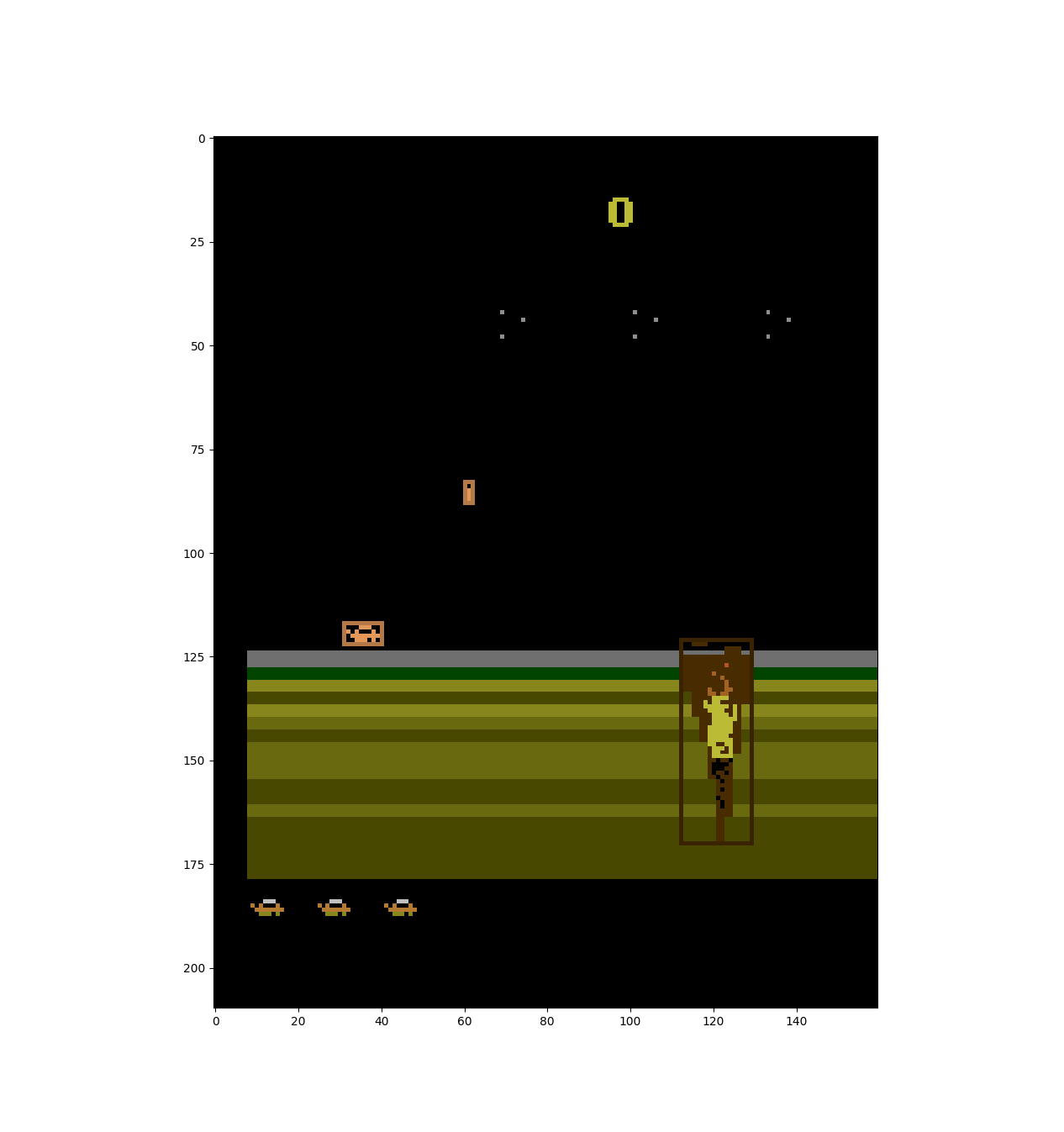}
\end{minipage}{}
}{}

\foreach \n in {Jamesbond}{\input{reports/\n_stat_reports.tex}}

\subsection{Kangaroo \ details}
\IfFileExists{environment_description/Kangaroo.tex}{
\begin{minipage}{0.60\textwidth}
The object of the game is to score as many points as you can while controlling Mother Kangaroo to rescue her precious baby. You start the game with three lives.During this rescue mission, Mother Kangaroo encounters many obstacles. You need to help her climb ladders, pick bonus fruit, and throw punches at monkeys.
\end{minipage}
}{}
\IfFileExists{environment_images/Kangaroo.png}{
\begin{minipage}{0.38\textwidth}
\raggedleft
\includegraphics[scale=0.25]{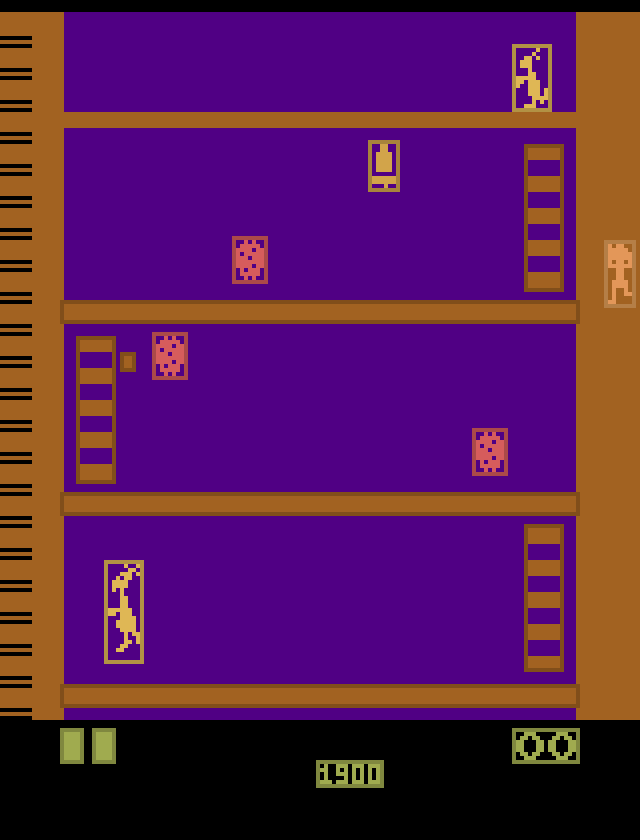}
\end{minipage}{}
}{}

\foreach \n in {Kangaroo}{\input{reports/\n_stat_reports.tex}}

\subsection{Krull \ details}
\IfFileExists{environment_description/Krull.tex}{
\begin{minipage}{0.60\textwidth}
Your mission is to find and enter the Beast’s Black Fortress, rescue Princess Lyssa, and destroy the Beast.The task is not an easy one, for the location of the Black Fortress changes with each sunrise on Krull.
\end{minipage}
}{}
\IfFileExists{environment_images/Krull.png}{
\begin{minipage}{0.38\textwidth}
\raggedleft
\includegraphics[scale=0.25]{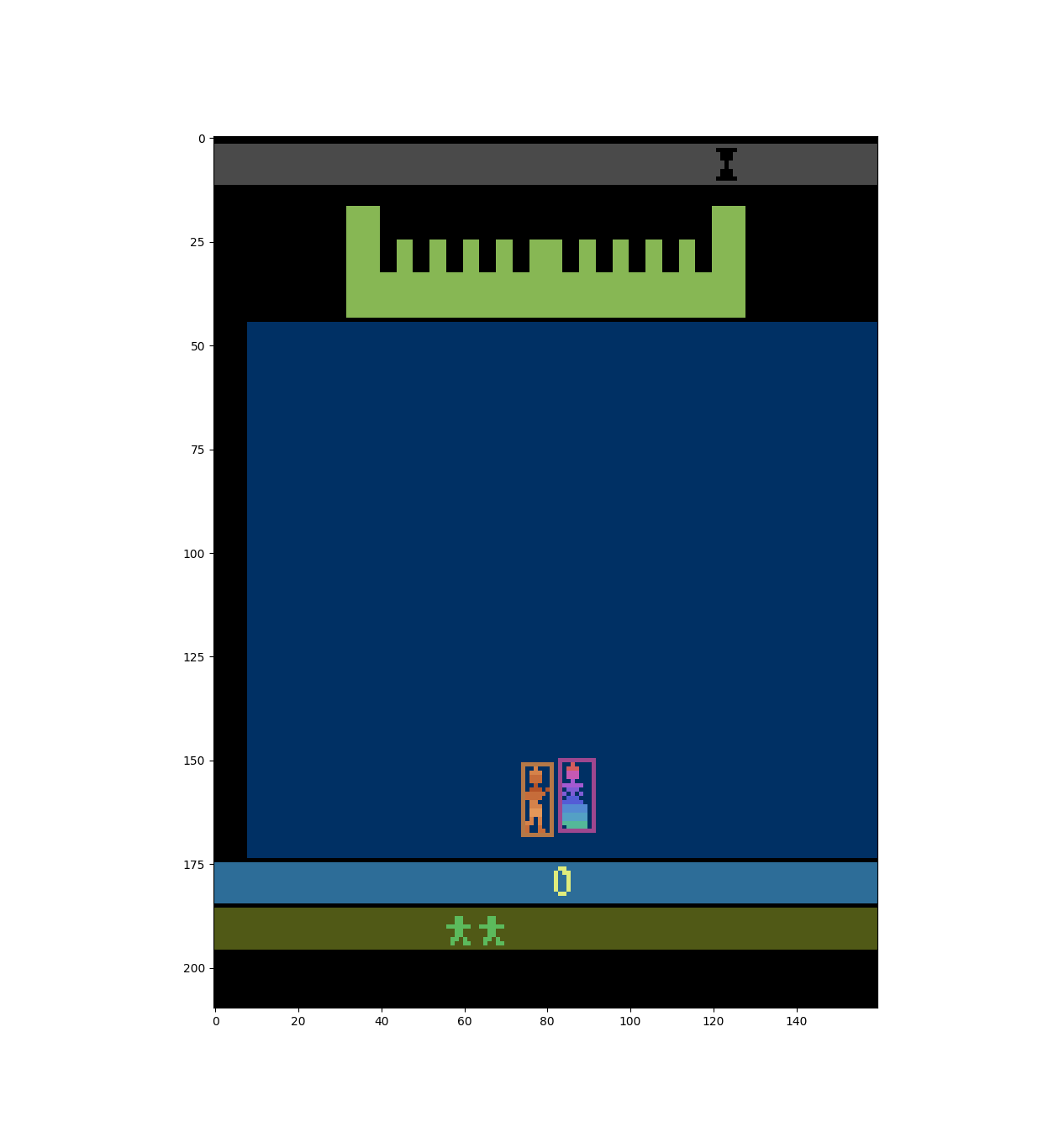}
\end{minipage}{}
}{}

\foreach \n in {Krull}{\input{reports/\n_stat_reports.tex}}

\subsection{MontezumaRevenge \ details}
\IfFileExists{environment_description/MontezumaRevenge.tex}{
\begin{minipage}{0.60\textwidth}
Your goal is to acquire Montezuma’s treasure by making your way through a maze of chambers within the emperor’s fortress. You must avoid deadly creatures while collecting valuables and tools which can help you escape with the treasure.
\end{minipage}
}{}
\IfFileExists{environment_images/MontezumaRevenge.png}{
\begin{minipage}{0.38\textwidth}
\raggedleft
\includegraphics[scale=0.25]{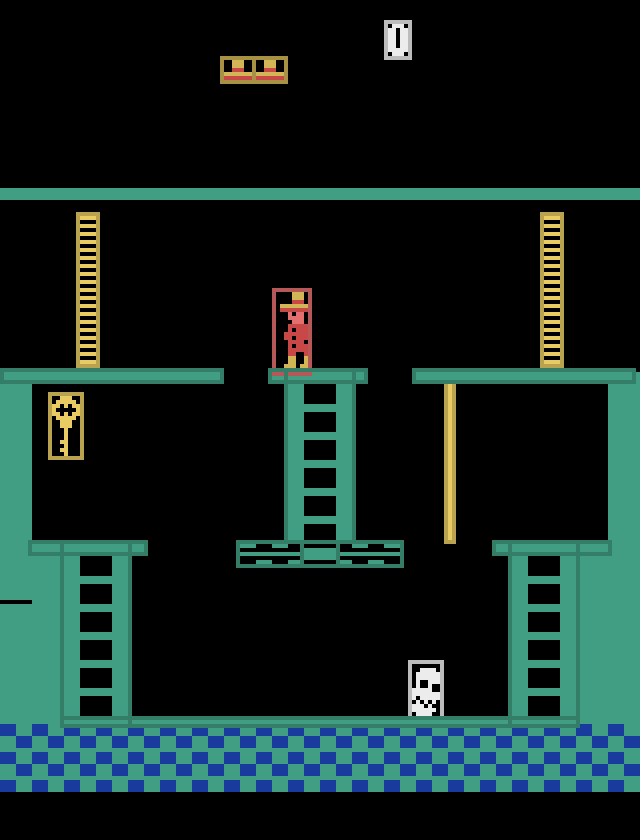}
\end{minipage}{}
}{}

\foreach \n in {MontezumaRevenge}{\input{reports/\n_stat_reports.tex}}

\subsection{MsPacman \ details}
\IfFileExists{environment_description/MsPacman.tex}{
\begin{minipage}{0.60\textwidth}
Your goal is to collect all of the pellets on the screen while avoiding the ghosts.
\end{minipage}
}{}
\IfFileExists{environment_images/MsPacman.png}{
\begin{minipage}{0.38\textwidth}
\raggedleft
\includegraphics[scale=0.25]{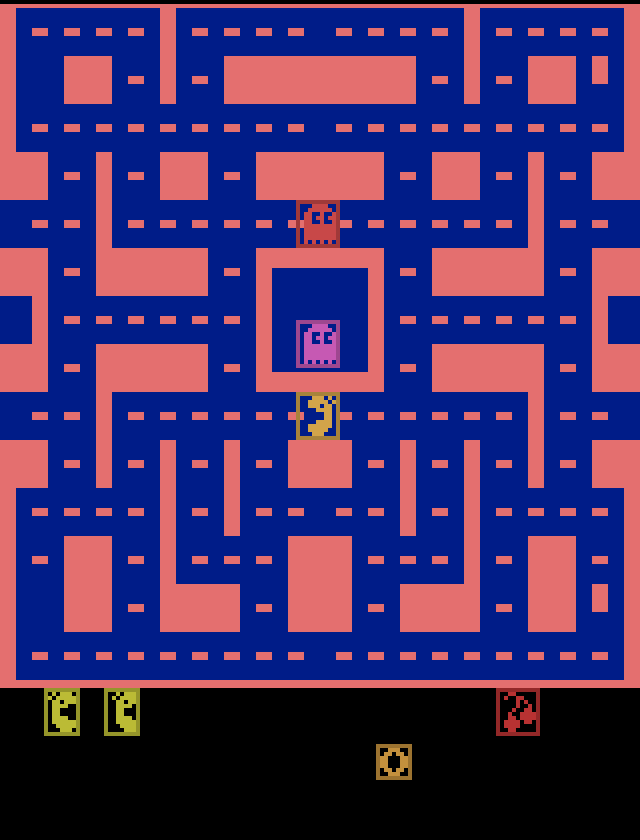}
\end{minipage}{}
}{}

\foreach \n in {MsPacman}{\input{reports/\n_stat_reports.tex}}

\subsection{Pacman \ details}
\IfFileExists{environment_description/Pacman.tex}{
\begin{minipage}{0.60\textwidth}
A classic arcade game. Move Pac Man around a maze collecting food and avoiding ghosts- unless you eat a Power Pellet, then you can eat the ghosts too!
\end{minipage}
}{}
\IfFileExists{environment_images/Pacman.png}{
\begin{minipage}{0.38\textwidth}
\raggedleft
\includegraphics[scale=0.25]{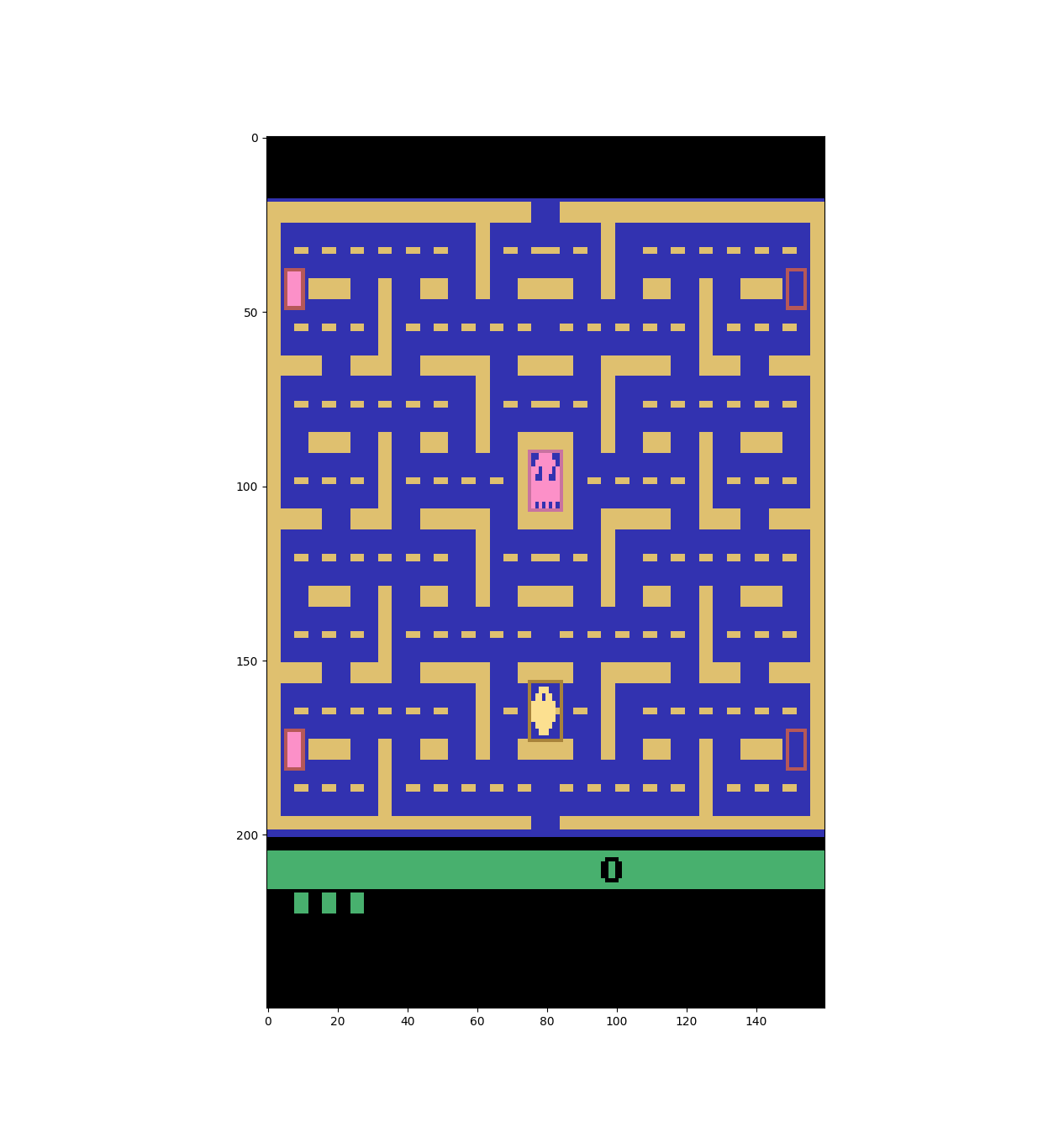}
\end{minipage}{}
}{}

\foreach \n in {Pacman}{\input{reports/\n_stat_reports.tex}}

\subsection{Pitfall \ details}
\IfFileExists{environment_description/Pitfall.tex}{
\begin{minipage}{0.60\textwidth}
You control Pitfall Harry and are tasked with collecting all the treasures in a jungle within 20 minutes. You have three lives. The game is over if you collect all the treasures or if you die or if the time runs out.
\end{minipage}
}{}
\IfFileExists{environment_images/Pitfall.png}{
\begin{minipage}{0.38\textwidth}
\raggedleft
\includegraphics[scale=0.25]{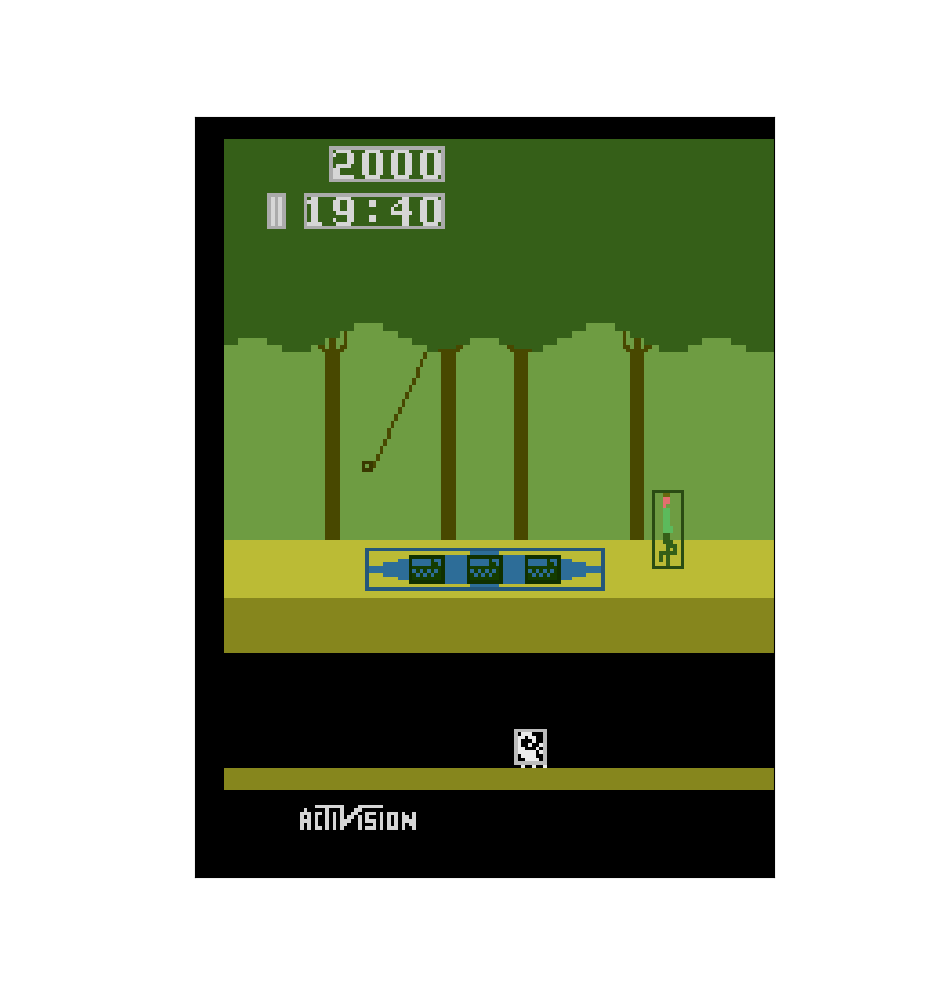}
\end{minipage}{}
}{}

\foreach \n in {Pitfall}{\input{reports/\n_stat_reports.tex}}

\subsection{Pong \ details}
\IfFileExists{environment_description/Pong.tex}{
\begin{minipage}{0.60\textwidth}
You control the right paddle, you compete against the left paddle controlled by the computer. You each try to keep deflecting the ball away from your goal and into your opponent’s goal.
\end{minipage}
}{}
\IfFileExists{environment_images/Pong.png}{
\begin{minipage}{0.38\textwidth}
\raggedleft
\includegraphics[scale=0.25]{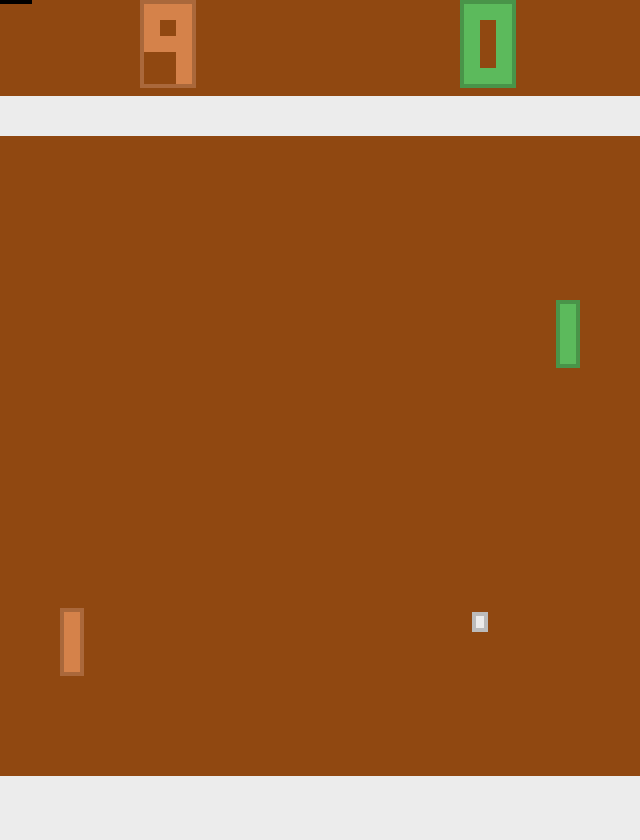}
\end{minipage}{}
}{}

\foreach \n in {Pong}{\input{reports/\n_stat_reports.tex}}

\subsection{PrivateEye \ details}
\IfFileExists{environment_description/PrivateEye.tex}{
\begin{minipage}{0.60\textwidth}
You control the French Private Eye Pierre Touche. Navigate the city streets, parks, secret passages, dead-ends and one-ways in search of the ringleader, Henri Le Fiend and his gang. You also need to find evidence and stolen goods that are scattered about. There are five cases, complete each case before its statute of limitations expires.
\end{minipage}
}{}
\IfFileExists{environment_images/PrivateEye.png}{
\begin{minipage}{0.38\textwidth}
\raggedleft
\includegraphics[scale=0.25]{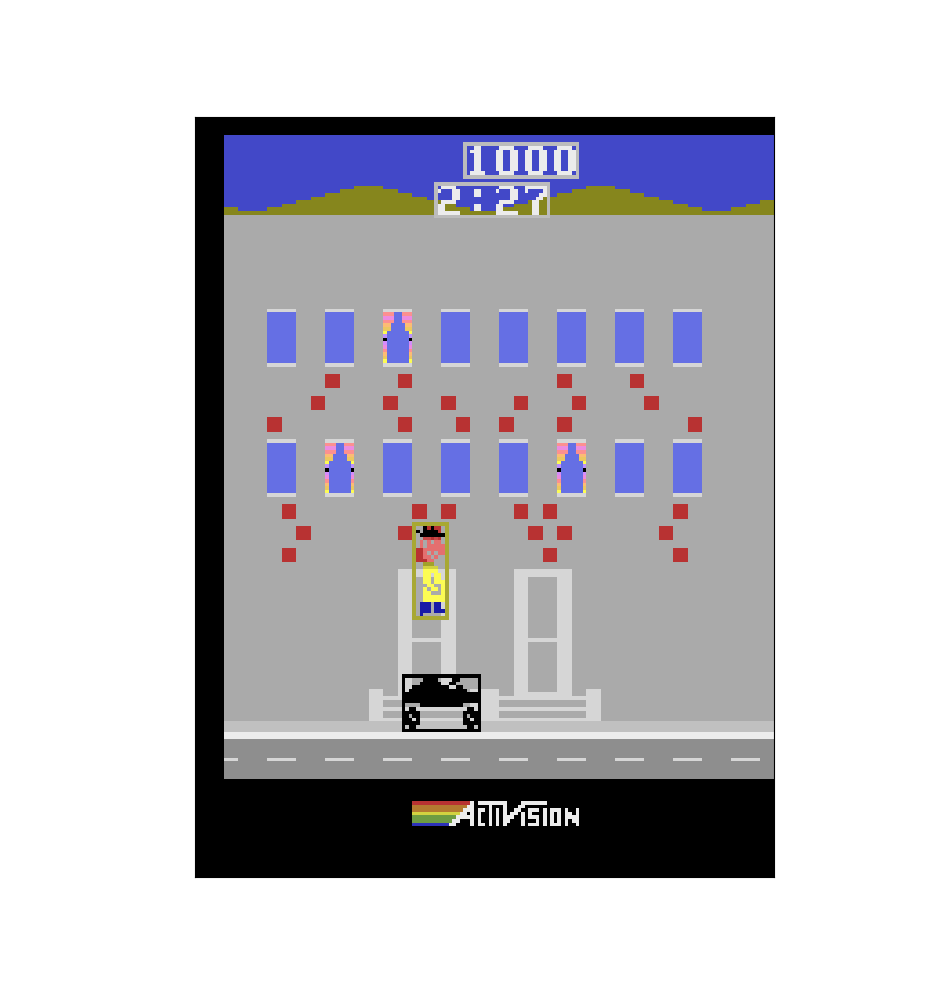}
\end{minipage}{}
}{}

\foreach \n in {PrivateEye}{\input{reports/\n_stat_reports.tex}}

\subsection{Qbert \ details}
\IfFileExists{environment_description/Qbert.tex}{
\begin{minipage}{0.60\textwidth}
You are Q*bert. Your goal is to change the color of all the cubes on the pyramid to the pyramid’s ‘destination’ color. To do this, you must hop on each cube on the pyramid one at a time while avoiding nasty creatures that lurk there.
\end{minipage}
}{}
\IfFileExists{environment_images/Qbert.png}{
\begin{minipage}{0.38\textwidth}
\raggedleft
\includegraphics[scale=0.25]{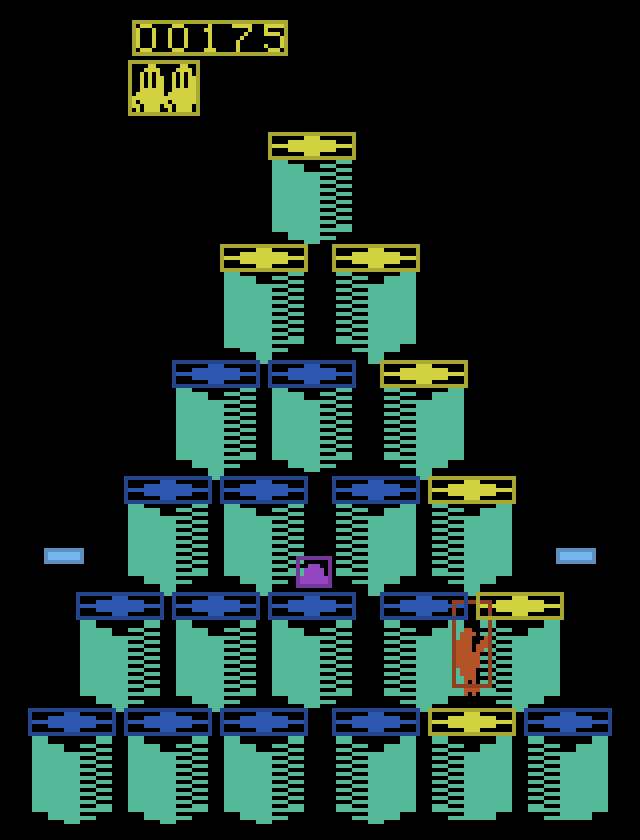}
\end{minipage}{}
}{}

\foreach \n in {Qbert}{\input{reports/\n_stat_reports.tex}}

\subsection{Riverraid \ details}
\IfFileExists{environment_description/Riverraid.tex}{
\begin{minipage}{0.60\textwidth}
You control a jet that flies over a river: you can move it sideways and fire missiles to destroy enemy objects. Each time an enemy object is destroyed you score points (i.e. rewards). You lose a jet when you run out of fuel: fly over a fuel depot when you begin to run low.You lose a jet even when it collides with the river bank or one of the enemy objects (except fuel depots). The game begins with a squadron of three jets in reserve and you’re given an additional jet (up to 9) for each 10,000 points you score.
\end{minipage}
}{}
\IfFileExists{environment_images/Riverraid.png}{
\begin{minipage}{0.38\textwidth}
\raggedleft
\includegraphics[scale=0.25]{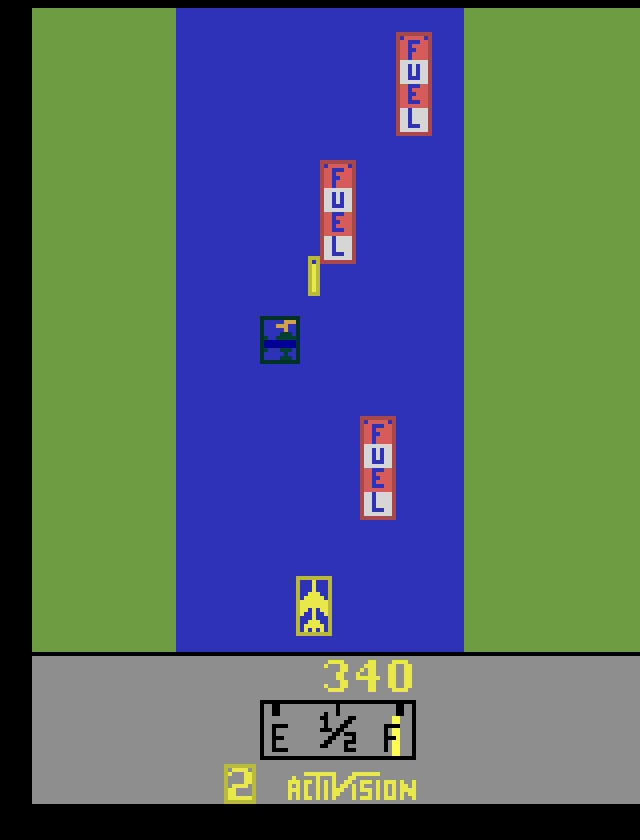}
\end{minipage}{}
}{}

\foreach \n in {Riverraid}{\input{reports/\n_stat_reports.tex}}

\subsection{RoadRunner \ details}
\IfFileExists{environment_description/RoadRunner.tex}{
\begin{minipage}{0.60\textwidth}
You control the Road Runner(TM) in a race; you can control the direction to run in and times to jumps.The goal is to outrun Wile E. Coyote(TM) while avoiding the hazards of the desert.The game begins with three lives. You lose a life when the coyote catches you, picks you up in a rocket, or shoots you with a cannon. You also lose a life when a truck hits you, you hit a land mine, you fall off a cliff,or you get hit by a falling rock.You score points (i.e. rewards) by eating seeds along the road, eating steel shot, and destroying the coyote.
\end{minipage}
}{}
\IfFileExists{environment_images/RoadRunner.png}{
\begin{minipage}{0.38\textwidth}
\raggedleft
\includegraphics[scale=0.25]{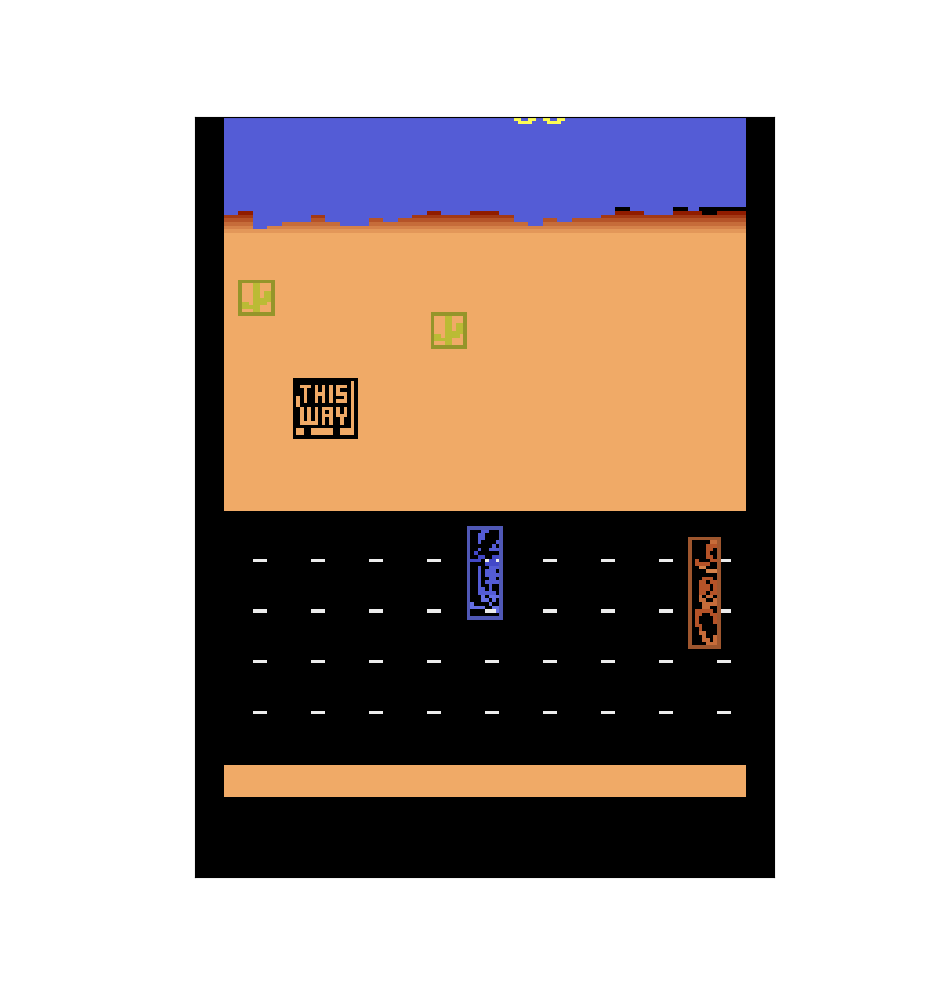}
\end{minipage}{}
}{}

\foreach \n in {RoadRunner}{\input{reports/\n_stat_reports.tex}}

\subsection{Seaquest \ details}
\IfFileExists{environment_description/Seaquest.tex}{
\begin{minipage}{0.60\textwidth}
You control a sub able to move in all directions and fire torpedoes. The goal is to retrieve as many divers as you can, while dodging and blasting enemy subs and killer sharks; points will be awarded accordingly. The game begins with one sub and three waiting on the horizon. Each time you increase your score by 10,000 points, an extra sub will be delivered to your base. You can only have six reserve subs on the screen at one time.Your sub will explode if it collides with anything except your own divers.The sub has a limited amount of oxygen that decreases at a constant rate during the game. When the oxygen tank is almost empty, you need to surface and if you don’t do it in time, your sub will blow up and you’ll lose one diver. Each time you’re forced to surface, with less than six divers, you lose one diver as well.
\end{minipage}
}{}
\IfFileExists{environment_images/Seaquest.png}{
\begin{minipage}{0.38\textwidth}
\raggedleft
\includegraphics[scale=0.25]{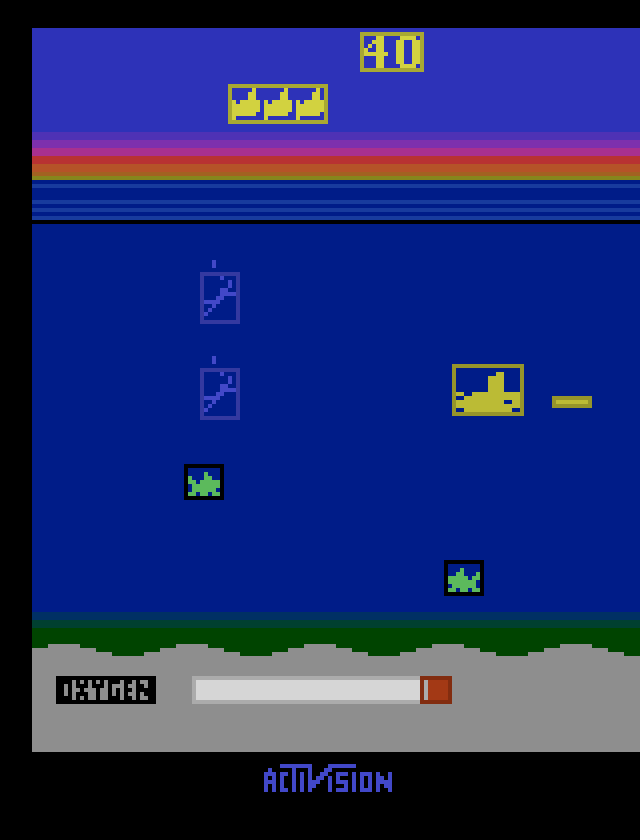}
\end{minipage}{}
}{}

\foreach \n in {Seaquest}{\input{reports/\n_stat_reports.tex}}

\subsection{Skiing \ details}
\IfFileExists{environment_description/Skiing.tex}{
\begin{minipage}{0.60\textwidth}
You control a skier who can move sideways. The goal is to run through all gates (between the poles) in the fastest time. You are penalized five seconds for each gate you miss. If you hit a gate or a tree, your skier will jump back up and keep going.
\end{minipage}
}{}
\IfFileExists{environment_images/Skiing.png}{
\begin{minipage}{0.38\textwidth}
\raggedleft
\includegraphics[scale=0.25]{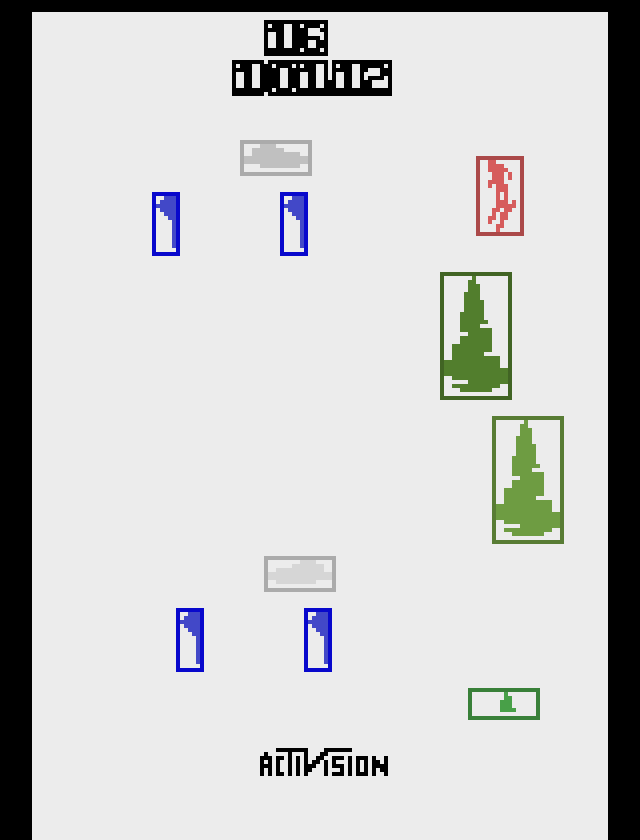}
\end{minipage}{}
}{}

\foreach \n in {Skiing}{\input{reports/\n_stat_reports.tex}}

\subsection{SpaceInvaders \ details}
\IfFileExists{environment_description/SpaceInvaders.tex}{
\begin{minipage}{0.60\textwidth}
Your objective is to destroy the space invaders by shooting your laser cannon at them before they reach the Earth. The game ends when all your lives are lost after taking enemy fire, or when they reach the earth.
\end{minipage}
}{}
\IfFileExists{environment_images/SpaceInvaders.png}{
\begin{minipage}{0.38\textwidth}
\raggedleft
\includegraphics[scale=0.25]{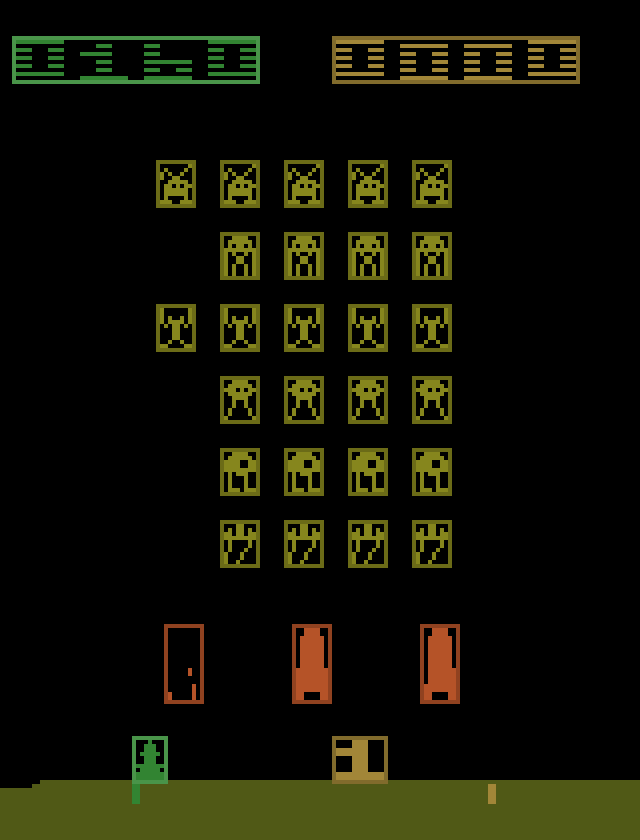}
\end{minipage}{}
}{}

\foreach \n in {SpaceInvaders}{\input{reports/\n_stat_reports.tex}}

\subsection{Tennis \ details}
\IfFileExists{environment_description/Tennis.tex}{
\begin{minipage}{0.60\textwidth}
You control the orange player playing against a computer-controlled blue player. The game follows the rules of tennis. The first player to win at least 6 games with a margin of at least two games wins the match. If the score is tied at 6-6, the first player to go 2 games up wins the match.
\end{minipage}
}{}
\IfFileExists{environment_images/Tennis.png}{
\begin{minipage}{0.38\textwidth}
\raggedleft
\includegraphics[scale=0.25]{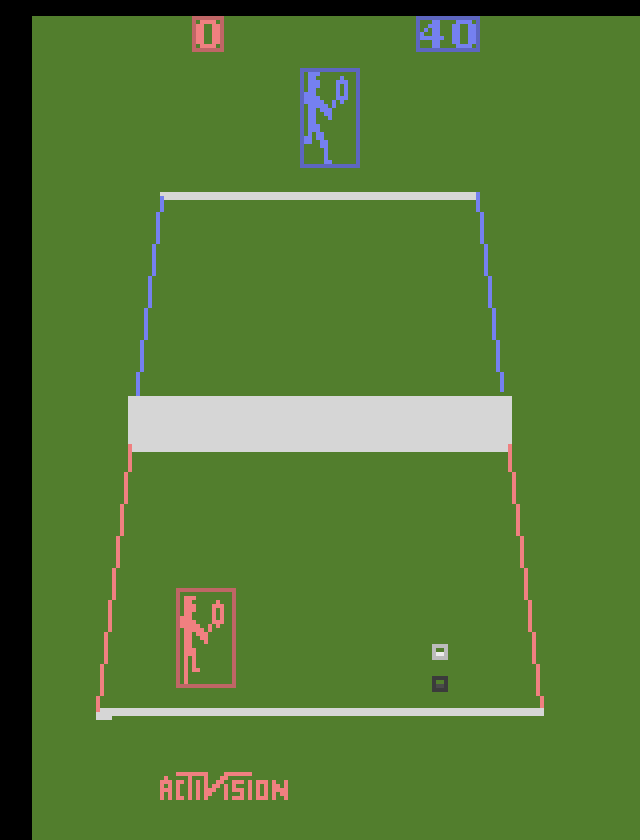}
\end{minipage}{}
}{}

\foreach \n in {Tennis}{\input{reports/\n_stat_reports.tex}}

\subsection{TimePilot \ details}
\IfFileExists{environment_description/TimePilot.tex}{
\begin{minipage}{0.60\textwidth}
You control an aircraft. Use it to destroy your enemies. As you progress in the game, you encounter enemies with technology that is increasingly from the future.
\end{minipage}
}{}
\IfFileExists{environment_images/TimePilot.png}{
\begin{minipage}{0.38\textwidth}
\raggedleft
\includegraphics[scale=0.25]{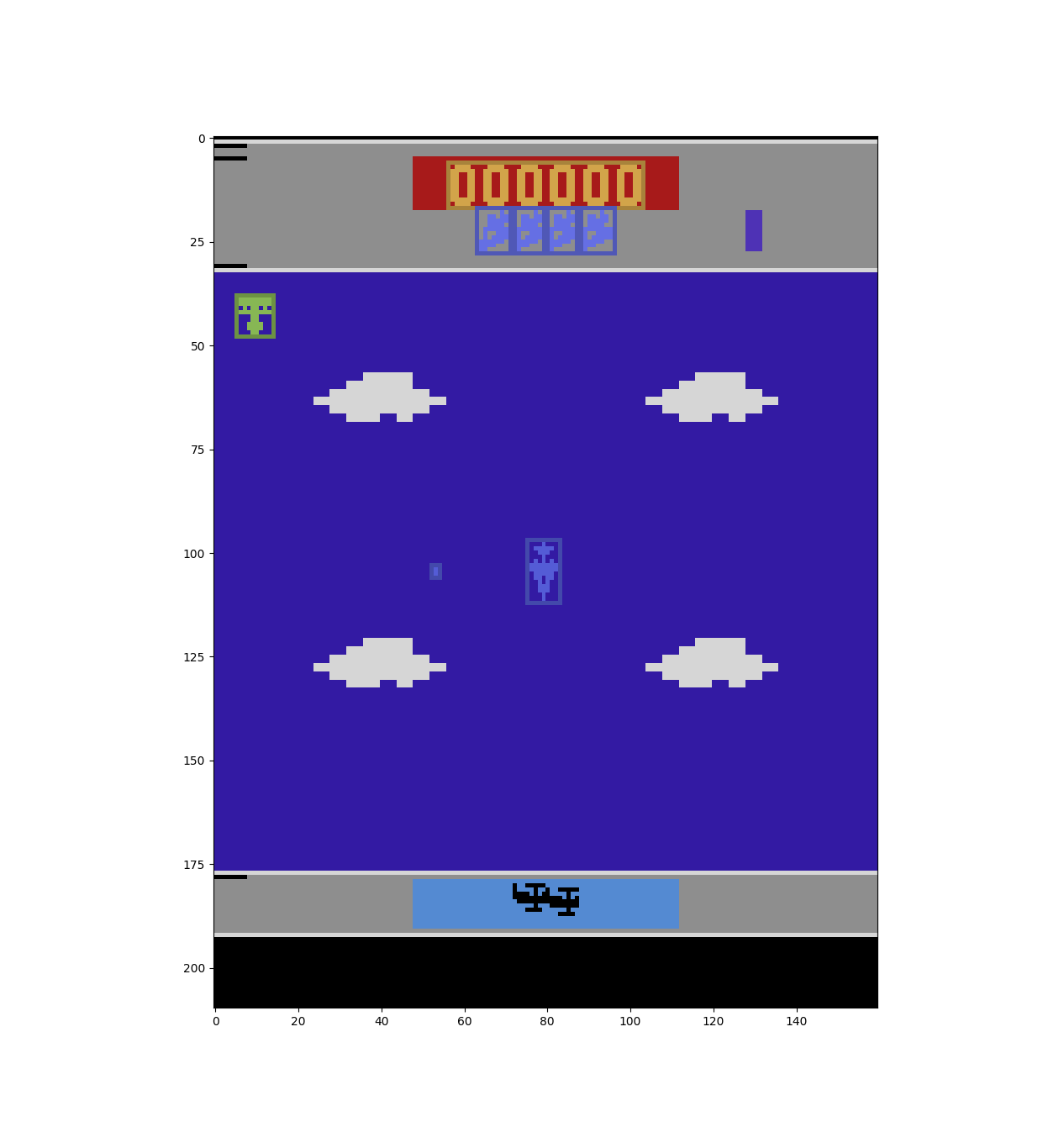}
\end{minipage}{}
}{}

\foreach \n in {TimePilot}{\input{reports/\n_stat_reports.tex}}

\subsection{UpNDown \ details}
\IfFileExists{environment_description/UpNDown.tex}{
\begin{minipage}{0.60\textwidth}
Your goal is to steer your baja bugger to collect prizes and eliminate opponents.
\end{minipage}
}{}
\IfFileExists{environment_images/UpNDown.png}{
\begin{minipage}{0.38\textwidth}
\raggedleft
\includegraphics[scale=0.25]{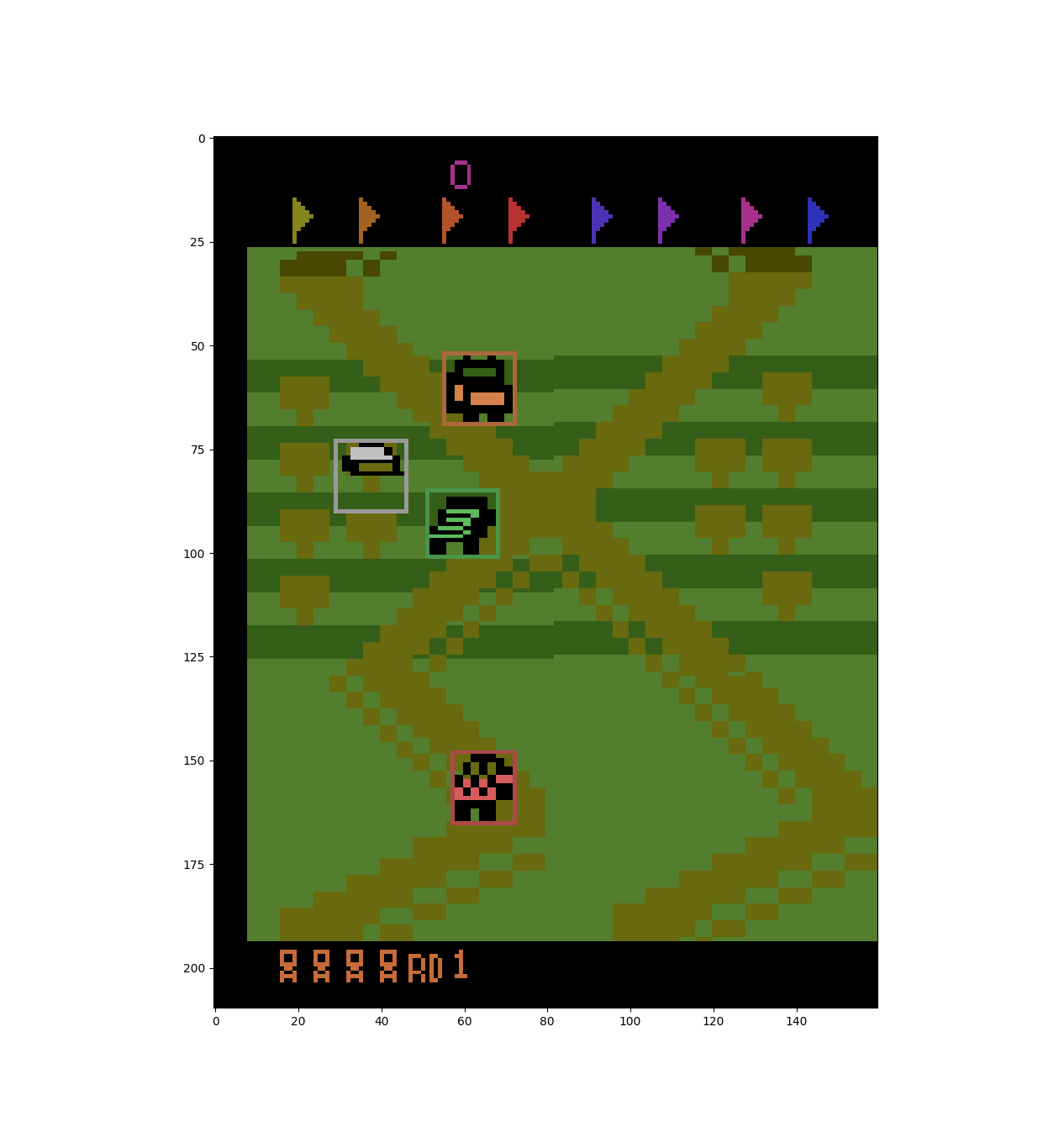}
\end{minipage}{}
}{}

\foreach \n in {UpNDown}{\input{reports/\n_stat_reports.tex}}

\subsection{Venture \ details}
\IfFileExists{environment_description/Venture.tex}{
\begin{minipage}{0.60\textwidth}
Your goal is to capture the treasure in every chamber of the dungeon while eliminating the monsters.
\end{minipage}
}{}
\IfFileExists{environment_images/Venture.png}{
\begin{minipage}{0.38\textwidth}
\raggedleft
\includegraphics[scale=0.25]{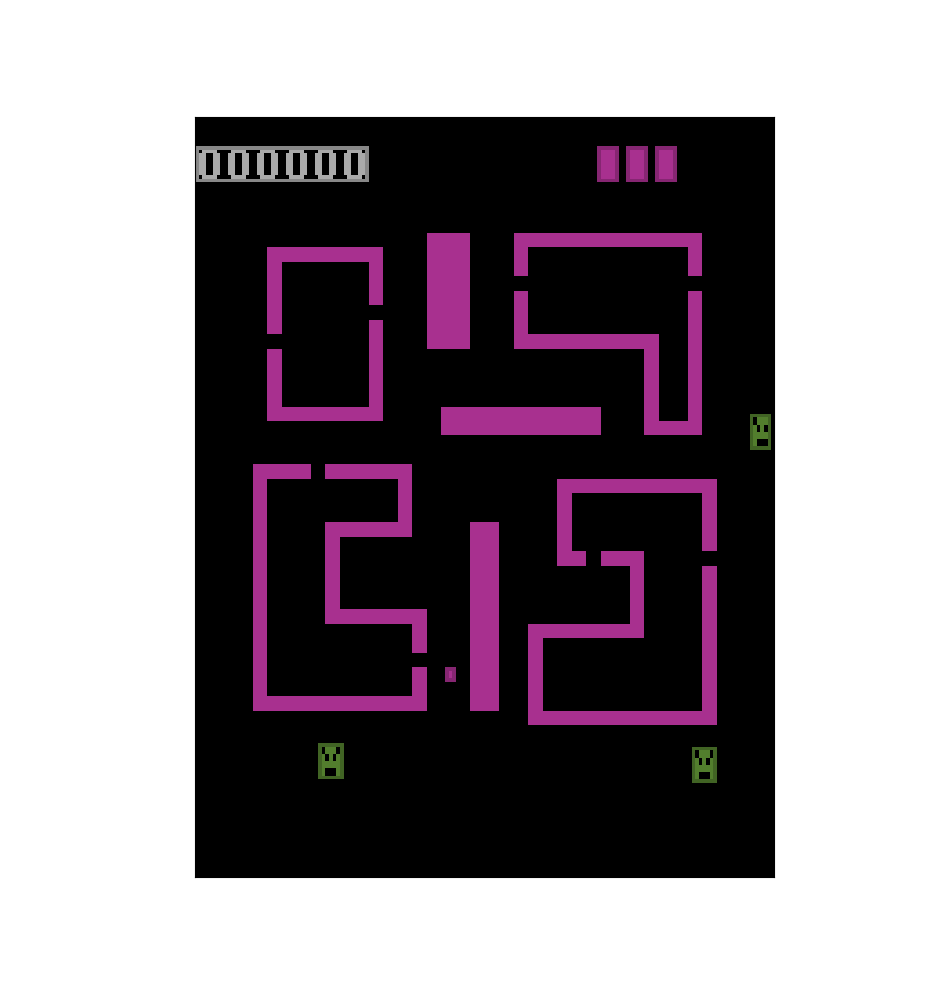}
\end{minipage}{}
}{}

\foreach \n in {Venture}{\input{reports/\n_stat_reports.tex}}

\subsection{VideoPinball \ details}
\IfFileExists{environment_description/VideoPinball.tex}{
\begin{minipage}{0.60\textwidth}
Your goal is to keep the ball in play as long as possible and to score as many points as possible.
\end{minipage}
}{}
\IfFileExists{environment_images/VideoPinball.png}{
\begin{minipage}{0.38\textwidth}
\raggedleft
\includegraphics[scale=0.25]{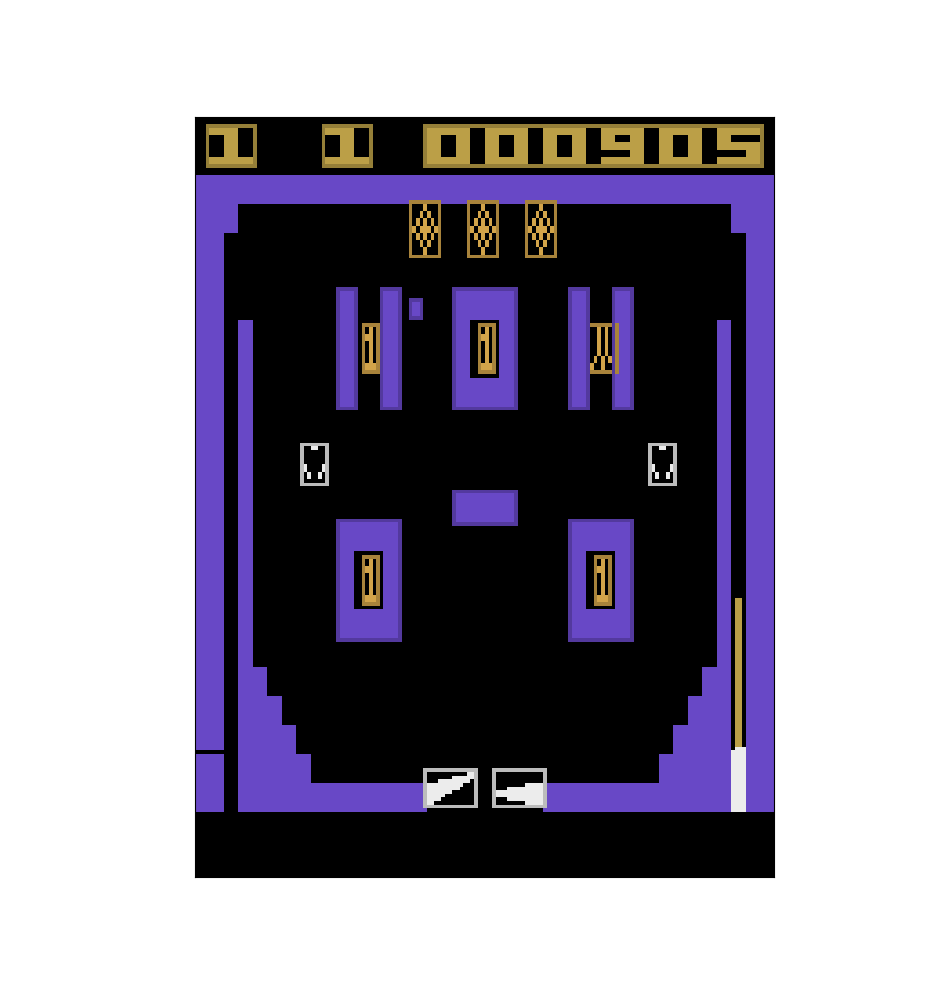}
\end{minipage}{}
}{}

\foreach \n in {VideoPinball}{\input{reports/\n_stat_reports.tex}}

\subsection{YarsRevenge \ details}
\IfFileExists{environment_description/YarsRevenge.tex}{
\begin{minipage}{0.60\textwidth}
The objective is to break a path through the shield and destroy the Qotile with a blast from the Zorlon Cannon.
\end{minipage}
}{}
\IfFileExists{environment_images/YarsRevenge.png}{
\begin{minipage}{0.38\textwidth}
\raggedleft
\includegraphics[scale=0.25]{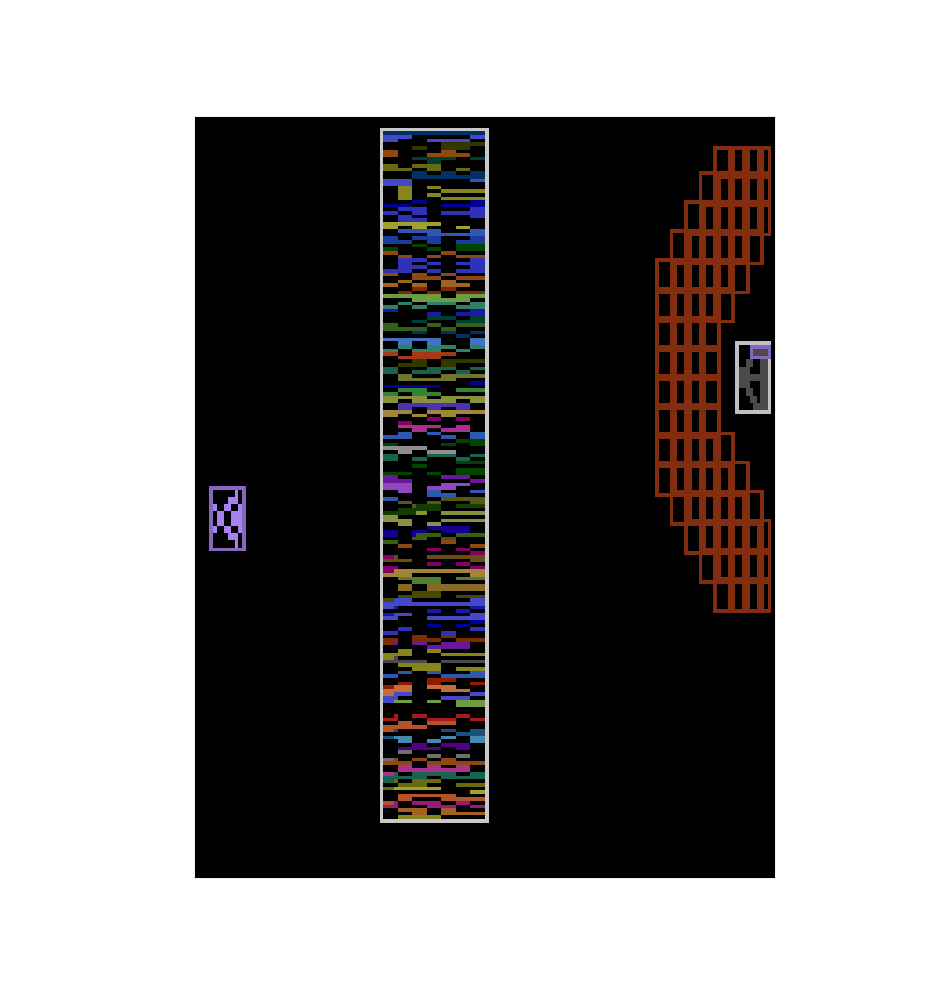}
\end{minipage}{}
}{}

\foreach \n in {YarsRevenge}{\input{reports/\n_stat_reports.tex}}

\clearpage

\rebut{
\section{Common Mistakes in Extracting and Detecting Objects}
}
\begin{figure}[h!]
\begin{subfigure}[c]{0.3\textwidth}
\includegraphics[width=\textwidth]{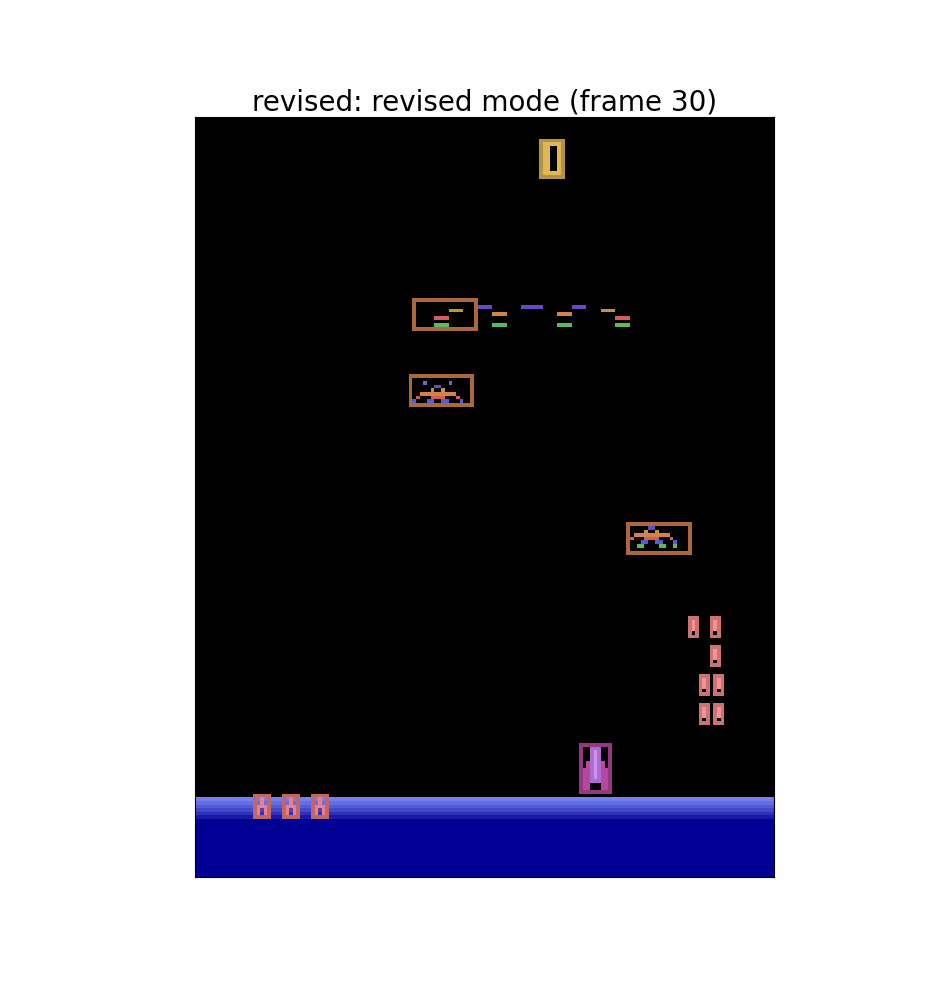}
\end{subfigure}
\begin{subfigure}[c]{0.3\textwidth}
\includegraphics[width=\textwidth]{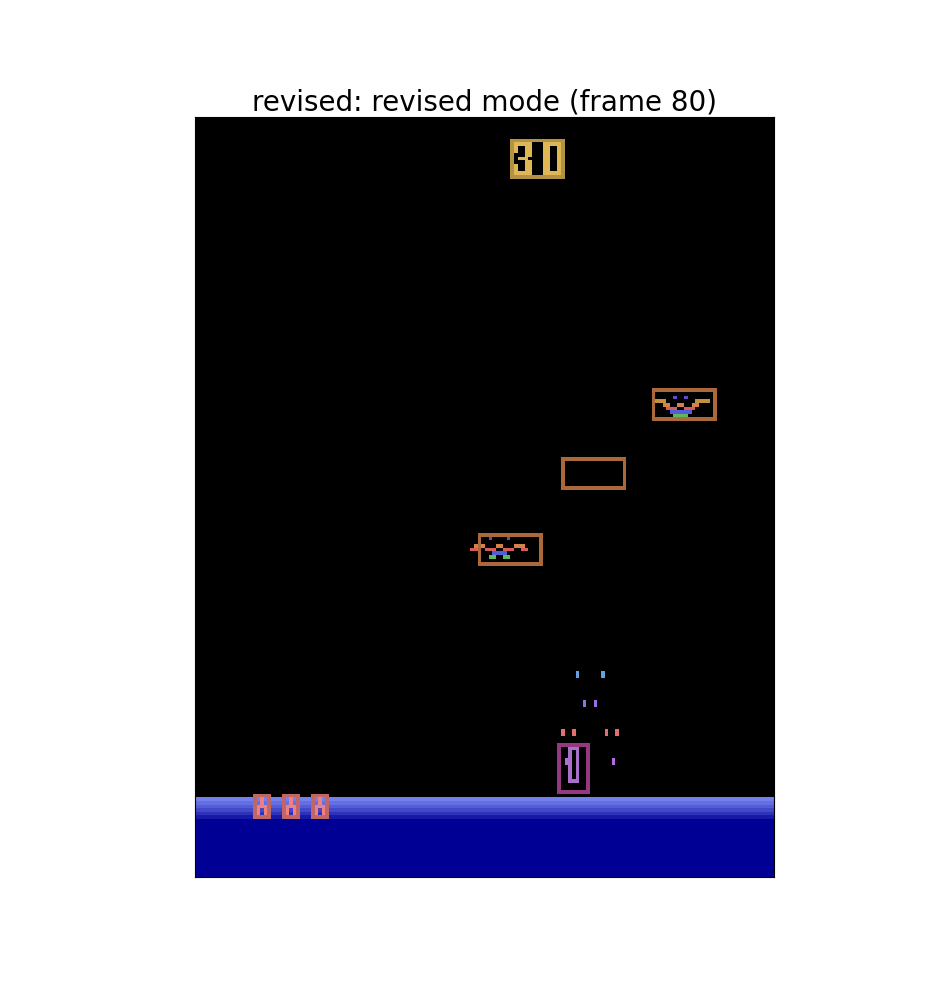}
\end{subfigure}
\begin{subfigure}[c]{0.3\textwidth}
\includegraphics[width=\textwidth]{environment_images/YarsRevenge.png}
\end{subfigure}
\caption{\rebut{Animation and errors in the game of DemonAttack and YarsRevenge. We can see multiple particle effects and invisible objects. In the left we see the spawn animation of an enemy, i n the second image we see the death animation of the player and in the last we see the invisible shields in YarsRevenge. In all cases the objects are already detected even if it is not yet or not anymore visible to the player.}}
\label{fig:demonAttackAnimation}
\end{figure}
\rebut{
In this section, we will briefly discuss 2 common errors that can occur during detection and extraction based on the games DemonAttack and YarsRevenge. 
}

\rebut{
\textbf{Case 1: Particle effects.} As described in Section 2, we primarily use positional information and the change of colors to identify objects in the visual detection of objects (VEM). It can happen that particle effects are incorrectly identified as objects, see Figure~\ref{fig:demonAttackAnimation}. In our RAM extraction we have defined the number and types of objects before extraction and concentrate on all game elements that are relevant for the game. Since these particle effects have no effect on the game, we deliberately do not detect them, which leads to a higher errors in F1 and IOU. \\
}

\rebut{
\textbf{Case 2: Invisible objects.} If objects disappear or appear in a game, there are various ways to realize this. The most common and simplest method, which is also used in most games, is to initialize objects only when they appear and to clear the memory when objects disappear. However, some games, such as DemonAttack or YarsRevenge  (Fig.~\ref{fig:demonAttackAnimation}) use a different method. Here the objects are only set to invisible when they disappear or already exist before the objects appear. As such, these objects are also found and tracked by our REM method at an early stage, even though they have not yet appeared, which leads to an increased error. In many games we have therefore tried to find binary information about which objects are active so that those that are not, are not detected. This helps to minimize the error and increase the scores, as you can see in the updates scores in DemonAttack. \\
}

\clearpage

\section{Difference between AtariARI and OCAtari}
\begin{table}[h]
    \centering
    \begin{tabular}{lp{6.5cm}p{7cm}}
    \toprule
        Game & Objects (AtariARI) & Objects (OCAtari) \\
        \midrule
        Asterix & Enemies, Player, Lives, Score, Missiles & Enemies, Player, Lives, Score, Missiles  \\ \midrule
        Berzerk & Player, Missiles, Lives, \textbf{Killcount}, Level, \textbf{Evil Otto}, Enemies & \textbf{Logo}, Player, Missiles, Enemies, Score, \textbf{RoomCleared}  \\ \midrule
        Bowling & Ball, Player, \textbf{FrameNumber}, Pins, Score & Pins, Player, PlayerScore, \textbf{PlayerRound}, \textbf{Player2Round}, Ball \\ \midrule
        Boxing & Player, Enemy, Scores, Clock & Enemy, Player, Scores, Clock, \textbf{Logo}  \\ \midrule
        Breakout & Ball, Player, Blocks, Score & Player, Blocks, \textbf{Live}, Score, Ball \\ \midrule
        Freeway & Player, Score, Cars & Player, Score, Cars, \textbf{Chicken} \\ \midrule
        Frostbite & Ice blocks, Lives, Igloo, Enemies, Player, Score & Ice blocks Blue, Ice blocks White, Score, Player Lives, Igloo, Enemies \\ \midrule
        Montezumas R. & \textbf{RoomNr}, Player, Skull, Monster, Level, Lives, \textbf{ItemsInInventory}, \textbf{RoomState}, Score & Player, Lives, Skull, Barrier, Key, Score, \textbf{Rope} \\ \midrule
        MsPacman & Enemies, Player, Fruits, \textbf{GhostsCount}, \textbf{DotsEaten}, Score, Lives & Lives, Score, Player, Enemies, Fruits \\ \midrule
        Pong & Player, Enemy, Ball, Scores & Player, Enemy, Ball, Scores \\ \midrule
        PrivateEye & Player, RoomNr, Clock, Score, Dove & \\ \midrule
        Q*Bert & Player, PlayerColumn, Red Enemy, Green Enemy, Score, TileColors & Cubes/Tiles, Score, Lives, \textbf{Disks}, Player, Sam, \textbf{PurpleBall}, \textbf{Coily}, GreenBall \\ \midrule
        Riverraid & Player, Missile, FuelMeter & \textbf{Score}, FuelMeter, \textbf{Tanker}, Lives, Player, \textbf{Helicopter}, Missile, \textbf{Bridge}, \textbf{Jet} \\ \midrule
        Seaquest & Enemy, Player, EnemyMissile, PlayerMissile, Score, Lives, DiversCount & Player, Lives, \textbf{OxygenBar}, Score, \textbf{Divers}, PlayerMissile, Enemy, EnemyMissile, DiverCount \\ \midrule
        SpaceInvaders & \textbf{InvadersCount}, Score, Lives, Player, Enemies, Missiles & Score, Lives, Player, Enemies, Missiles, \textbf{Satellite}, \textbf{Shield} \\ \midrule
        Tennis & Enemy, Scores, Ball, Player & Enemy, Scores, Ball, \textbf{BallShadow}, Player, Logo \\

        \bottomrule
    \end{tabular}
    \vspace{2mm}
    \caption{All games, supported by both AtariARI and OCAtari with their respective object lists. Note that OCAtari returns a list of (x,y,w,h) per object and AtariARI provides the value written at a specific RAM position (x and y positions or the direct value, \eg, scores and so on)}
    \label{tab:obj}
\end{table}
\clearpage
 
\section{Insufficent Information in AtariARI}
\label{app:insufficent}
\begin{table}[h]
    \centering
    \begin{tabular}{lp{10cm}}
    \toprule
        Game & Reason \\
        \midrule
        Battlezone\footnotemark[1] & Unfinished \\
        DemonAttack & Not all Demons are spotted  \\
        Hero & Missing Enemies \\
        Q*Bert & Some Enemies, like Coily (Snake) are missing \\
        Skiing\footnotemark[1] & Unfinished \\
        RiverRaid & Important Elements (see above) are missing \\
        Seaquest & Oxygenbar, Divers are missing \\
        SpaceInvaders & Shields are missing \\
        \bottomrule
    \end{tabular}
    \vspace{2mm}
    \caption{In \autoref{tab:supported games} some games are marked with a $\sim$ to show that the RAM information provided by AtariARI are insufficient. This table gives a short reason while we marked each game.}
    \label{tab:check}
\end{table}
\footnotetext[1]{The games appear in the Github for AtariARI, but not in the associated publication~\citep{Anand19AtariARI}. Also, the information does not seem sufficient to play with them alone so we did not indicate these games in \autoref{tab:supported games} at all.}

\section{REM vs VEM: Speed performance}
\label{app:speed_perfs}
The following graph shows that we the RAM Extraction Method of OCAtari is, in average, $50 \times$ computationally more efficient than the Vision Extraction method. 

\end{document}